\documentclass{article}

\usepackage[preprint]{neurips_2022}

\usepackage{elasticrow}

\usepackage[utf8]{inputenc} % allow utf-8 input
\usepackage[T1]{fontenc}    % use 8-bit T1 fonts
\usepackage{hyperref}       % hyperlinks
\usepackage{url}            % simple URL typesetting
\usepackage{multirow}
\usepackage{booktabs}       % professional-quality tables
\usepackage{amsfonts}       % blackboard math symbols
\usepackage{nicefrac}       % compact symbols for 1/2, etc.
\usepackage{microtype}      % microtypography
\usepackage{xcolor}         % colors

\usepackage[]{graphicx}       % figures
\usepackage{amsmath}

\usepackage{multicol}
\usepackage{tabularx}
\usepackage{booktabs}
\newcolumntype{C}{>{\centering\arraybackslash}X}
\newcolumntype{R}{>{\raggedleft\arraybackslash}X}

\newcommand{\alignedlabel}[2]{%
\parbox{#1}{\scriptsize\centering{\strut#2}}
}

\def\imagepaddingy{0.4cm}
\def\imagepaddingtiny{0.1cm}

\title{Generating Sparse Counterfactual Explanations For Multivariate Time Series}

\author{
  Jana Lang \\
  Section Computational Sensomotorics\\
  Hertie Institute for Clinical Brain Research \\ Centre for Integrative Neuroscience\\
  University of Tübingen, Germany \\
  \texttt{jana.lang@uni-tuebingen.de}
  \And
    Martin Giese\\
  Section Computational Sensomotorics\\
  Hertie Institute for Clinical Brain Research \\ Centre for Integrative Neuroscience\\
  University of Tübingen, Germany \\
  \texttt{martin.giese@uni-tuebingen.de}
    \And
    Winfried Ilg\\
  Section Computational Sensomotorics\\
  Hertie Institute for Clinical Brain Research \\ Centre for Integrative Neuroscience\\
  University of Tübingen, Germany \\
  \texttt{winfried.ilg@uni-tuebingen.de}
  \And
    Sebastian Otte\\
  Neuro-Cognitive Modeling\\
  University of Tübingen, Germany \\
  \texttt{sebastian.otte@uni-tuebingen.de} 
}

\renewcommand{\vec}[1]{\mathbf{#1}}

\newcommand{\query}{\vec{x}_{q}}
\newcommand{\target}{\vec{x}_{t}}
\newcommand{\targetclass}{\vec{c}_{t}}
\newcommand{\residuals}{\vec{\delta}}
\newcommand{\cf}{\vec{x}_{cf}}
\newcommand{\gen}{\mathcal{G}}
\newcommand{\disc}{\mathcal{D}}
\newcommand{\cls}{\mathcal{C}}
\newcommand{\genloss}{\mathcal{L}_{G}}
\newcommand{\discloss}{\mathcal{L}_{D}}
\newcommand{\advloss}{\mathcal{L}_{adv}}
\newcommand{\clsloss}{\mathcal{L}_{cls}}
\newcommand{\simloss}{\mathcal{L}_{sim}}
\newcommand{\sparseloss}{\mathcal{L}_{sparse}}
\newcommand{\jerkloss}{\mathcal{L}_{jerk}}
\newcommand{\lone}{\mathcal{L}_{1}}
\newcommand{\lzero}{\mathcal{L}_{0}}
\newcommand{\ltwo}{\mathcal{L}_{2}}
\newcommand{\ours}{SPARCE} % usage \ours{}

\begin{document}

\maketitle

\begin{abstract}

% Buzzwords die wir haben:
% 
% sparsity
% counterfactual explanations
% (multivariate) time series
% recurrent
% regularized
% LSTM
% GAN
% actionable, plausible, meaningful (haben die anderen aber auch)

%   ***** take two *****\\
Since neural networks play an increasingly important role in critical sectors, explaining network predictions has become a key research topic. Counterfactual explanations can help to understand why classifier models decide for particular class assignments and, moreover, how the respective input samples would have to be modified such that the class prediction changes. Previous approaches mainly focus on image and tabular data. In this work we propose SPARCE \footnote{Code: https://github.com/janalang/SPARCE}, a generative adversarial network (GAN) architecture that generates SPARse Counterfactual Explanations for multivariate time series. Our approach provides a custom sparsity layer and regularizes the counterfactual loss function in terms of similarity, sparsity, and smoothness of trajectories. We evaluate our approach on real-world human motion datasets as well as a synthetic time series interpretability benchmark. Although we make significantly sparser modifications than other approaches, we achieve comparable or better performance on all metrics. Moreover, we demonstrate that our approach predominantly modifies salient time steps and features, leaving non-salient inputs untouched.

%   TL;DR ("Too Long; Didn't Read"): a short sentence describing your paper: \\
%   In this work we propose SPARCE, a GAN architecture that generates SPArse Recurrent Counterfactual Explanations for multivariate time series.
%   \\
  
%   Comma separated list of keywords: Explainable Artificial Intelligence, Counterfactual Explanations, Multivariate Time Series, Generative Adversarial Networks, Long Short-Term Memory

\end{abstract}

\section{Introduction} 
With the advent of machine learning for decision making in critical sectors like healthcare, predictive maintenance, or traffic, serious concerns have been raised about the trustworthiness of these algorithms. In recent years, the field of explainable artificial intelligence (XAI) has therefore gained increasing popularity. While manifold techniques for explaining tabular data and image classifiers have been proposed, temporal data has largely been neglected. In contrast to image data, time series interpretability poses manifold challenges, including the presence of distinct time and space dimensions and an increased difficulty of visualizing information in a meaningful way. Recent work has raised strong concerns about the adaptability of prevalent XAI methods to multivariate time series \citep{ismail2020benchmarking}.

\begin{figure}[h]
\centering
\includegraphics[width=1.4cm]{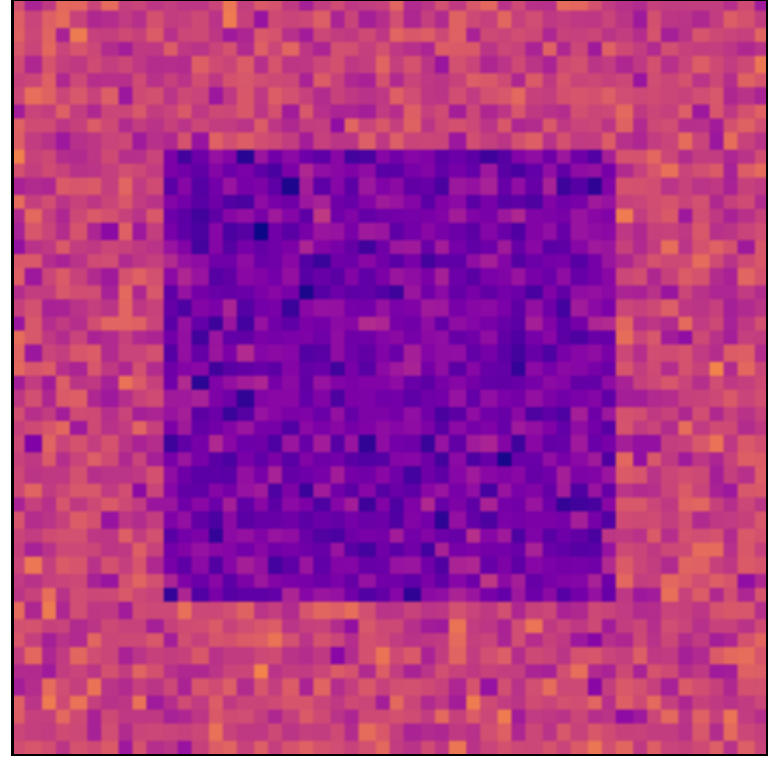} 
\includegraphics[width=1.8cm,trim=0.2cm -4cm 0 0]{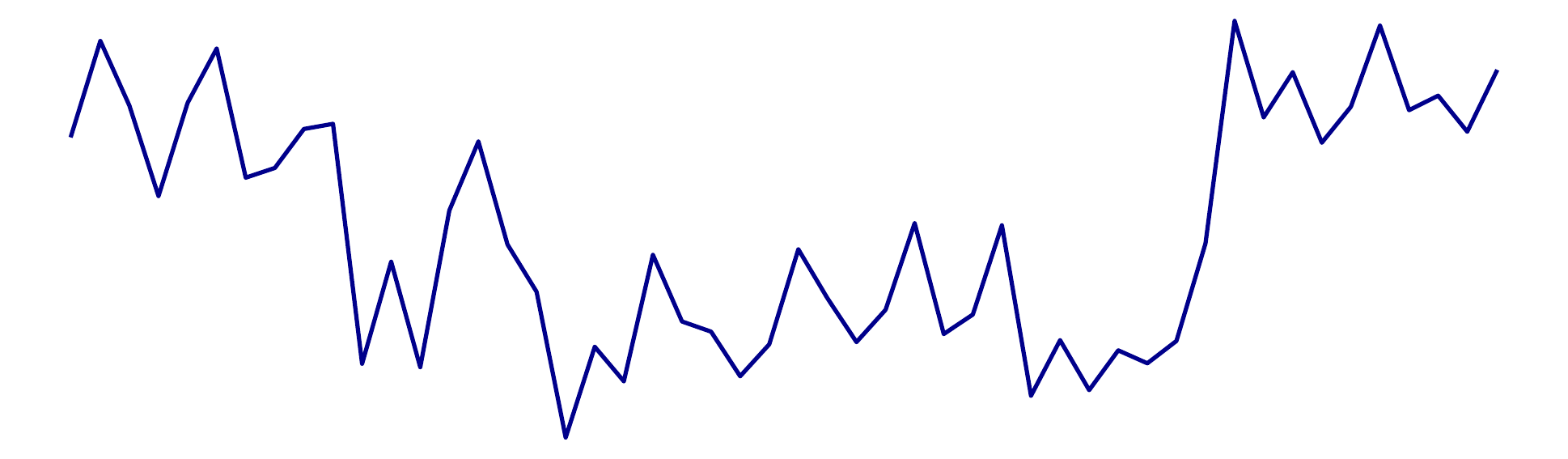} 
\hfill
\includegraphics[width=1.4cm]{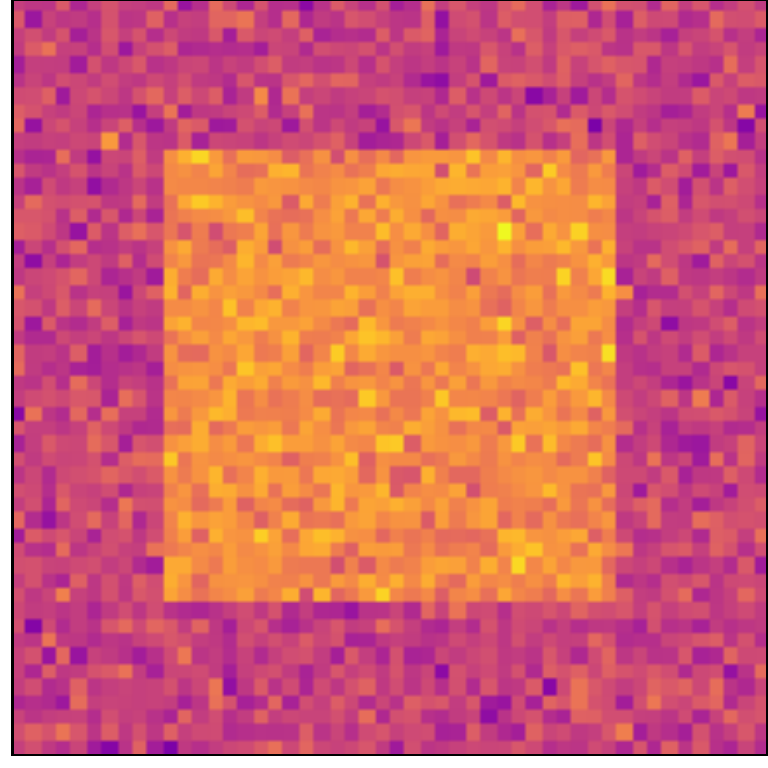}
\includegraphics[width=1.8cm,trim=0 -4cm 0 0]{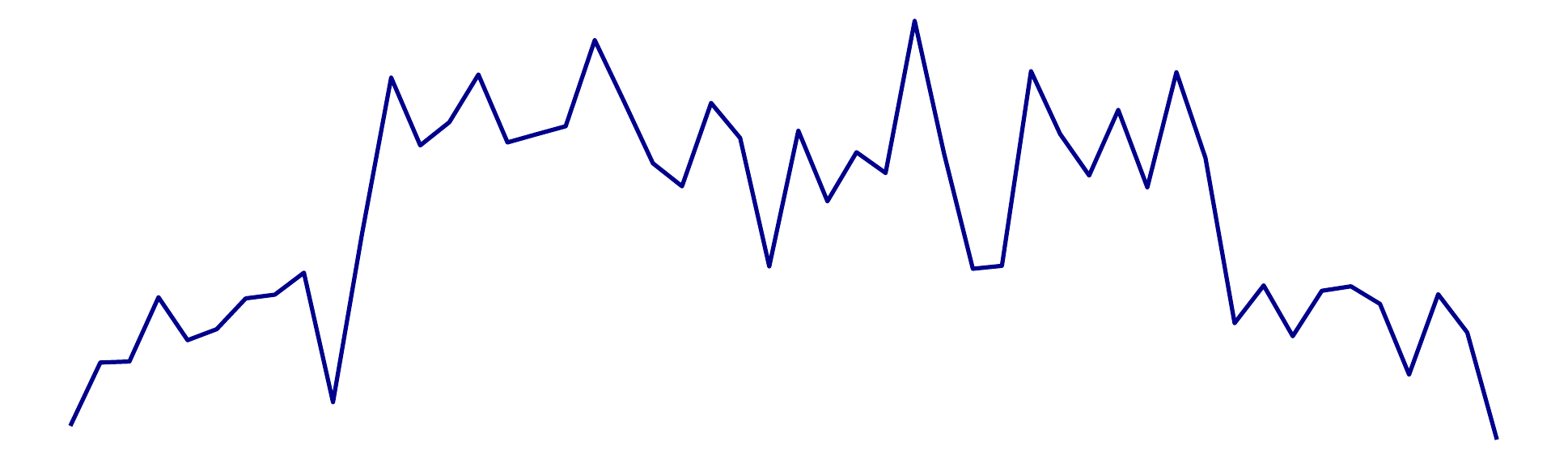} 
\hfill
\includegraphics[width=1.4cm]{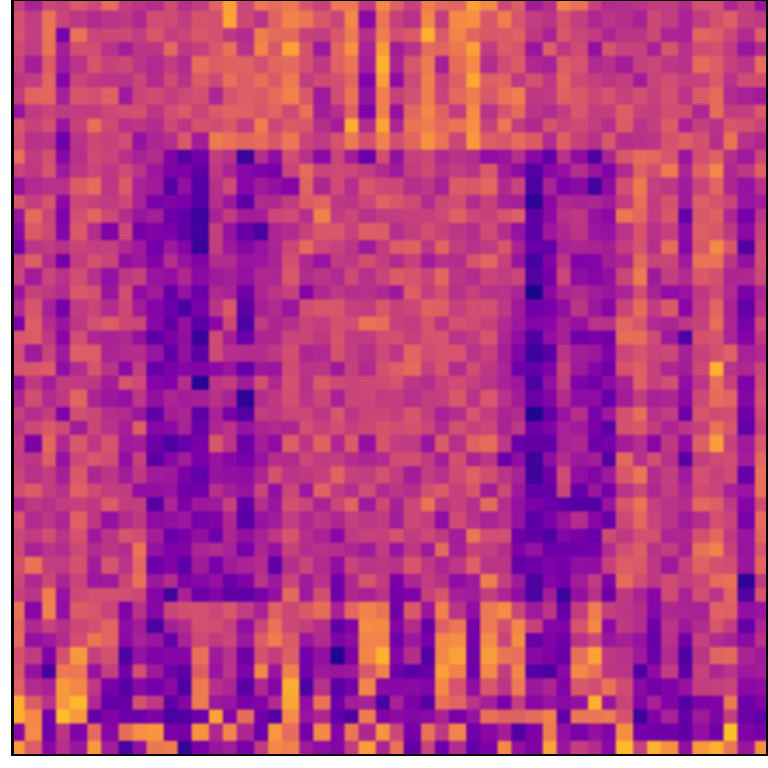}
\includegraphics[width=1.8cm,trim=0 -4cm 0 0]{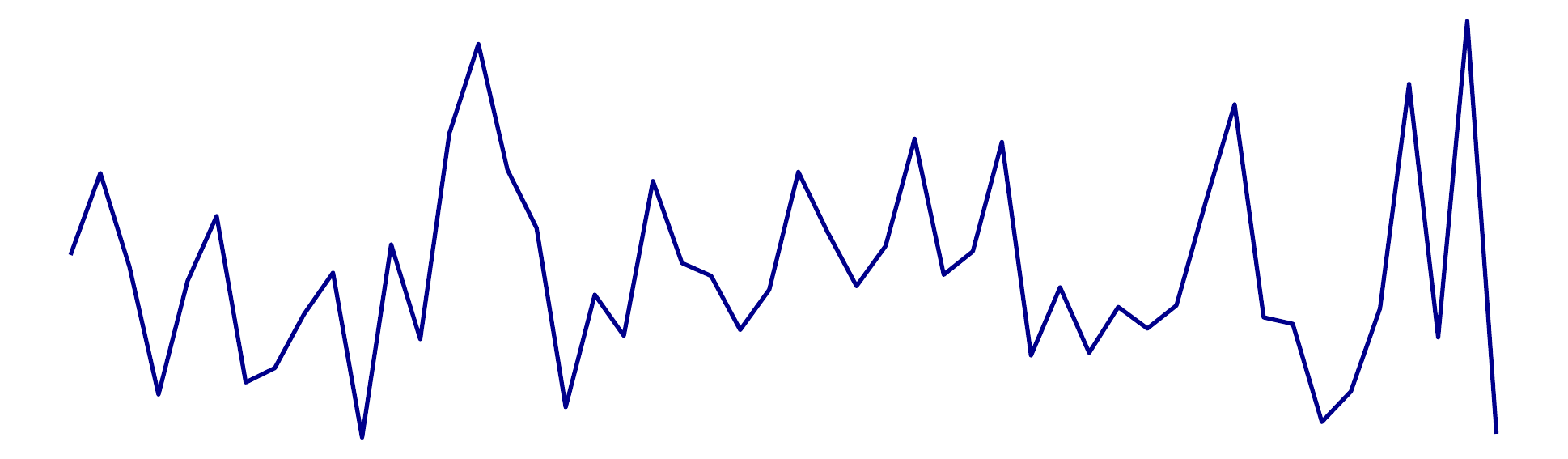} 
\hfill
\includegraphics[width=1.4cm]{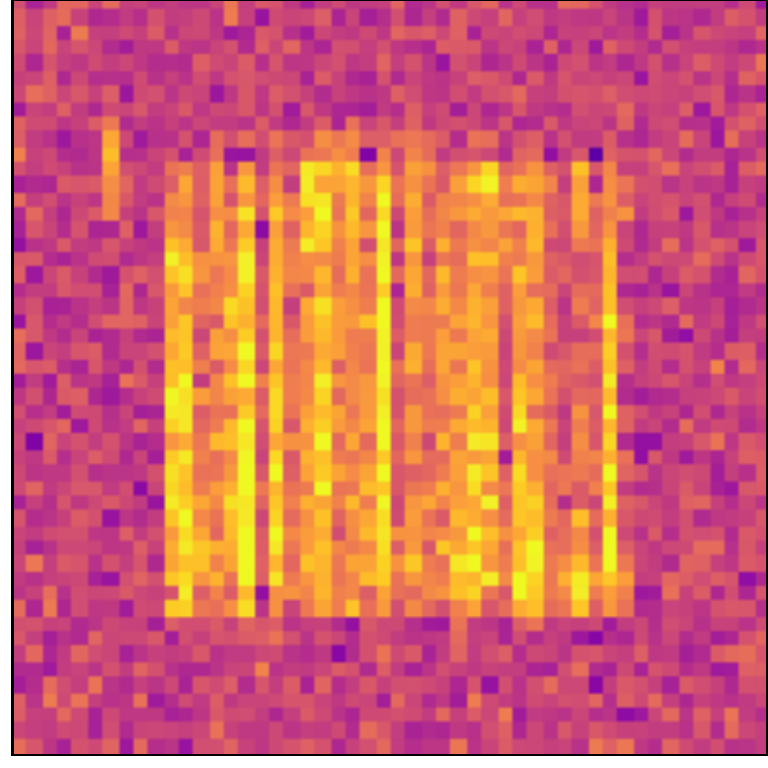}
\includegraphics[width=1.8cm,trim=0 -4cm 0 0]{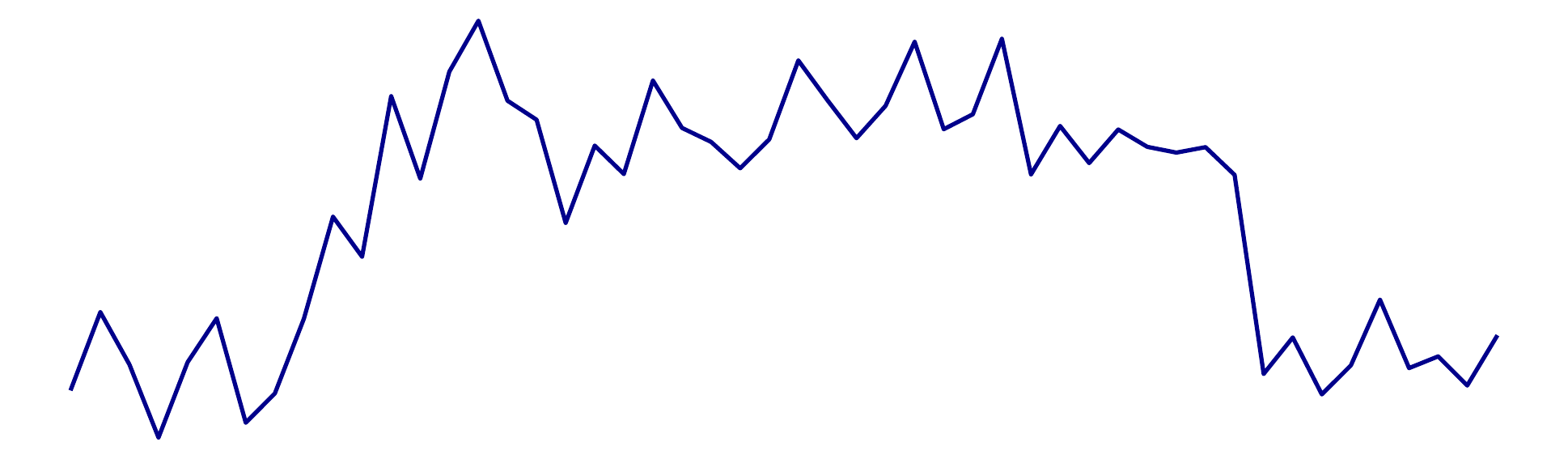} 

\hspace{0.3cm}{\scriptsize{Original}}
\hspace{2.75cm}{\scriptsize{Target}}
\hspace{2.4cm}{\scriptsize{State-of-the-art}}
\hspace{2.0cm}{\scriptsize{\ours{} (ours)}}\hfill{}

%\alignedleftlabel{0.24\linewidth}{~~~~Original}\hfill
%\alignedleftlabel{0.24\linewidth}{~~~~~Target}\hfill
%\alignedleftlabel{0.24\linewidth}{~~~~State-of-the-art}\hfill
%\alignedleftlabel{0.24\linewidth}{~~~~\ours (ours)}

%    Original & Salient & ICS & GAN & CounteRGAN & \ours (Ours) & Ours $\lambda_{4,5} = 0$
\caption{Counterfactuals generated using a state-of-the-art approach \citep{nemirovsky2020countergan} and our approach for a multivariate time series. Columns represent features and rows represent time steps. The curves arranged right to the boxes depict respective sequences for one of the center features.}
\label{fig:sparse}

\end{figure}

\paragraph{Counterfactual Explanations}
Derived from philosophical reasoning, \textit{counterfactual explanations} try to find modifications to an input query so that the classification changes to a desired class \citep{wachter2017counterfactual}. Features of the input query can be mutable, i.e. the values can and may be modified, or immutable. A valid counterfactual should only modify mutable input features \citep{karimi2021algorithmic}. Meaningful counterfactual explanations can guide users towards a better understanding of decisions made by a system. If a classifier predicts a certain disease risk based on a patient's medical record, it is helpful to understand not only what factors led to the decision, but also what factors would have to change and in which way to minimize the risk.

\subsection{Objectives For Counterfactual Explanations}
\paragraph{Precision, Similarity \& Realism} A valuable counterfactual explanation is close to the original data point, looks plausible and realistic and suggests actionable modifications \citep{dandl2020multi, keane2020good}. The choice of distance functions to measure the actionability of a counterfactual has been a topic of discussion. The original approach by \citet{wachter2017counterfactual} iteratively minimizes the distance between the predicted class for the counterfactual and the target class (via $\ltwo$ norm) as well as the distance between query and counterfactual (via $\lone$ norm) using gradient descent. \citet{dandl2020multi} additionally assess realism of the generated counterfactual by measuring how likely it is that the counterfactual stems from the observed data distribution.

%\paragraph{Sparsity \& Plausibility} 
\paragraph{Sparsity} \citet{dandl2020multi} implement sparsity as the $\lzero$ norm between query and counterfactual, that measures how many features were changed to go from the original data point to the counterfactual. \citet{mothilal2020explaining} do not include sparsity into the loss function, but modify the generated counterfactuals post-hoc using a greedy algorithm to set increasingly more features with smaller modifications zero until the prediction changes. In contrast, \citet{keane2020good} define a rigid threshold for sparsity stating that a good counterfactual for tabular data may only modify up to two features. Adapting this paradigm to time series, \citet{delaney2021instance} only allow for modification of one single contiguous section of the time series. Others only ensure feature sparsity, while modifying all time steps of the sequence \citep{ates2021counterfactual}. 
%\citep{keane2021if} highlighted that a counterfactual can be very close to the query instance while at the same time violating common-sense. Some approaches have therefore introduced plausibility constraints to ensure that the query is modified in a plausible way, e.g. by defining preset feature ranges for each feature in the dataset \citep{mothilal2020explaining, karimi2020model}.

\paragraph{Similarity vs. Sparsity}
Figure \ref{fig:sparse} demonstrates why similarity alone does not guarantee actionability. The counterfactual generated by our approach makes sparse, but more substantial modifications, while the counterfactual generated using a state-of-the-art approach makes minor changes in all time steps and features. If solely regularized by the $\lone$ norm (i.e. the similarity constraint), the latter would be preferred. Taking the actionability of the counterfactual into account, one would most likely prefer the counterfactual generated by SPARCE despite the higher $\lone$ loss. As a consequence, sparsity plays a central role in our approach.

\subsection{Generative Approaches}
To generate more realistic and plausible counterfactuals, while overcoming high computational costs of iterative optimization methods, generative adversarial networks (GANs) have recently been introduced for the generation of counterfactual explanations \citep{nemirovsky2020countergan, van2021conditional}. GANs have become popular for generating realistic looking fake images by training a generator to create fake samples that a discriminator would erroneously perceive as real samples \citep{goodfellow2014generative}. GAN-based architectures for counterfactual search add a classifier to the standard GAN approach. In this way, the generator learns to produce realistic looking counterfactuals that change the classifier's prediction to a target class.

While \citet{nemirovsky2020countergan} only evaluate their model on image and tabular data, \citet{van2021conditional} also assess their approach on univariate time series. Both approaches use $\lone$ or $\ltwo$ norms as regularization terms that act on the generator's loss function. Particularly for multivariate time series, this formulation is problematic, since it creates proximate, but not sparse counterfactuals. Indeed, sample counterfactuals generated by \citet{van2021conditional} modify every single time step of the query sequence. In some domains, this might be necessary. However, it is questionable whether such a counterfactual explanation would have any explanatory power. Besides, it is unclear whether these modifications could actually be acted upon in reality. Our approach is thus designed to create truly sparse counterfactual explanations for multivariate time series without compromising other important objectives of counterfactuals, including realism, similarity, and plausibility. 

\section{Method}
\label{sec:method}
Motivated by the insufficient adaptation of counterfactual approaches to multivariate time series, we propose SPARCE: a novel framework to efficiently generate SPARse Counterfactual Explanations for multivariate time series data. Our approach aims to change the class label of an original time series to a target class (\textit{precision}). Generated counterfactuals should be within the distribution of the original data points (\textit{realism}) and stay as close as possible to the query sequence (\textit{similarity}). In contrast to related approaches for multivariate time series, we postulate that counterfactuals are time- and feature-sparse, i.e. that only a subset of features and time steps is modified (\textit{sparsity}). Finally, for applications where time series evolve smoothly over time, we aim to modify the original data point in a temporally plausible manner (\textit{smoothness}).

\subsection{Generating Counterfactual Explanations}
Basing our approach on a generator-discriminator architecture, we ensure realism of the generated samples. In line with \citet{nemirovsky2020countergan} we define a modified generator $\gen$ which learns to generate residuals $\residuals = \gen(\query)$ from the input sample $\query$. In contrast to standard GANs, the generator does not use a random seed, but real samples as inputs. Thus, original samples are first divided into queries $\query$ and targets $\target$. Targets are samples labeled as the target class $\targetclass$ and are used as real examples for the discriminator $\disc$. The query subset contains all other samples and is presented as inputs to the generator $\gen$. Residuals created by the generator are added to the query to produce a counterfactual ($\cf = \query + \residuals$). A pre-trained classifier $\cls$ determines the class prediction for the generated counterfactual. At the same time, the counterfactual is presented to the discriminator as a fake sample. In combination with real target samples, the discriminator tries to distinguish between real and fake (i.e. generated) samples. The realism of the generated counterfactual examples increases as the generator learns to fool the discriminator. The classifier prevents the generator from producing zero-residuals, i.e. from learning the identity function.

\begin{figure}
  \centering
  %fbox{\rule[-.5cm]{0cm}{4cm} \rule[-.5cm]{4cm}{0cm}}
  % \includegraphics[width=1.0\linewidth]{approach.pdf}
  \includegraphics[width=1.0\linewidth]{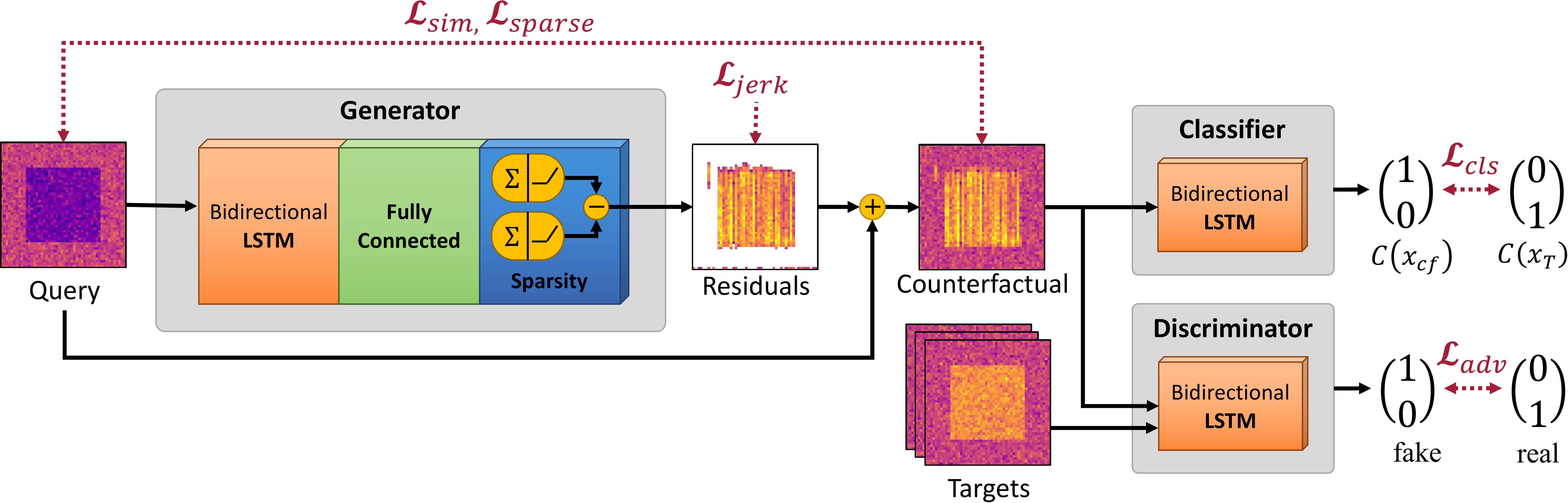}
  \caption{Schematic illustration of our GAN-based approach for counterfactual search. Inputs are divided into query and target time series (displayed as heatmaps) according to the desired target class. A recurrent generator with sparsity activation generates residuals for each query. Residuals are added to the corresponding query to create a counterfactual explanation. A pretrained sequence classifier predicts the class label of the counterfactual. A recurrent discriminator tries to distinguish counterfactuals from real targets.}
  \label{fig:approach}
\end{figure}

\begin{minipage}[t]{0.495\textwidth}
\begin{equation}
\label{eqn:advloss}
\advloss = -\log(\disc (\query + \gen(\query)))
\end{equation}
\end{minipage}\hfill
\begin{minipage}[t]{0.495\textwidth}
\begin{equation}
\label{eqn:clsloss}
\clsloss = - \targetclass \log(\cls(\query + \gen(\query)))
\end{equation}
\end{minipage}

\begin{minipage}[t]{0.495\textwidth}
\begin{equation}
\label{eqn:simloss}
\simloss = \| \query - \cf \|_1
\end{equation}
\end{minipage}\hfill
\begin{minipage}[t]{0.495\textwidth}
\begin{equation}
\label{eqn:sparseloss}
\sparseloss = \| \query - \cf \|_0
\end{equation}
\end{minipage}

\begin{equation}
\label{eqn:jerkloss}
\jerkloss = \sum_{t=0}^{T-1} \| \residuals^{t+1} - \residuals^{t} \|_2
\end{equation}
\begin{equation}
\label{eqn:genloss}
\genloss = \mathbb{E}_{\query} \left[ \lambda_{1} \advloss + \lambda_{2} \clsloss + \lambda_{3} \simloss + \lambda_{4} \sparseloss + \lambda_{5} \jerkloss \right]
\end{equation}
\begin{equation}
\label{eqn:discloss}
\discloss = \frac{1}{2}\ \mathbb{E}_{\query} \left[ - \log(\disc (\query)) \right] - \mathbb{E}_{\query} \left[ \log(1 - (\disc (\query + \gen(\query)))) \right]
\end{equation}

\paragraph{Generator}
\label{sec:generator}
The generator is realized with a many-to-many sequence prediction model trained to generate modifications to a query sequence. To capture temporal dependencies in the input, different types of sequence models can be chosen, including long short-term memories (LSTMs), gated recurrent units, or temporal convolutional neural networks. Input and output of the generator are of the same shape. Loss functions for generator and discriminator derive from the minimax loss suggested by \citet{goodfellow2014generative}. The generator maximizes the discriminator's estimate that the counterfactual is real (Equation \ref{eqn:advloss}). One important aspect of the generator is the subtractive dual ReLU \citep{nair2010rectified} output in the sparsity layer. Instead of a single linear output the two contrastive outputs allow the network to produce positive and negative residuals while it is still easy to generate exact zero-residuals ($\residuals = ReLU(\residuals_{pos}) - ReLU(\residuals_{neg})$).

\paragraph{Immutable Features} In case of immutable features in the original dataset, the generator only produces residuals for all mutable features. In this specific case, the input to the generator is larger than its output. Generated residuals for the mutable features are then likewise added to the respective mutable features in the query sequence. All immutable features of the query instance remain untouched.

\paragraph{Discriminator} The discriminator takes on the role of distinguishing between real samples (i.e. samples from the original dataset) and fake samples (i.e. generated counterfactuals). It aims to maximize its estimate that the counterfactual is fake and the query is real (Equation \ref{eqn:discloss}). It is implemented as a binary many-to-one sequence classification model with sigmoid activation that takes in a multivariate time series and produces a probability between 0 and 1, indicating whether the given sample looks like a real or fake sample. As the counterfactuals begin to look more realistic, the discriminator's accuracy drops towards 50\% (chance).

\paragraph{Classifier} Unlike vanilla GANs, a counterfactual GAN needs a third neural network, the classifier. In our approach the classifier is realized with a many-to-one sequence classification model. The classifier is pretrained on the original dataset and learns to classify the label of a sequence. In contrast to \citet{nemirovsky2020countergan}, our classifier does not only distinguish between samples which belong and samples that do not belong to the target class. Instead, we train a full classifier which learns to distinguish all classes in the original dataset. That said, our classifier can either be binary (with sigmoid activation) or multi-class (with softmax activation) in case of two or multiple original class labels, respectively. This property allows us to flexibly alter the desired target class for the generated counterfactuals without retraining the classifier. Moreover, our approach could also simultaneously be trained on all target classes. In this case, generating counterfactuals for different target classes would not require retraining of any network element of our approach.

\paragraph{Regularization}
\label{sec:regularization}
The combination of adversarial loss $\advloss$ and classification loss $\clsloss$ loss ensures that the generated counterfactual changes the class label, while resembling a sample from the original data distribution. The classification loss between the predicted class for the counterfactual and the target class is derived from the cross-entropy loss (Equation \ref{eqn:clsloss}). In line with other counterfactual approaches, we apply the $\lone$ norm as a similarity regularization term $\simloss$ on the generator loss (Equation \ref{eqn:simloss}). Importantly, we also use the $\lzero$ norm as a real sparsity constraint $\sparseloss$ which ensures that the number of modifications stays low (Equation \ref{eqn:sparseloss}). It was shown that $\lzero$ regularization effectively fosters sparse hidden state updates in RNNs  \citep{gumbsch2021sparsely}. To address the sequentiality of time series, we introduce another regularization term, the \textit{jerk} constraint $\jerkloss$. This term ensures that changes are evenly distributed over time by penalizing large differences between modifications in consecutive time steps (Equation \ref{eqn:jerkloss}). Additional weighting factors $\lambda_{1-5}$ allow each component of the generator loss to be switched on or off to meet the specific needs of individual datasets. A more fine-grained weighting with weighting factors between 0 and 1 enables a direct influence on the loss balance (Equation \ref{eqn:genloss}).

\subsubsection{On the Sparsity of Generated Counterfactuals}
One key difference of our model in comparison with other counterfactual approaches is the clear distinction between similarity and sparsity. The combination of the sparsity constraint $\sparseloss$ and the sparsity layer as part of the generator architecture produces truly sparse counterfactuals with zero-residuals in a number of time steps and features. Importantly, we let the system inherently learn the trade-off between realism, precision, similarity, sparsity and smoothness during the training process. As a consequence, unlike other counterfactual approaches for time series, there is no need to define a fixed number of time steps and features that which may be changed. On the same lines, there is not only one specific section of the series which can be modified. Instead, we demonstrate that our approach identifies and modifies salient time steps and features while leaving most non-salient time steps and features untouched. 

\section{Experiments}
\label{sec:experiments}
Our approach is evaluated on three different multivariate time series datasets in comparison with three related counterfactual methods. The evaluated tasks comprise two movement datasets for multi-class classification and one synthetic time series interpretability benchmark for binary classification. Human motion datasets are anonymized and cannot be mapped back to individual subjects.

\subsection{Datasets}
\label{sec:datasets}
%\paragraph{MotionSense}
\textbf{MotionSense:} The human motion dataset \textit{MotionSense} \citep{Malekzadeh:2019:MSD:3302505.3310068} (Open Database License ODbL) provides multivariate time series collected by accelerometer and gyroscope sensors of a smartphone stored in a subject's pocket as they perform different actions. Actions include walking downstairs, walking upstairs, sitting, standing, walking and jogging. For this work, we only used active movement sequences and thus excluded sitting and standing trials which yielded a total number of 11194 samples. Each time series was truncated to a length of 100 time steps. All twelve features describing attitude, gravity, rotation and user acceleration are treated as mutable.

%\paragraph{Catching}
\textbf{Catching:} The \textit{Catching} dataset \citep{lang2021early} (provided by personal permission) contains multivariate two-dimensional movement trajectories of healthy and pathological ball catching trials over 60 time steps. At each time step, 20 features capture the catcher's arm position as well as the position of the ball. Each of the 1975 catching trials is assigned a label indicating the subject's disease status: healthy control, patient with Autism Spectrum Disorder or patient with Spinocerebellar Ataxia. All features specifying the catcher's body posture are defined as mutable features, while the two features describing the ball position are treated as immutable.

%\paragraph{Moving Box}
\textbf{Moving Box:} The synthetic \textit{Moving Box} dataset was introduced to benchmark interpretability in time series predictions \citep{ismail2020benchmarking}. It portrays a wide range of temporal and spatial properties commonly found in multivariate time series. Each time series spans 50 time steps and 50 features of which only a subset is salient. Samples are assigned a binary label (0: negative class, 1: positive class) and have a defined start and end point of salient time steps and features per sample. In this dataset, all features are mutable. We used a representative subset containing 13950 samples with boxes of different sizes and at varying positions as well as a variety of generating time series processes.
\subsection{Approaches}
\label{sec:approaches}
\textbf{ICS:} We loosely follow \citet{wachter2017counterfactual} for an implementation of an iterative counterfactual search algorithm. Each counterfactual is initialized with a random uniform distribution between the minimum and maximum values of the query sequence. The class of the generated counterfactual is predicted using a pretrained classifier. We use the $\ltwo$ distance to measure the classification loss and the unweighted $\lone$ norm to enforce similarity. 

All following approaches are based on GANs combined with a pretrained classifier. To account for temporal dependencies in the data, generator and discriminator are implemented as bidirectional LSTMs \citep{hochreiter1997long}. The generator is a two-layer many-to-many bidirectional LSTM with 256 hidden neuron and dropout of 0.4. The discriminator is built up as a one-layer many-to-one bidirectional LSTM with 16 hidden neurons, sigmoid output activation and dropout of 0.4. For both networks, the final LSTM layer is followed by a fully-connected output layer. 

\textbf{GAN:} This approach consists of a counterfactual LSTM-GAN producing complete counterfactuals based on query sequences. The fully-connected output layer of the generator is followed by a tanh activation. The generator loss is regularized using the $\lone$ norm to optimize the distance between counterfactual and query. 

\textbf{CounteRGAN:} This approach is a time series specific implementation of \citet{nemirovsky2020countergan} and implements an LSTM generator that produces residuals based on query sequences. All other aspects of the implementation are equal to the GAN approach. 

\textbf{SPARCE:} Our approach likewise generates residuals instead of complete counterfactuals. In comparison to CounteRGAN, we additionally regularize the generator loss via sparsity and smoothness constraints (cf. Section \ref{sec:regularization}). Moreover, we add weighting factors $\lambda_{1-5}$ to enable the (de-)activation of single regularization constraints if required. Most importantly, the LSTM generator implemented in our approach does not conclude with a linear or tanh activation layer, but instead uses a custom sparsity layer of two interoperating ReLU activations (cf. Section \ref{sec:generator}).

\subsection{Evaluation Metrics}
\label{sec:metrics}
%In accordance with the previously defined objectives for meaningful and actionable counterfactuals (cf. Section \ref{sec:method}), we evaluate all approaches using the following criteria: 

%\paragraph{Realism} 
\textbf{Realism: } In line with \citet{yoon2019time}, we use \textit{t-distributed stochastic neighbor embedding} (\textit{t-SNE}) for a visual assessment of the in-distributionness of the generated counterfactuals \citep{van2008visualizing}. We separately plot query and target samples of the original dataset along with the counterfactuals generated by each approach to determine whether the generated counterfactuals rather resemble queries or targets.

%\paragraph{Precision} 
\textbf{Precision: } Classification error of generated counterfactuals is measured by the $\ltwo$ norm between the classifier's prediction for a counterfactual sequence and the target class. The metric is indicated as the average distance across all test samples. The lower the metric, the higher the precision of the counterfactual approach. A precision value of 0.0 means that all generated counterfactuals were correctly classified as the target class. 

%\paragraph{Similarity}
\textbf{Similarity: } The $\lone$ distance between each query and the corresponding counterfactual is used to assess similarity. The metric is averaged over all test samples and normalized using the number of time steps and features in the dataset. Lower values indicate higher mean proximity of the generated counterfactuals to the corresponding queries.

%\paragraph{Sparsity} 
\textbf{Sparsity: } Generated counterfactuals of each approach are evaluated on the number of modified time steps and features to transform the query into the counterfactual using the $\lzero$ norm between queries and corresponding counterfactual examples. Values are averaged and normalized in the same way as the similarity metric. Here lower values represent higher average sparsity, i.e. fewer modifications in the time and feature dimensions. In the case of immutable features in the dataset, the sparsity metric is only computed on all mutable features. As a consequence, the maximum sparsity value equals 1.0 indicating that all features in all time steps have been modified in each counterfactual.

%\paragraph{Smoothness} 
\textbf{Smoothness: } This time series specific metric is assessed with the $\ltwo$ distance between modifications of consecutive time steps. High values indicate large differences between modifications in subsequent steps. Lower values represent modifications that are more smoothly distributed over the course of the sequence. This metric is likewise averaged across all samples and normalized using the number of time steps and features.

\section{Results}
\subsection{Quantitative Evaluation}
\label{sec:quanteval}
All results are reported on the held-out subsets for testing (20\% of each dataset). Unless otherwise stated, results are averaged over five repetitions with random seeds. The target class for counterfactuals is healthy control for \textit{Catching}, walking upstairs for \textit{MotionSense} and class 1 for \textit{Moving Box}. ICS is performed for 100 steps ($\lambda_{init} = 1.0$, max. $\lambda$ steps = 10). The loss is minimized with Adam \citep{kingma2014adam} optimization ($lr = 0.4, \beta_1 = 0.9, \beta_2 = 0.999$). All GAN-based approaches are trained for 100 epochs in batches of 32 samples using Adam optimization ($lr = 0.0002, \beta_1 = 0.5, \beta_2 = 0.999$). For all quantitative metrics, lower values represent better performance. The best value for each metric is printed in bold numbers. 

Considering the \textit{Catching} dataset, GAN and CounteRGAN achieve a precision of 100\%, however closely followed by our approach (Table \ref{tab:results}). \ours{} outperforms ICS, GAN and CounteRGAN on the similarity and sparsity of generated counterfactuals and shares the best smoothness value with the CounteRGAN approach. It can be seen that no tested approach besides ours can generate sparse counterfactuals. This observation also holds for the \textit{MotionSense} and \textit{Moving Box} datasets. SPARCE reaches the best or second-best performance on each metric in spite of making considerably sparser modifications than the other approaches. 

\begin{table}
  \caption{Quantitative results for all datasets}
  \label{tab:results}
  \centering
  \footnotesize
  \begin{tabularx}{\linewidth}{XXCCCc}
    \toprule
    %\cmidrule(r){1-5}
    Dataset & Measure  & ICS  & GAN & CounteRGAN & \ours{} (Ours) \\
%    \midrule
%    \multicolumn{5}{c}{\textit{Catching} dataset}\\
    \midrule
\multirow{4}{*}{\textit{Catching}}
& Precision  & 0.24 $\pm$ .05 & \textbf{0.00} $\pm$ .00 & \textbf{0.00} $\pm$ .00 & 0.01 $\pm$ .01           \\
    & Similarity & 1.66 $\pm$ .04 & 0.22 $\pm$ .02 & 0.12 $\pm$ .01 & \textbf{0.09} $\pm$ .04       \\
    & Sparsity  & 1.00 $\pm$ .00 & 1.00 $\pm$ .00 & 1.00 $\pm$ .00 & \textbf{0.27} $\pm$ .10       \\
    & Smoothness & 0.55 $\pm$ .01 & 0.07 $\pm$ .01 & \textbf{0.01} $\pm$ .01 & \textbf{0.01} $\pm$ .01\\
%    \midrule
%         & ICS & GAN & CounteRGAN & Ours \\
%    \multicolumn{5}{c}{\textit{MotionSense} dataset}\\
    \midrule
    \multirow{4}{*}{\textit{MotionSense}}
    & Precision  & 0.37 $\pm$ .07 & \textbf{0.00} $\pm$ .00 & \textbf{0.00} $\pm$ .01 & 0.04 $\pm$ .06            \\
    & Similarity & 1.32 $\pm$ .01 & 0.71 $\pm$ .21 & 0.33 $\pm$ .09 & \textbf{0.22} $\pm$ .13       \\
    & Sparsity  & 1.00 $\pm$ .00 & 1.00 $\pm$ .00 & 1.00 $\pm$ .00 & \textbf{0.22} $\pm$ .14       \\
    & Smoothness & 0.58 $\pm$ .00 & 0.09 $\pm$ .01 & \textbf{0.03} $\pm$ .02 & 0.04 $\pm$ .03      \\
%    \midrule
%    \multicolumn{5}{c}{\textit{Moving Box} dataset}\\
    \midrule
    \multirow{4}{*}{\textit{Moving Box}}
    & Precision  & 0.99 $\pm$ .00 & \textbf{0.00} $\pm$ .00 & 0.01 $\pm$ .01 & \textbf{0.00} $\pm$ .00\\
    & Similarity & 1.32 $\pm$ .00 & 0.87 $\pm$ .17 & 0.59 $\pm$ .05 & \textbf{0.40} $\pm$ .06       \\
    & Sparsity  & 1.00 $\pm$ .00 & 1.00 $\pm$ .00 & 1.00 $\pm$ .00 & \textbf{0.30} $\pm$ .05       \\
    & Smoothness & 0.29 $\pm$ .00 & 0.12 $\pm$ .01 & 0.03 $\pm$ .00 & \textbf{0.02} $\pm$ .00      \\
\bottomrule
  \end{tabularx}
 \end{table}

\subsection{Realism}
\begin{figure}
\centering
\begin{elasticrow}[\imagepaddingtiny]
    \elasticfigure{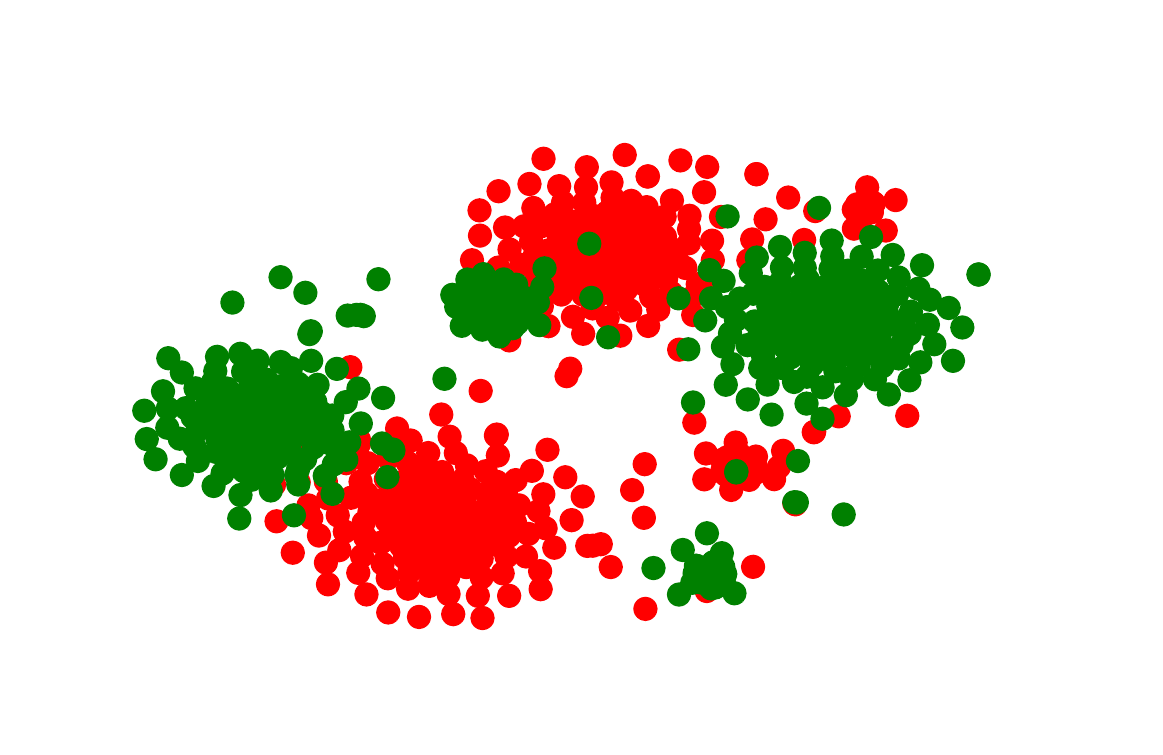}
    \elasticfigure{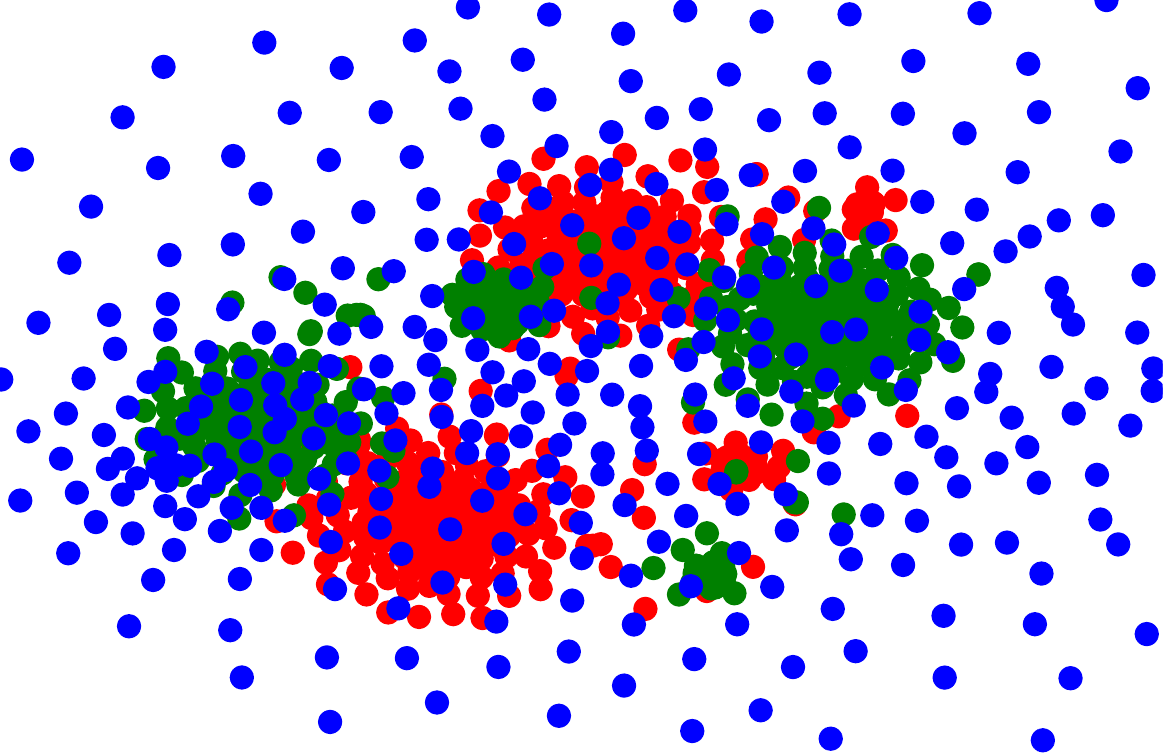}
    \elasticfigure{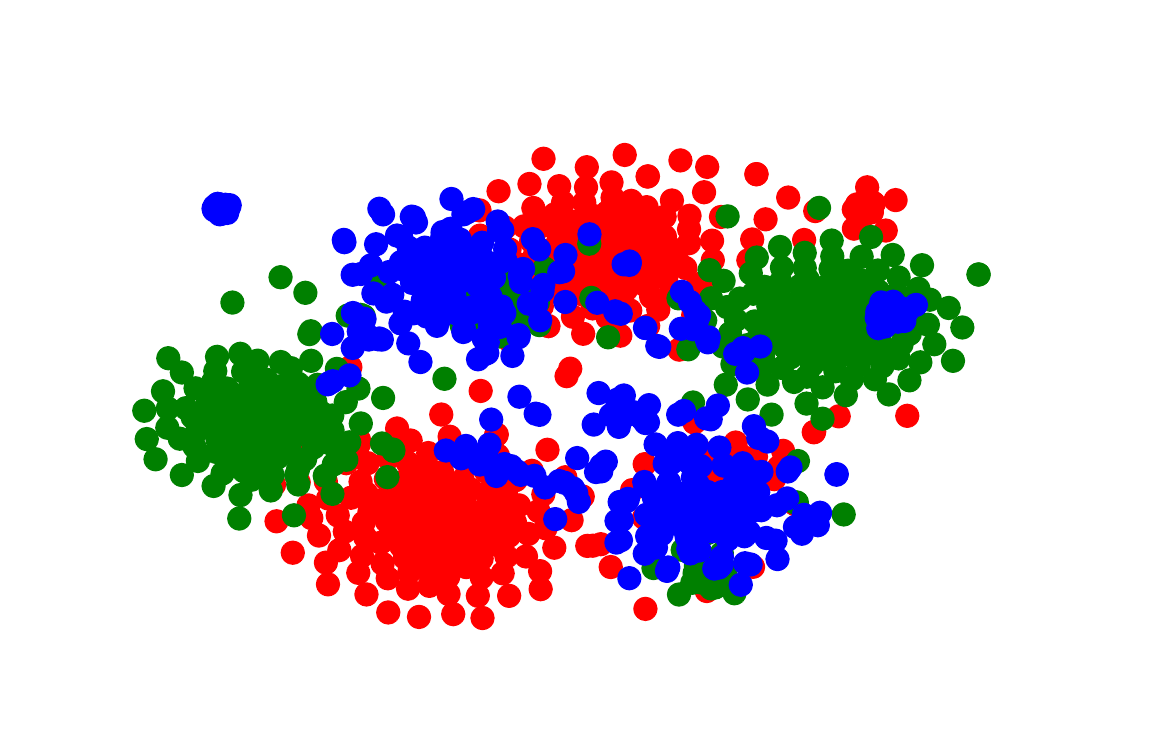}
    \elasticfigure{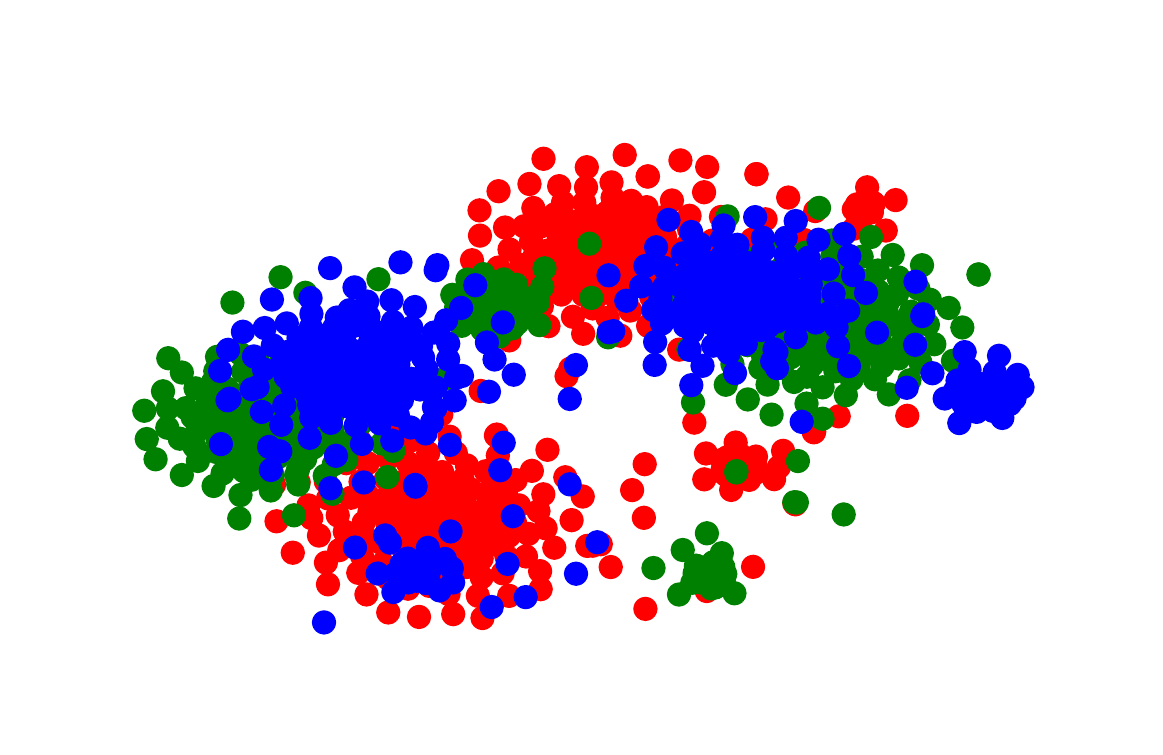}
    \elasticfigure{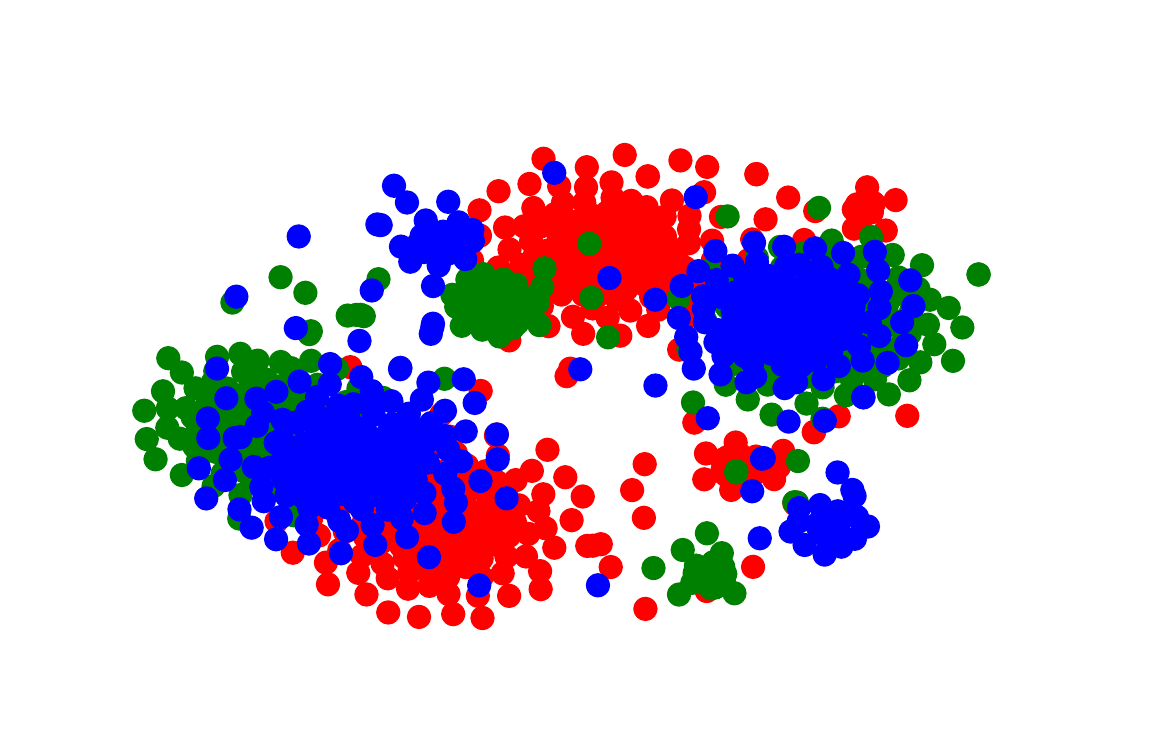}
\end{elasticrow}
\\%   
\alignedlabel{0.19\linewidth}{Original}\hfill
\alignedlabel{0.19\linewidth}{ICS}\hfill
\alignedlabel{0.19\linewidth}{GAN}\hfill
\alignedlabel{0.19\linewidth}{CounteRGAN}\hfill
\alignedlabel{0.19\linewidth}{\ours{} (Ours)}
        
\caption{t-SNE plots for \textit{Moving Box} dataset. Queries are plotted in red, targets in green and generated counterfactuals in blue.}
    \label{fig:tsne}
   
\end{figure}

We qualitatively assess the in-distributionness of generated counterfactuals for the synthetic \textit{Moving Box} dataset via t-SNE visualization ($components = 2, perplexity = 4.4, iterations = 300$). In Figure \ref{fig:tsne}, the first subplot illustrates the distribution of queries and targets in the original dataset. The remaining subplots additionally show the distribution of counterfactuals generated by the respective approaches. While counterfactuals generated by ICS lie within but also largely out of the original distribution, those generated by GAN form separate groups next to queries and targets. Since the task of counterfactual search is to find samples that modify a query sample to look like a target, counterfactuals generated by CounteRGAN and SPARCE show the most promising distributions. Indeed, counterfactuals of both approaches modify queries in a way that the resulting sequences approximate and even overlap with target samples.

\begin{figure}
\centering
%\resizebox{1.0\linewidth}{!}{%
%.135
\begin{elasticrow}[\imagepaddingy]
    \elasticfigure{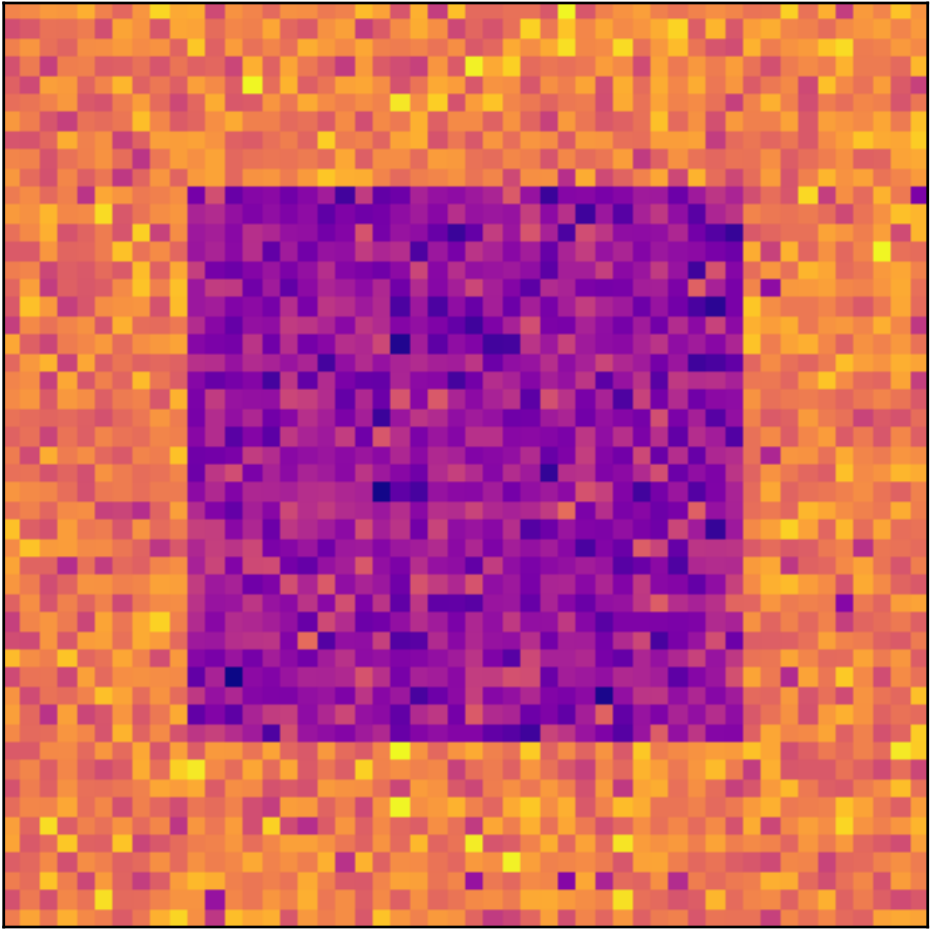} \elasticfigure{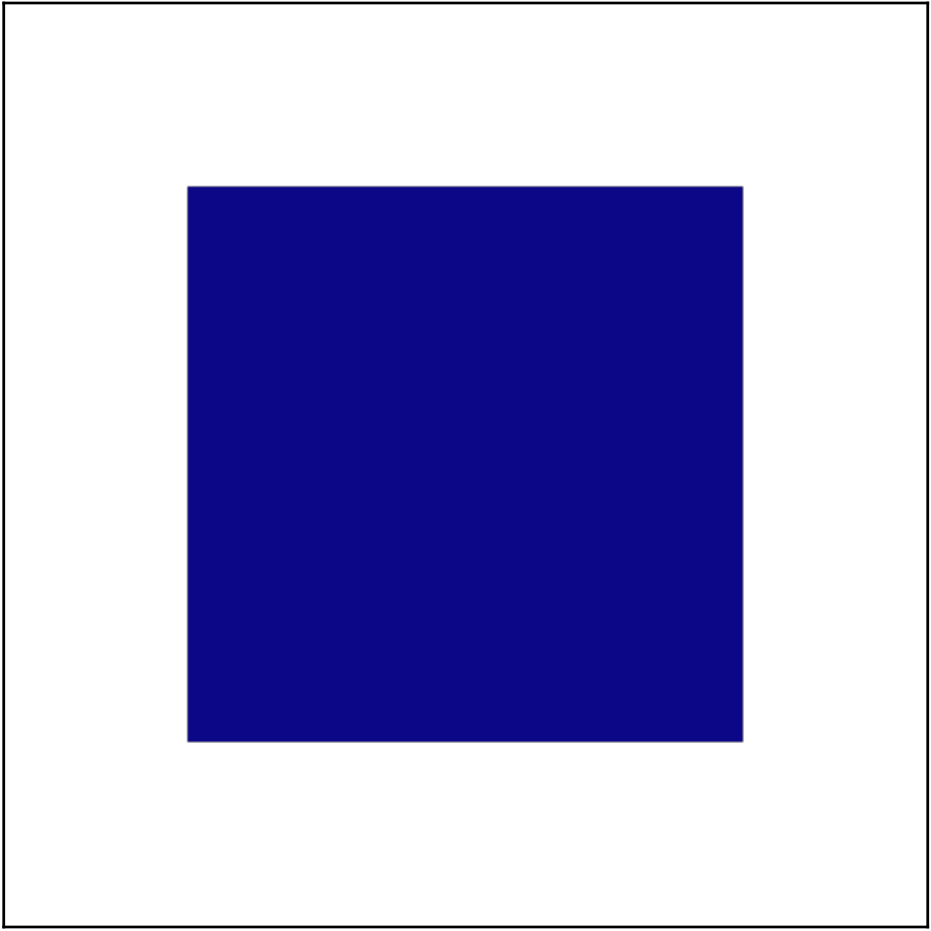}
    \elasticfigure{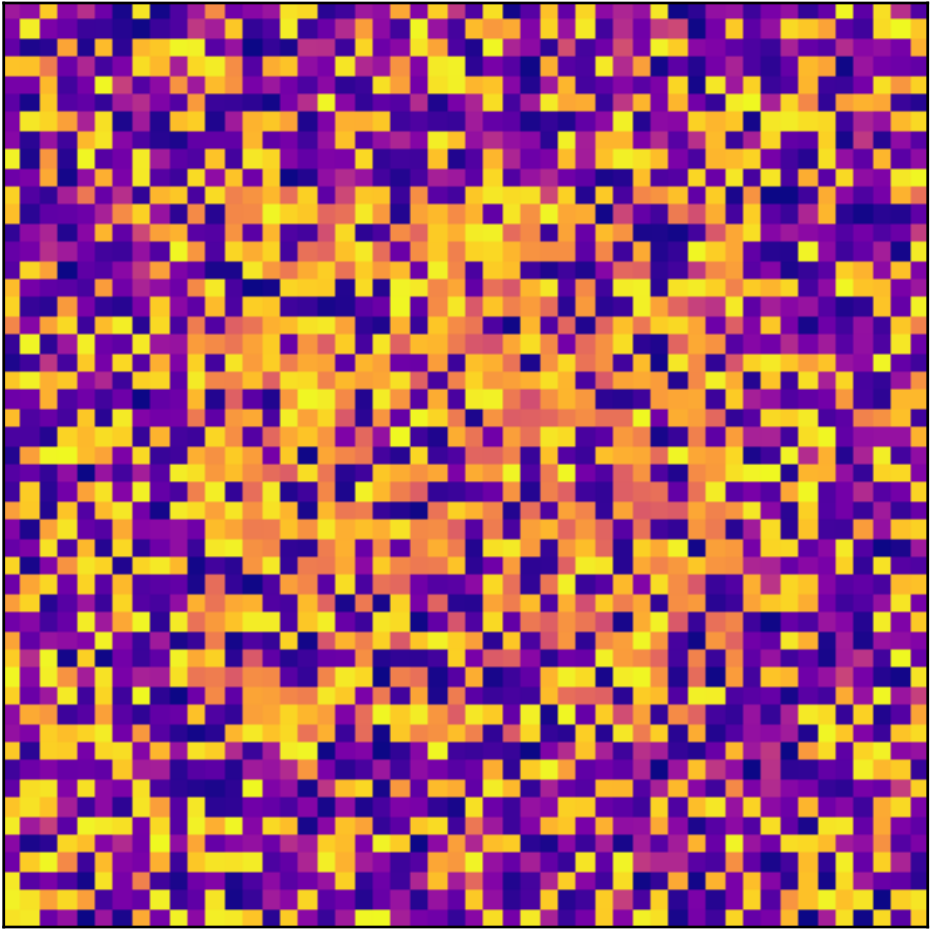}
    \elasticfigure{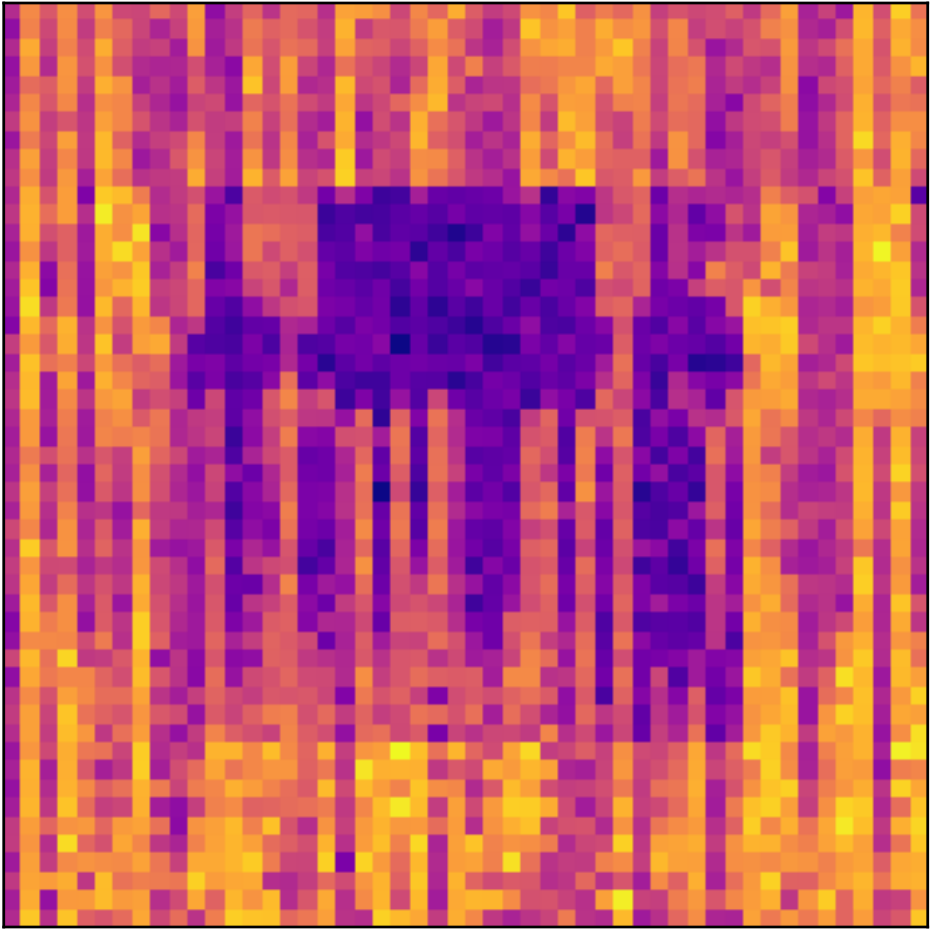}
    \elasticfigure{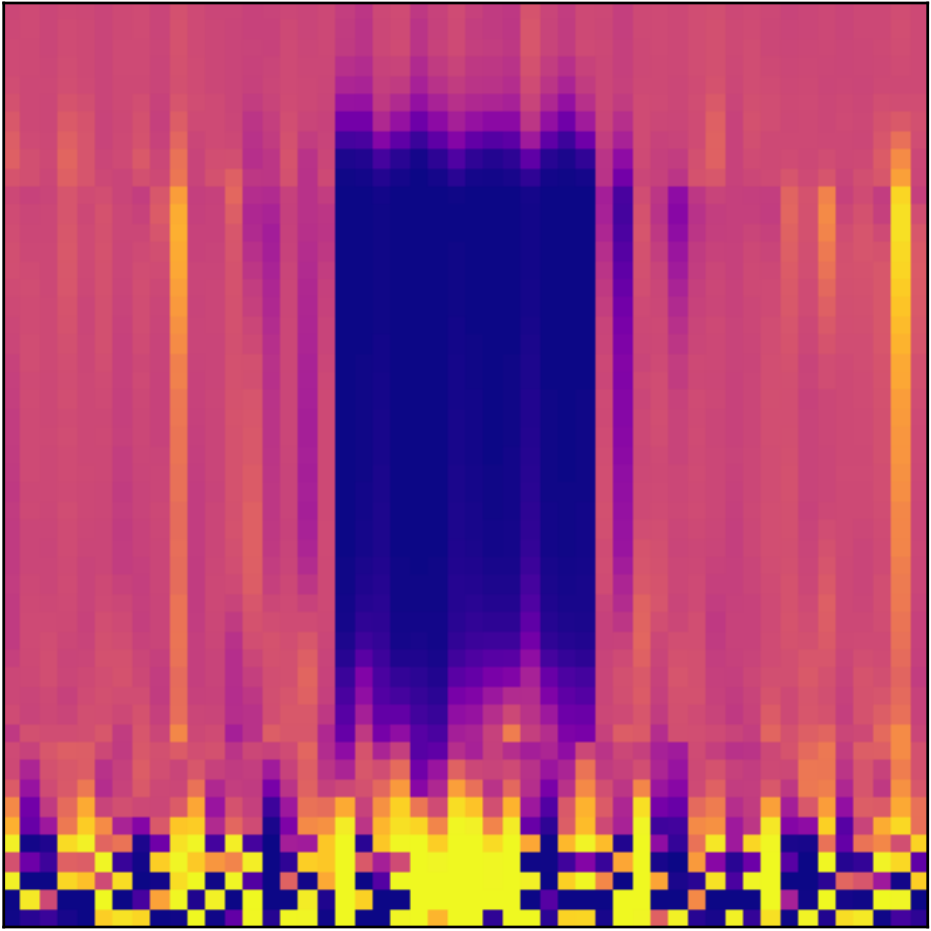}
    \elasticfigure{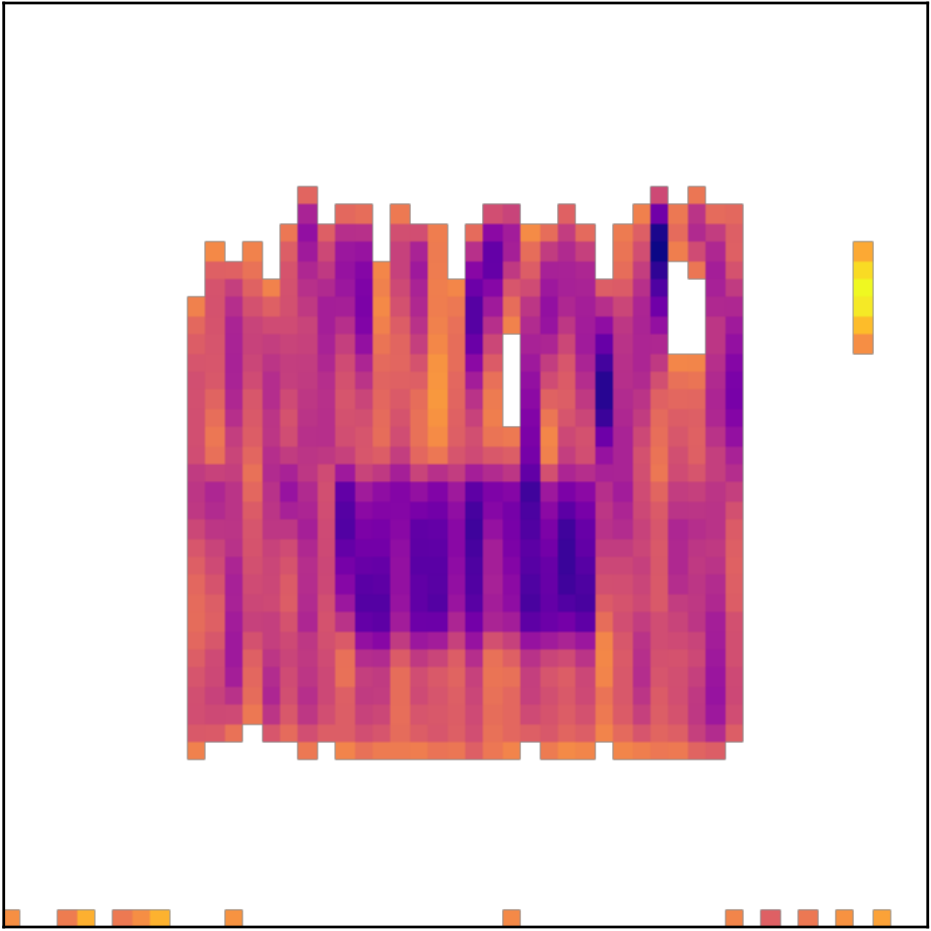}
    \elasticfigure{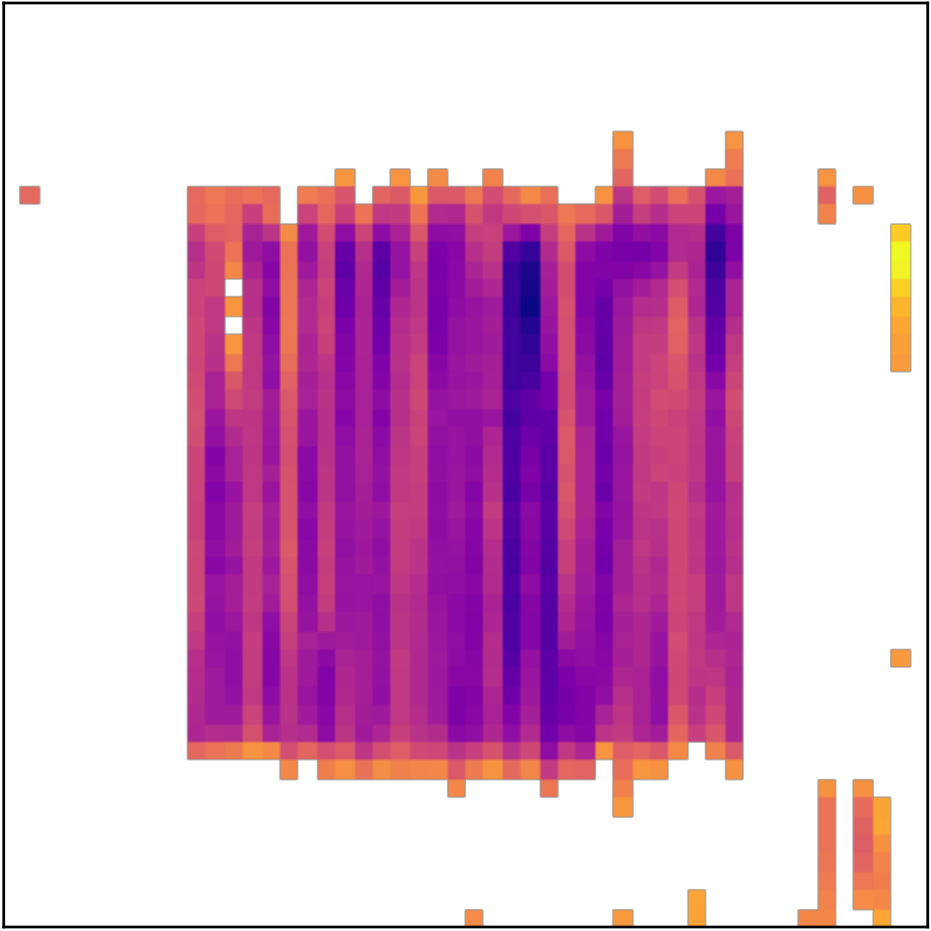}
\end{elasticrow}
\vskip\imagepaddingtiny
\begin{elasticrow}[\imagepaddingy]
    \elasticfigure{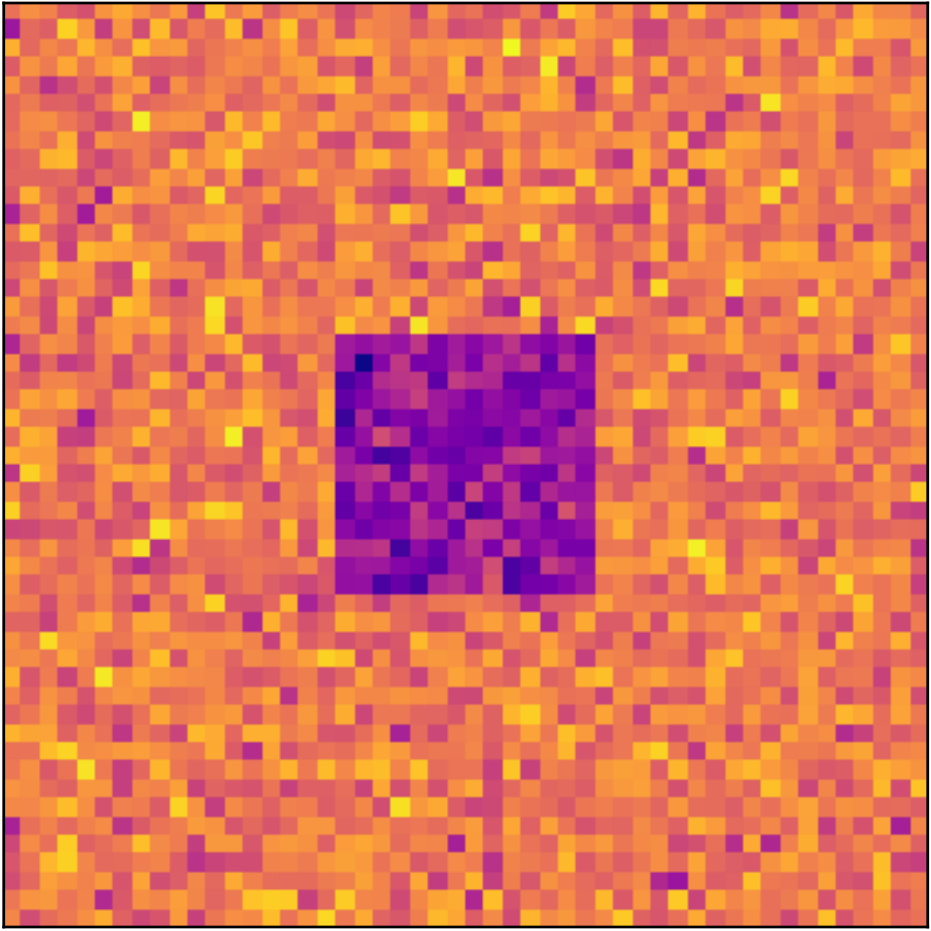}
    \elasticfigure{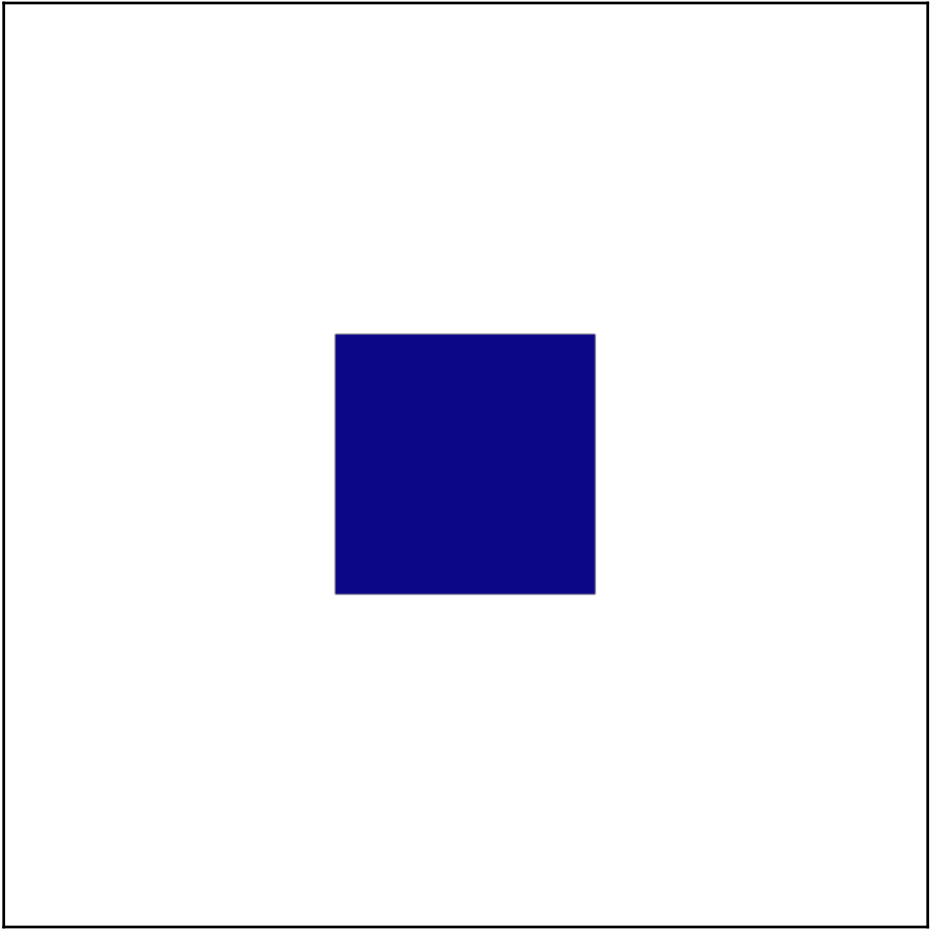}
    \elasticfigure{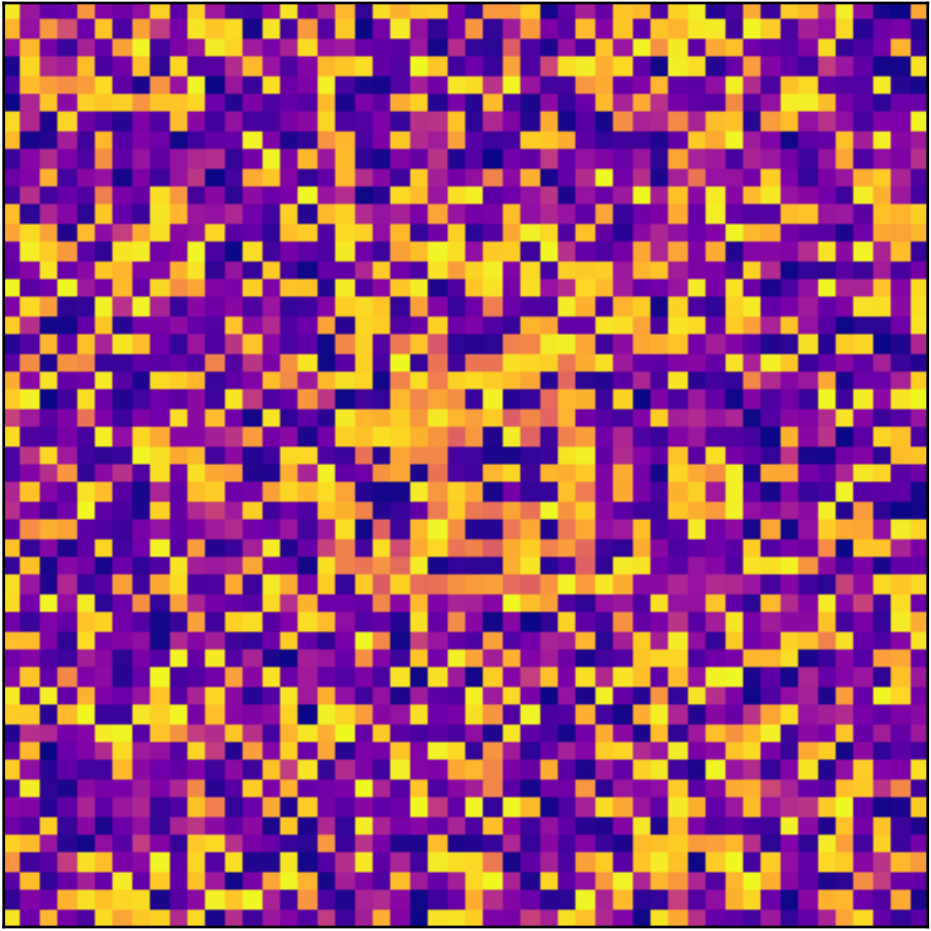}
    \elasticfigure{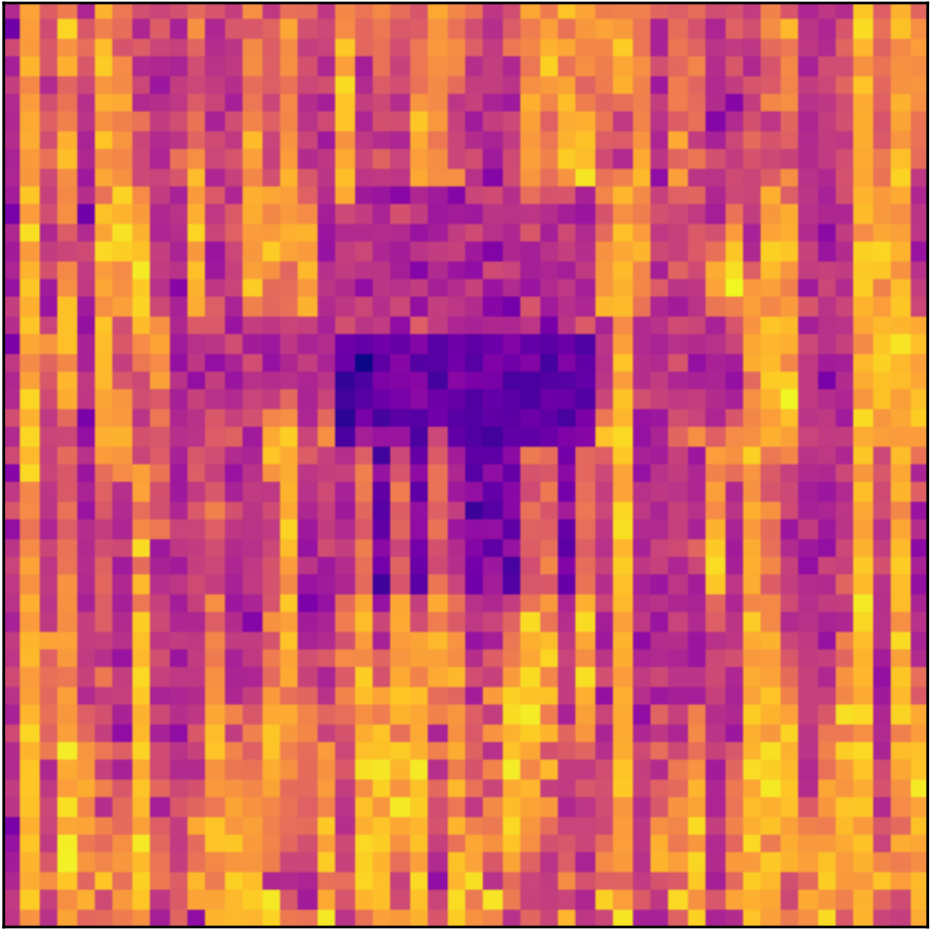}
    \elasticfigure{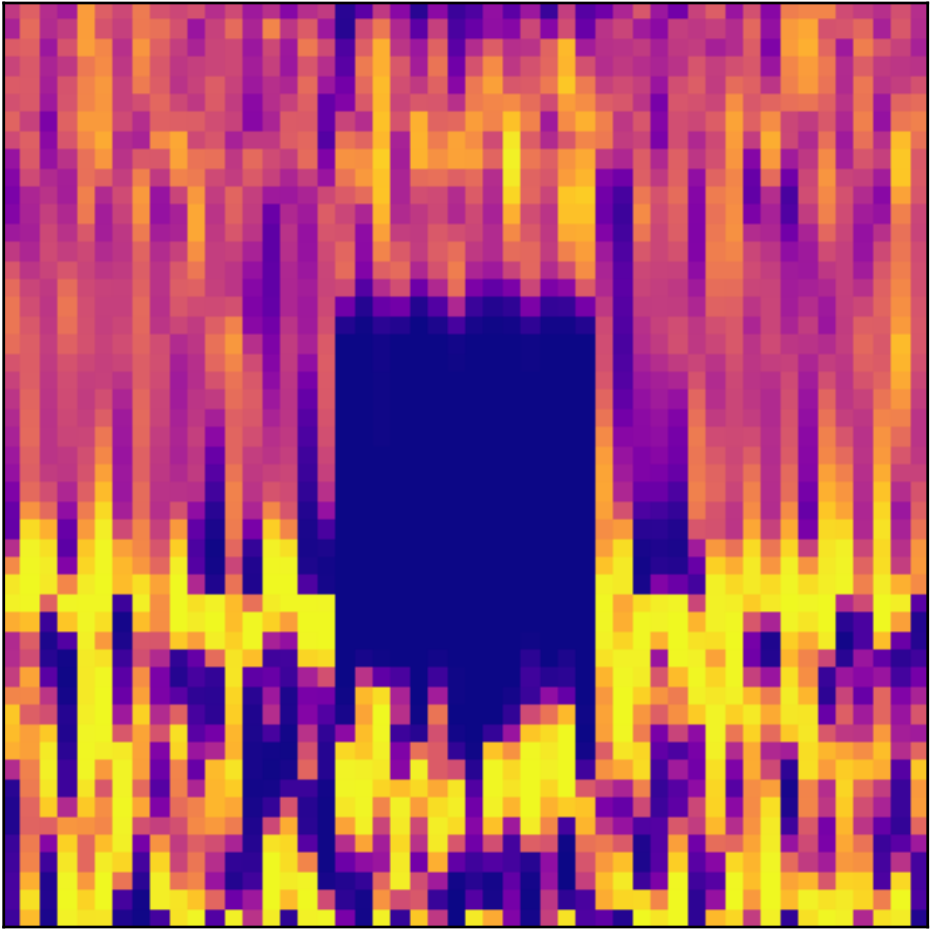}
    \elasticfigure{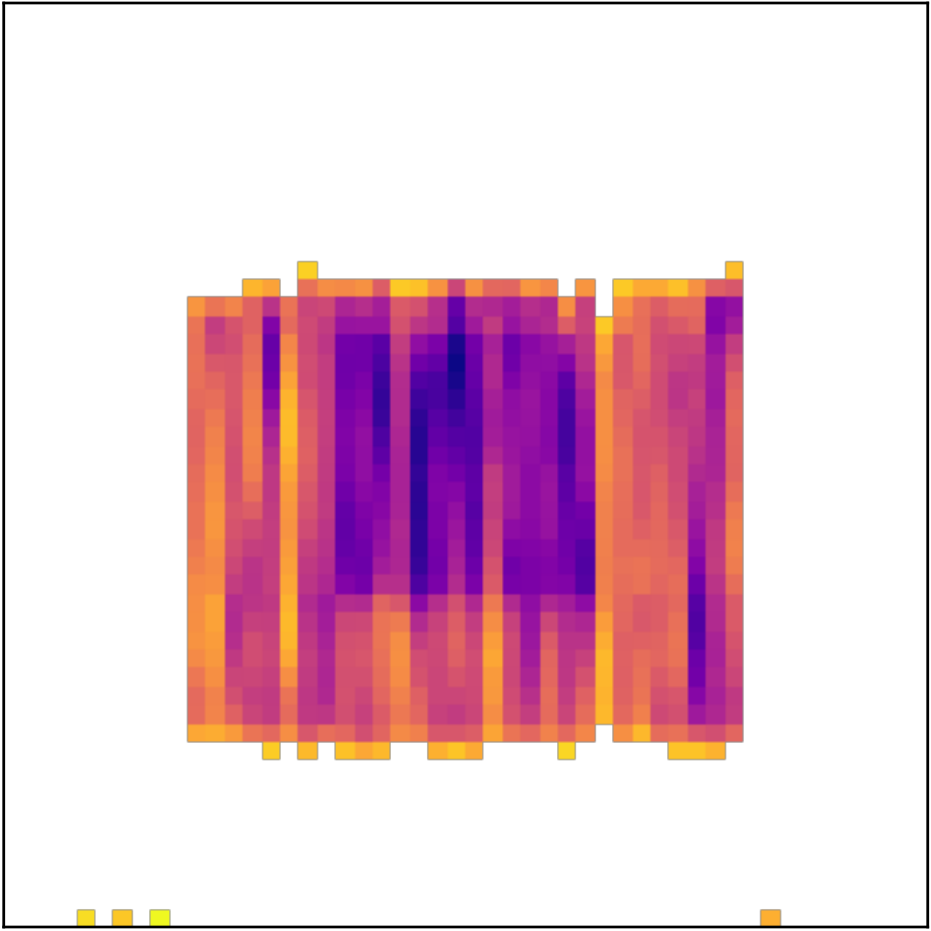} \elasticfigure{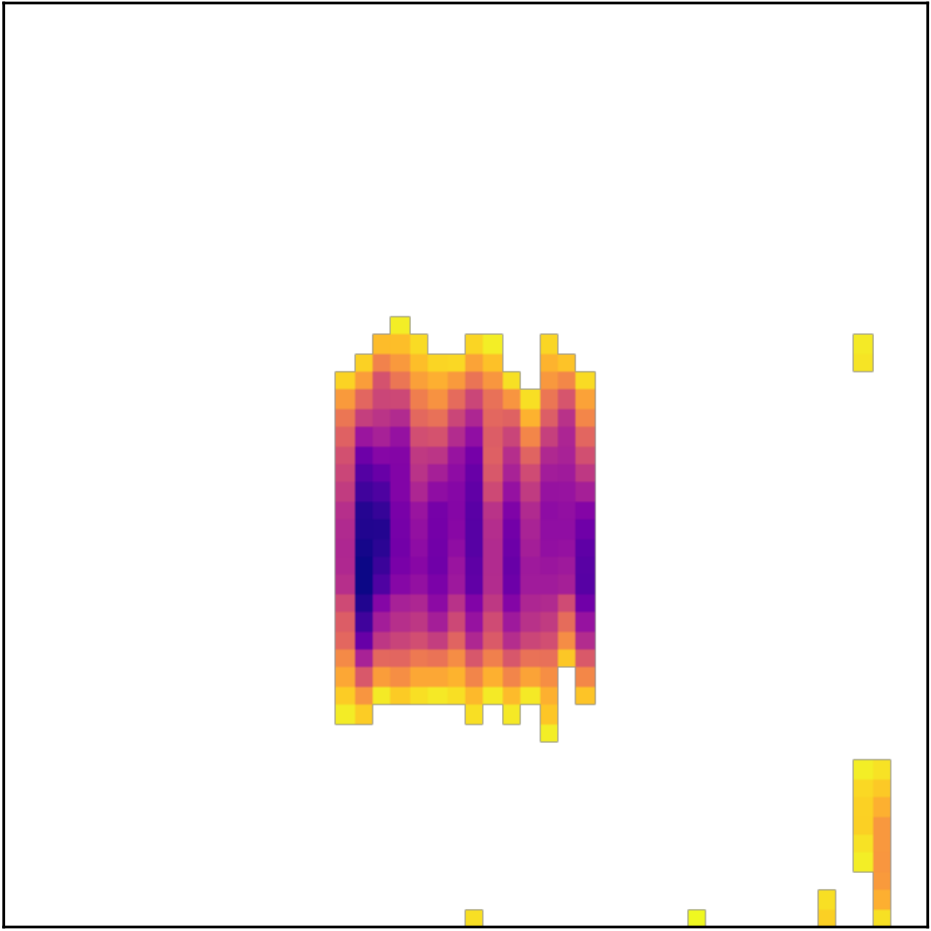}
\end{elasticrow}
\vskip\imagepaddingtiny
\begin{elasticrow}[\imagepaddingy]
    \elasticfigure{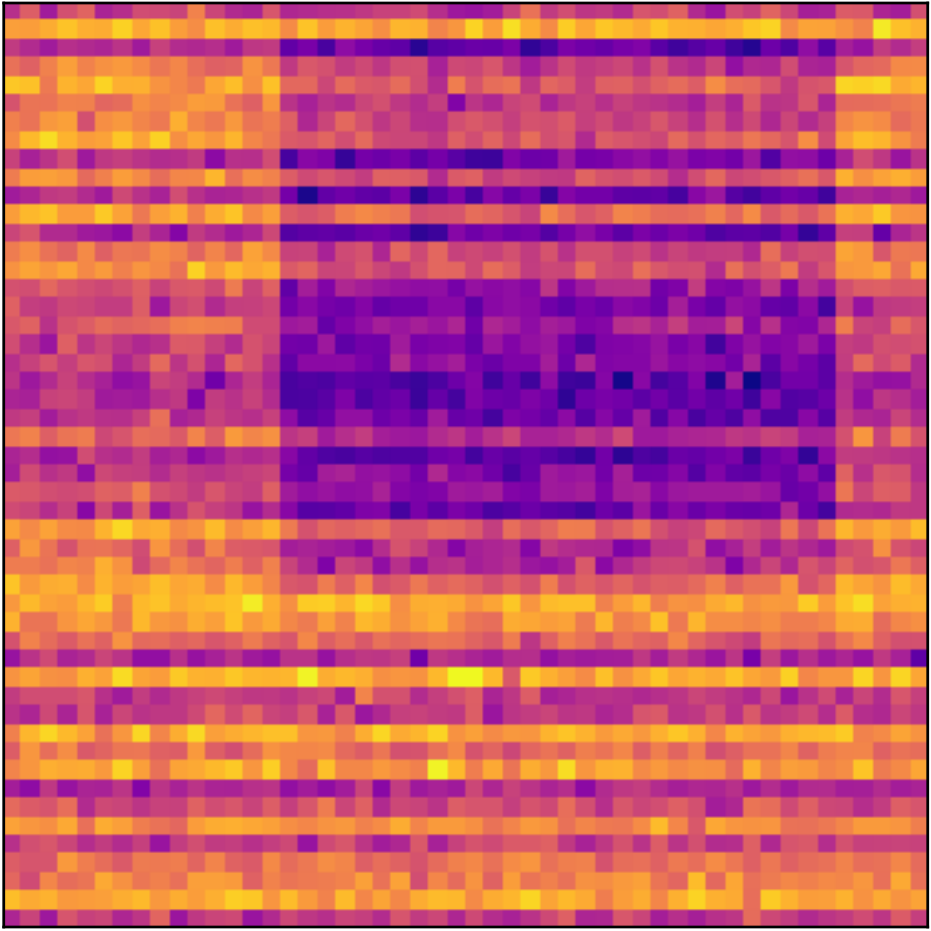}
    \elasticfigure{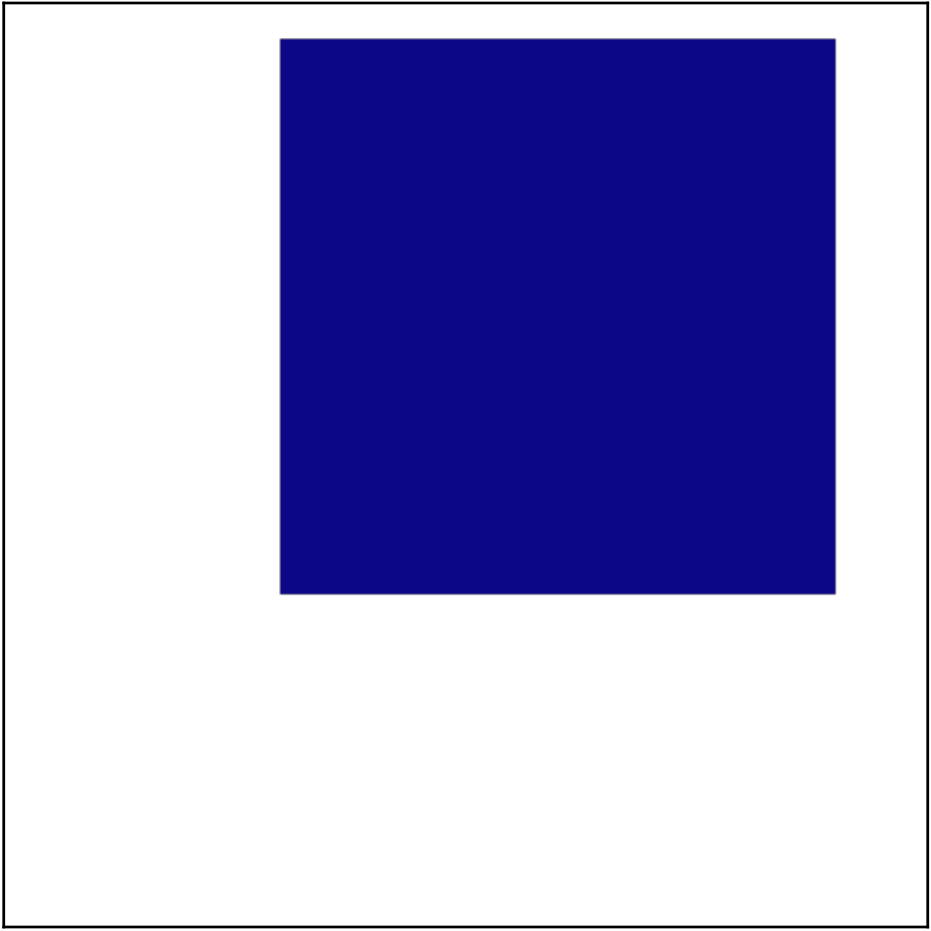}
    \elasticfigure{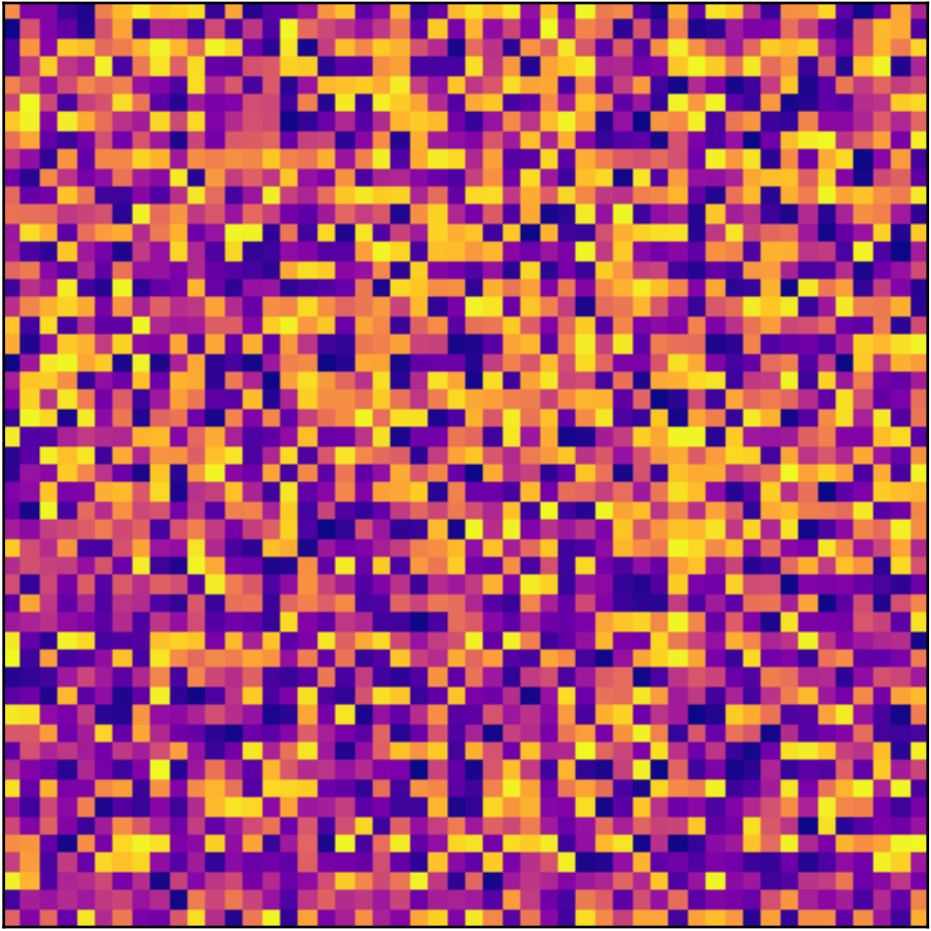}
    \elasticfigure{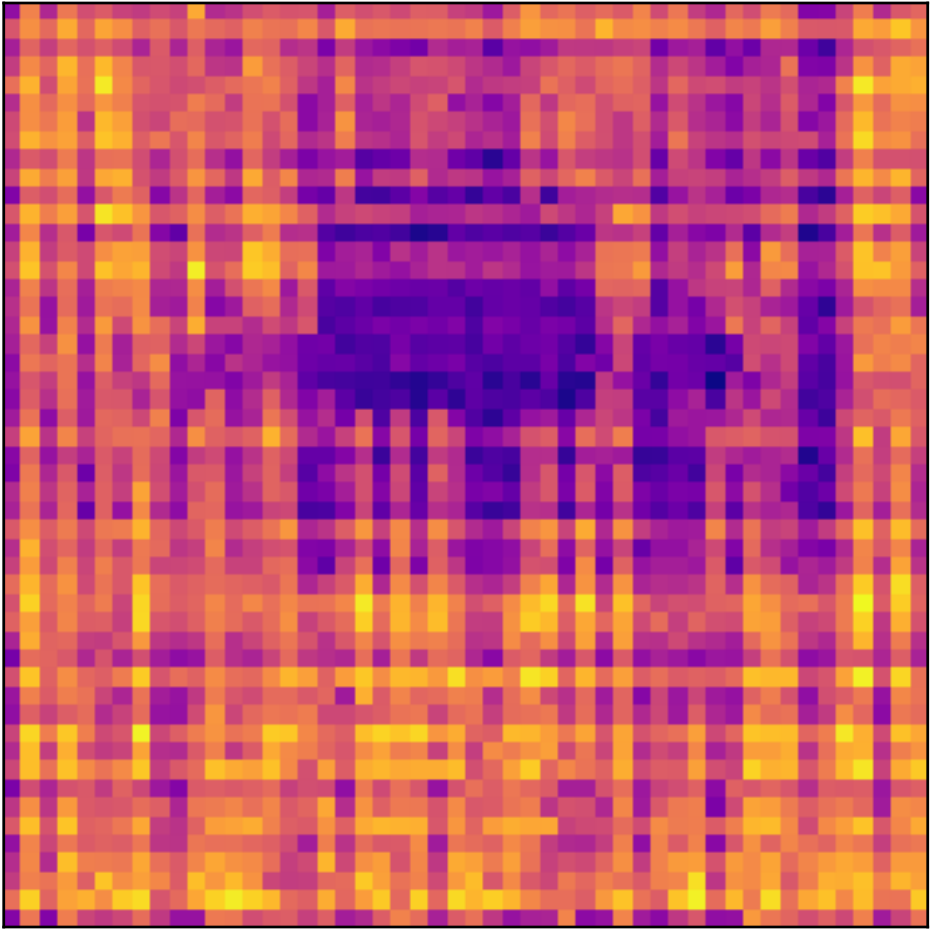}
    \elasticfigure{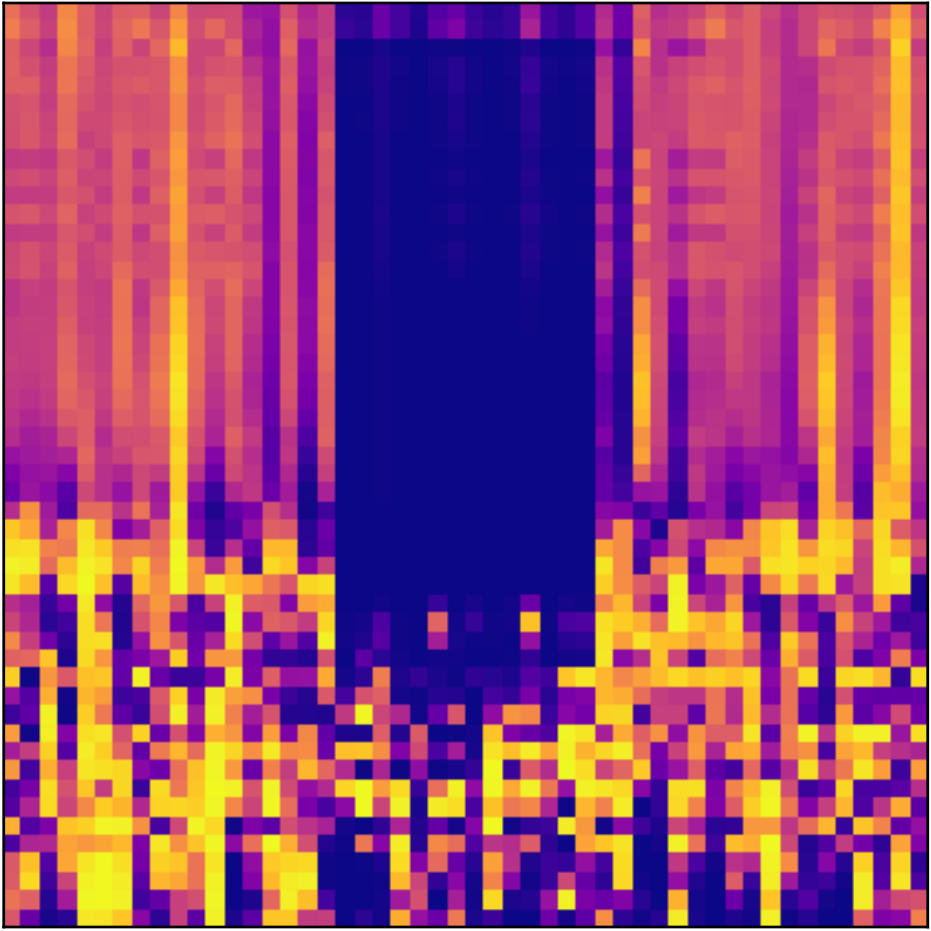}
    \elasticfigure{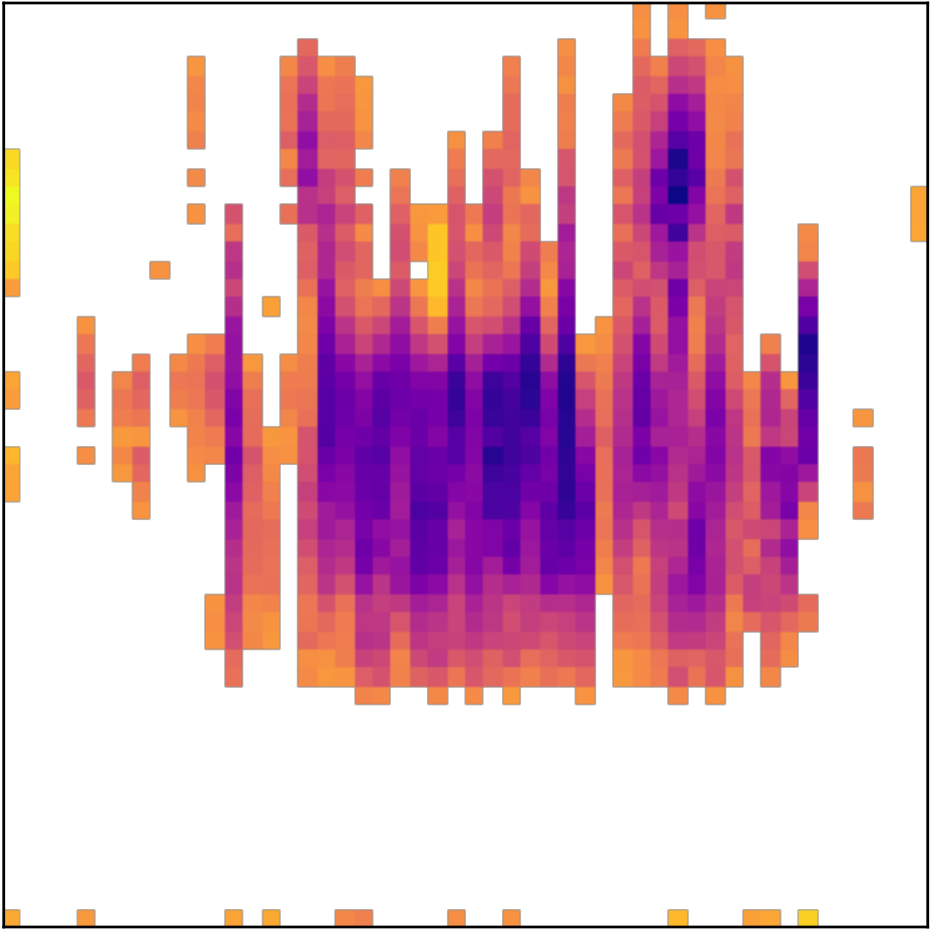}
    \elasticfigure{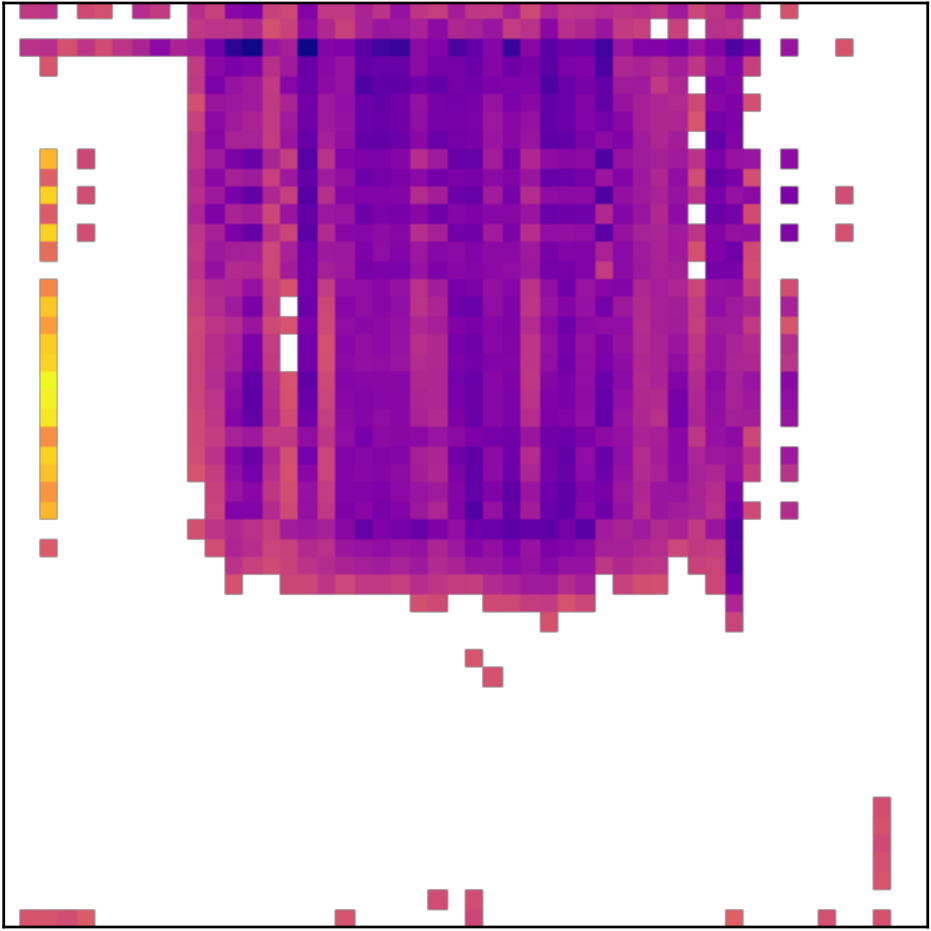}
\end{elasticrow}

\alignedlabel{0.13\linewidth}{Original}\hfill
\alignedlabel{0.13\linewidth}{Salient}\hfill
\alignedlabel{0.13\linewidth}{ICS}\hfill
\alignedlabel{0.13\linewidth}{GAN}\hfill
\alignedlabel{0.13\linewidth}{CounteRGAN}\hfill
\alignedlabel{0.13\linewidth}{\ours{} (Ours)}\hfill
\alignedlabel{0.13\linewidth}{Ours $\lambda_{4,5} = 0$}
%    Original & Salient & ICS & GAN & CounteRGAN & \ours (Ours) & Ours $\lambda_{4,5} = 0$
    \caption{Predefined salient inputs vs. counterfactual modifications for the \textit{Moving Box} dataset.}
    \label{fig:salient}
   
\end{figure}
\subsection{Saliency}

Since salient features and time steps are known upfront for the synthetic \textit{Moving Box} dataset, we compare the overlap with time steps modified by each approach. A perfect counterfactual would only modify salient inputs. We first visually compare modifications for queries with boxes of different sizes and positions (Figure \ref{fig:salient}). In all heatmaps, the x-axis represents the feature axis and time is on the y-axis. In the second column, the salient features and time steps corresponding to each query are shown in color. All remaining sub-figures demonstrate the modifications to the queries. White spaces are zero-residuals (i.e. sparse time steps and features without modifications). Darker colors indicate stronger modifications.

ICS largely fails to identify salient points in the input. All GAN-based methods detect the position of most salient inputs. However, GAN and CounteRGAN additionally modify non-salient inputs. In contrast, SPARCE modifies far fewer inputs overall and focuses on salient inputs. For this dataset, the performance of our approach can be further improved by switching off $\lambda_{4,5}$, i.e. sparsity and jerk regularization. This shows that sparsity is primarily induced by the sparsity layer. It also demonstrates that the application of the jerk constraint depends on the problem. Here, a clear value increase marks the transition from non-salient to salient inputs. In human motion datasets, in contrast, smooth movements are natural and desired.

\begin{figure}
\centering
%{\setlength{\tabcolsep}{0.1cm}\renewcommand{\arraystretch}{1.5} \scriptsize
%\begin{tabularx}{\linewidth}{@{}CC@{}} %{@{}ccccccc@{}}
\alignedlabel{0.35\linewidth}{Target Class 1}
\alignedlabel{0.35\linewidth}{Target Class 0}\\
\includegraphics[width=0.35\linewidth]{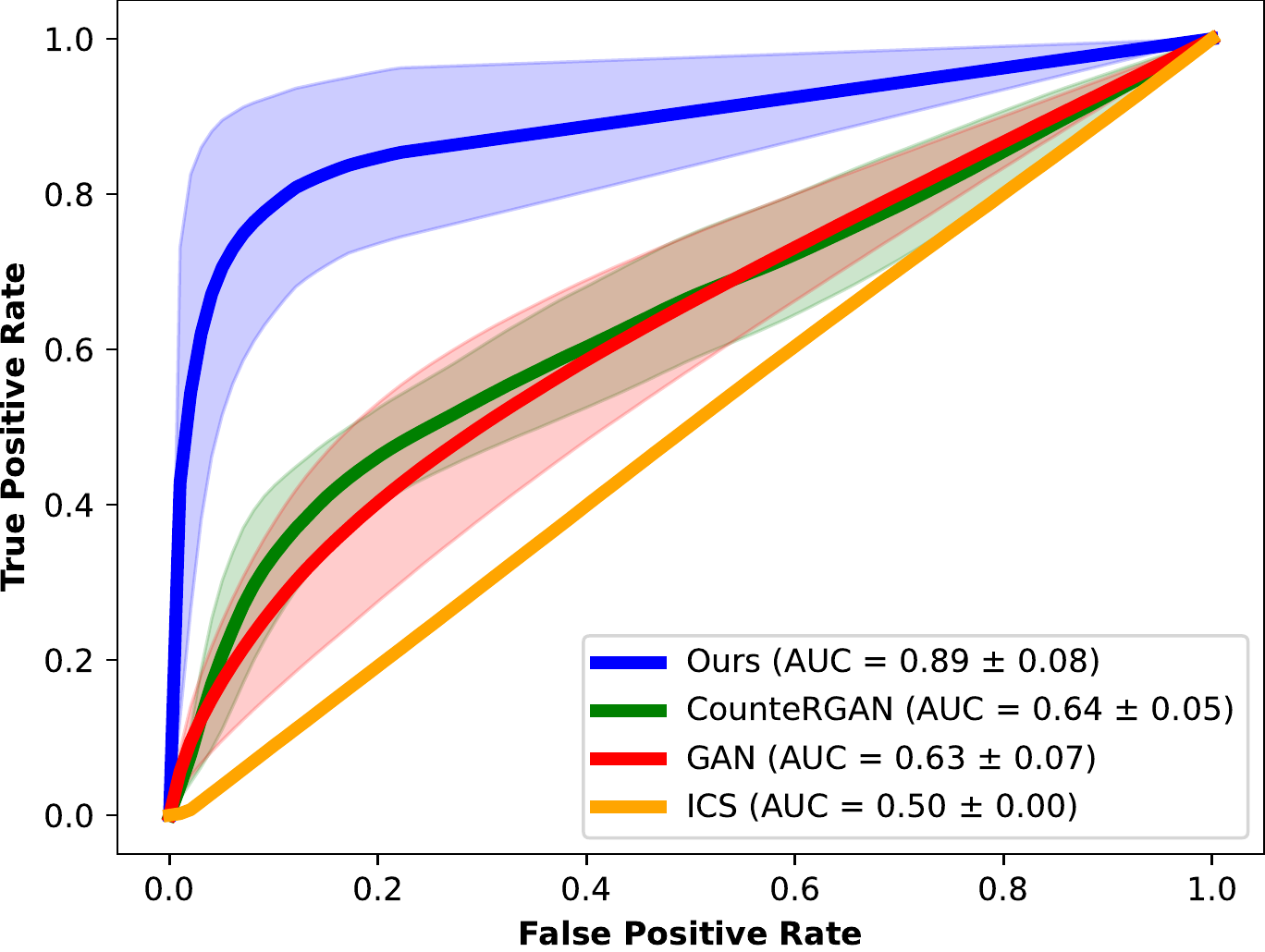} \includegraphics[width=0.35\linewidth]{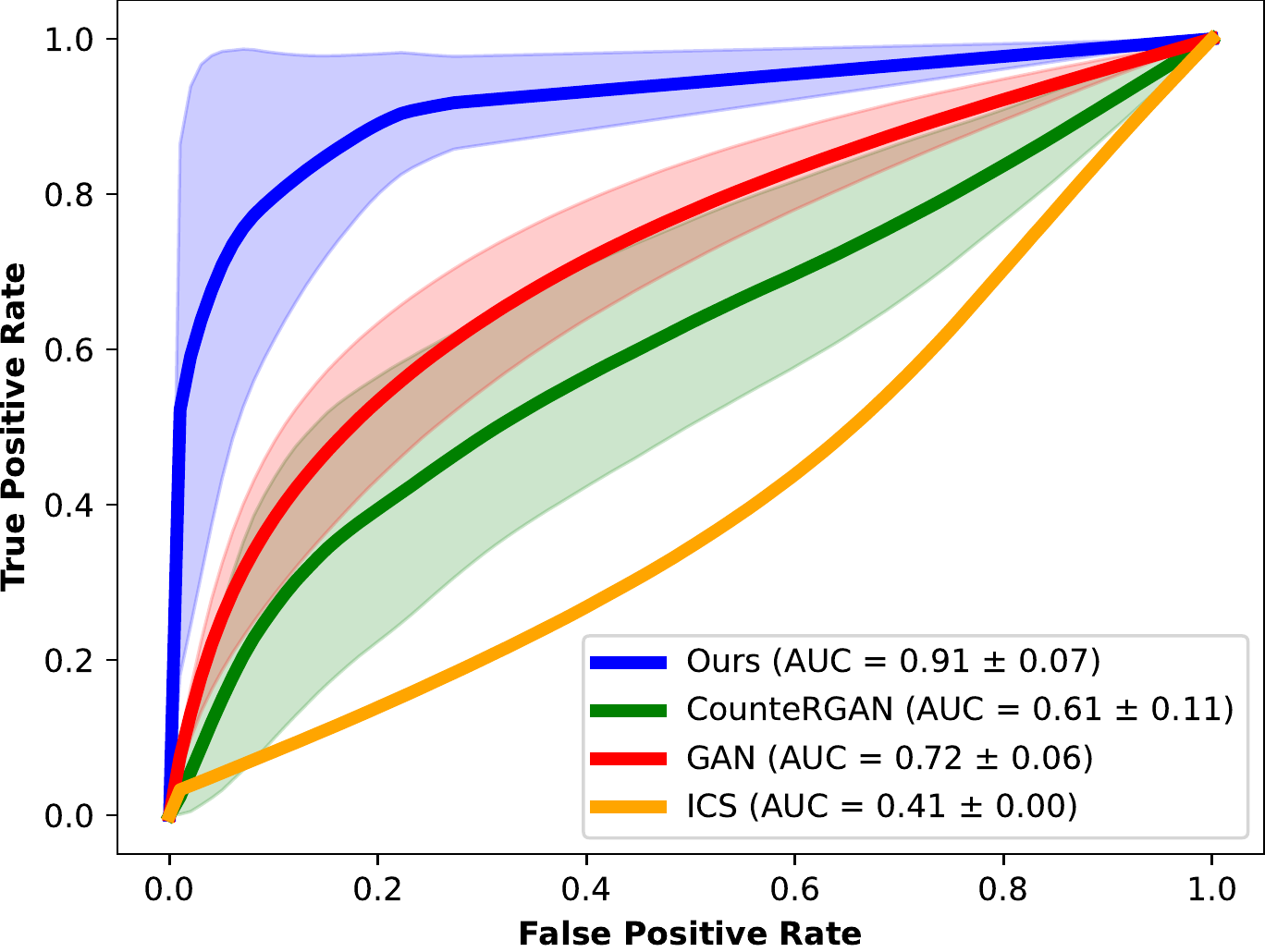} \\%   
     %Target Class 1 & Target Class 0 
%\end{tabularx}}
    \caption{Mean ROC curve measuring the overlap between predefined salient inputs and counterfactual modifications for the \textit{Moving Box} dataset. Shaded areas describe one standard deviation of the mean ROC.}
    \label{fig:auc}
   
\end{figure}

We furthermore assess the salience overlap in a quantitative manner via the receiver operating characteristic (ROC) curve in combination with the area under the curve (AUC) score. Higher AUC scores indicate better discrimination performance between salient and non-salient inputs. Figure \ref{fig:auc} visualizes mean ROC curves over five repetitions for both target classes. In both cases, we see that our approach produces counterfactuals that show a substantially higher overlap with predefined salient inputs than other approaches. Visual and quantitative evaluation therefore demonstrates that our approach creates sparse counterfactual explanations and is also suitable for the identification of salient inputs in multivariate time series.

\begin{figure}
    \centering
    \alignedlabel{0.35\linewidth}{Shoulder-Elbow Distance}
    \alignedlabel{0.35\linewidth}{Elbow-Wrist Distance}\\
    %\begin{subfigure}{0.49\textwidth}
    %    \centering
    \includegraphics[width=0.35\linewidth]{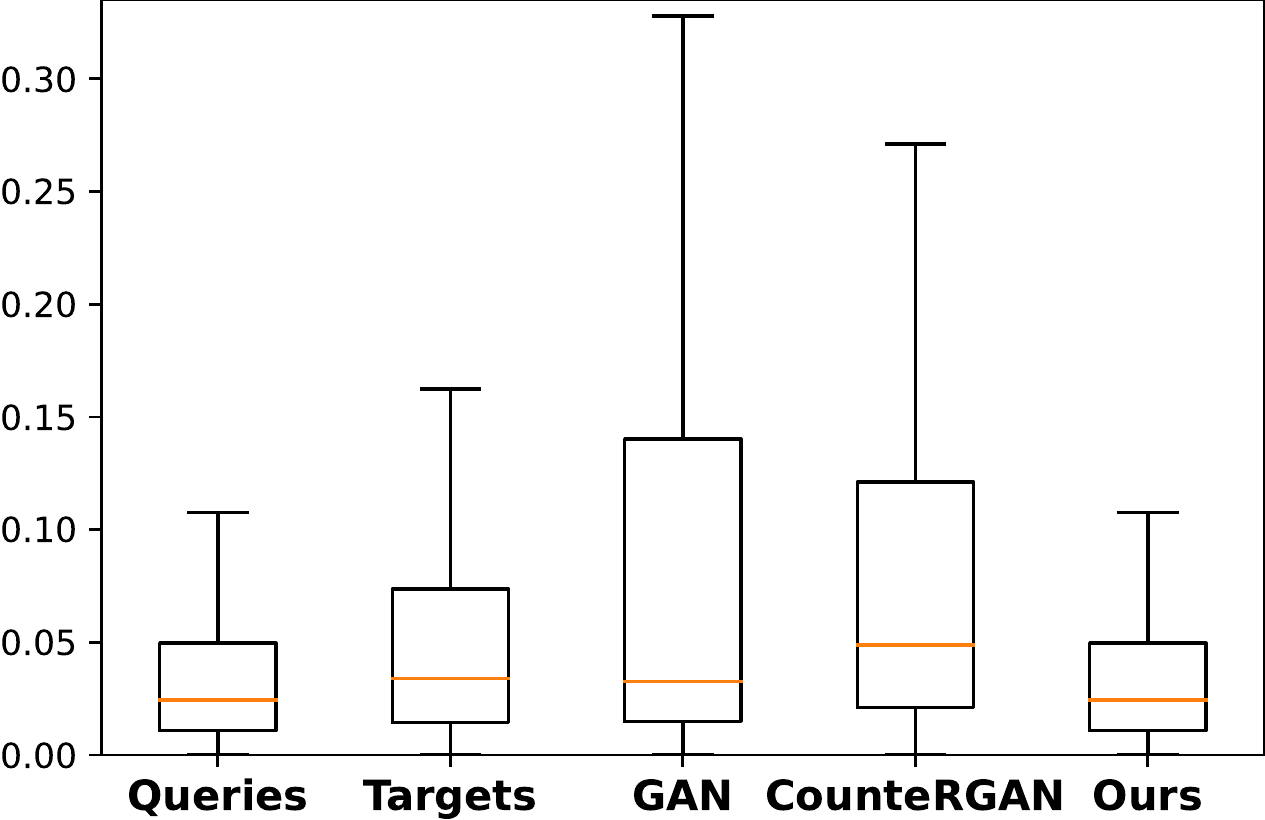}
%        \caption{Shoulder-Elbow Distance}
%    \end{subfigure}
%    \begin{subfigure}{0.49\linewidth}
%        \centering
    \includegraphics[width=0.35\linewidth]{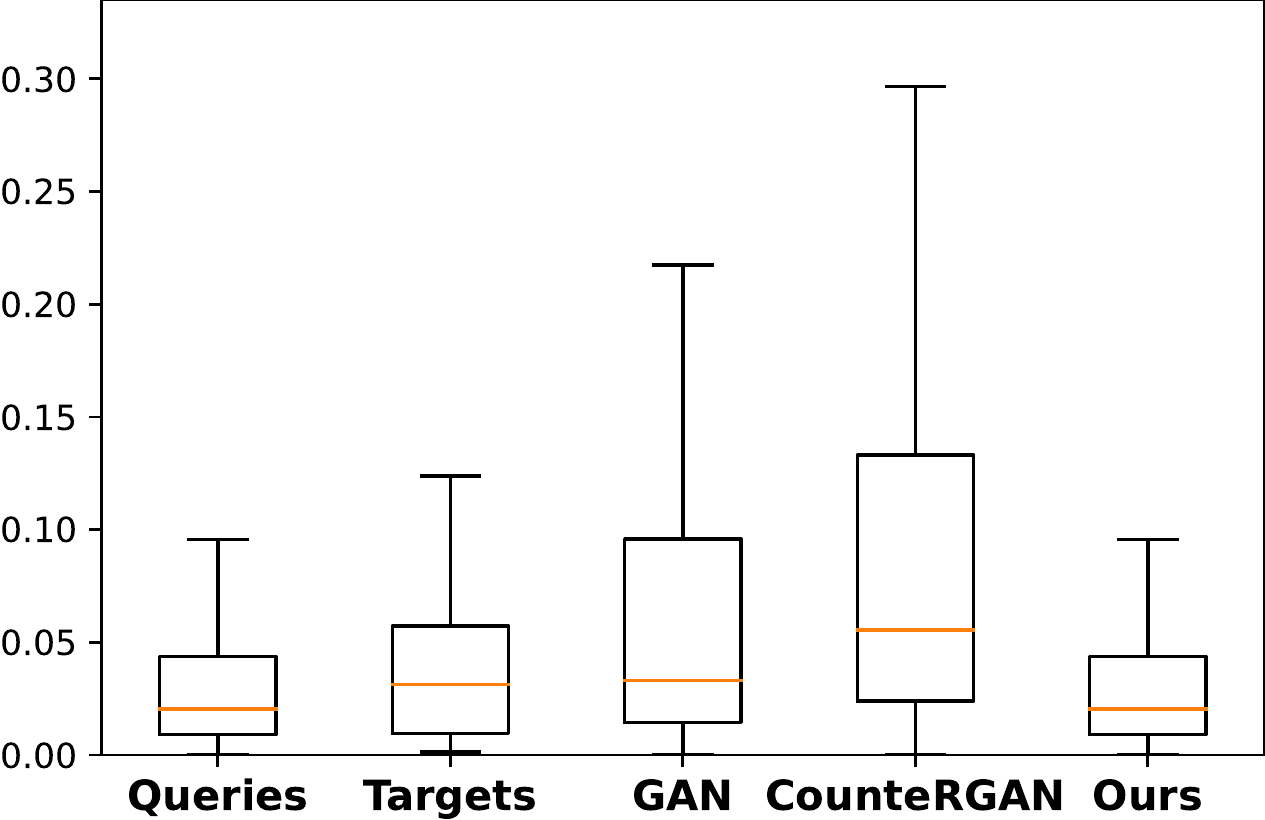}
%        \caption{Elbow-Wrist Distance}
%    \end{subfigure}
    \caption{Measured distances between body parts for originals and counterfactuals for the \textit{Catching} dataset. Orange horizontal lines denote the median distance. Boxes contain all values between lower and upper quartiles.}
    \label{fig:boxplots}
\end{figure}

\subsection{Geometric Plausibility}
To assess geometric plausibility of the \textit{Catching} dataset, we compute the Euclidean distances between body parts for the original dataset and the generated counterfactuals (Figure \ref{fig:boxplots}). ICS is excluded from the figure, since the corresponding values lie outside of the displayed area. Counterfactuals generated by our approach most closely resemble the body-part distances found in the original data. This can indicate higher geometric plausibility of our generated counterfactuals. In order to fully inspect geometric plausibility, however, the angles at which the joints are positioned in relation to one another would also have to be examined.

\section{Conclusions}
\label{sec:conclusions}
We proposed GAN architecture for generating time- and feature-sparse counterfactual explanations for multivariate time series. Our approach extends previous methods by a custom sparsity layer and additional loss regularization for sparsity and smoothness. In extensive experiments, we demonstrate that in spite of making substantially sparser modifications \ours{} achieves comparable or superior performance on common metrics for counterfactual search. Benchmarking our approach on a synthetic interpretability dataset, we show that it can also be used for feature attribution. The application to real-world human motion datasets demonstrates that our approach generates sparser and more plausible counterfactuals than related approaches.

The design of our approach allows for a flexible change of the desired target class, as well as an easy adaptation of the counterfactual value function catering to the needs of other applications. Future extensions can consider other applications (e.g. weather, stocks) and domain-specific regularization terms. In critical sectors such as healthcare, misinterpretation of systems can have severe consequences. XAI systems should thus always be validated by human experts. To enhance the understandability of generated counterfactuals, further work can investigate the visualization of explanations for end-users (e.g. in textual or visual form).

%Furthermore, an analysis of the correlation between modifications of different features can guide the understanding of feature interaction in a dataset.

%In conclusion, we believe that our approach can be of great value for interpreting decisions of neural networks trained on multivariate time series classification.

\begin{ack}
The authors thank the International Max Planck Research School for Intelligent Systems (IMPRS-IS) and the German Academic Scholarship Foundation (Studienstiftung des deutschen Volkes) for supporting Jana Lang. Martin Giese received support from the BMG project SStepKiZ and the EU ERC SYNERGY Grant RELEVANCE.
\end{ack}

%\section*{References}
%\medskip

\small
\bibliographystyle{abbrvnat}
\bibliography{bibliography}

%%%%%%%%%%%%%%%%%%%%%%%%%%%%%%%%%%%%%%%%%%%%%%%%%%%%%%%%%%%%

%%%%%%%%%%%%%%%%%%%%%%%%%%%%%%%%%%%%%%%%%%%%%%%%%%%%%%%%%%%%

\clearpage

\appendix

\section{Appendix}

\subsection{Additional Results}

\subsubsection{Saliency Heatmaps}
\begin{figure}[h!]
\centering
%\resizebox{1.0\linewidth}{!}{%
%.135
\begin{elasticrow}[\imagepaddingy]
    \elasticfigure{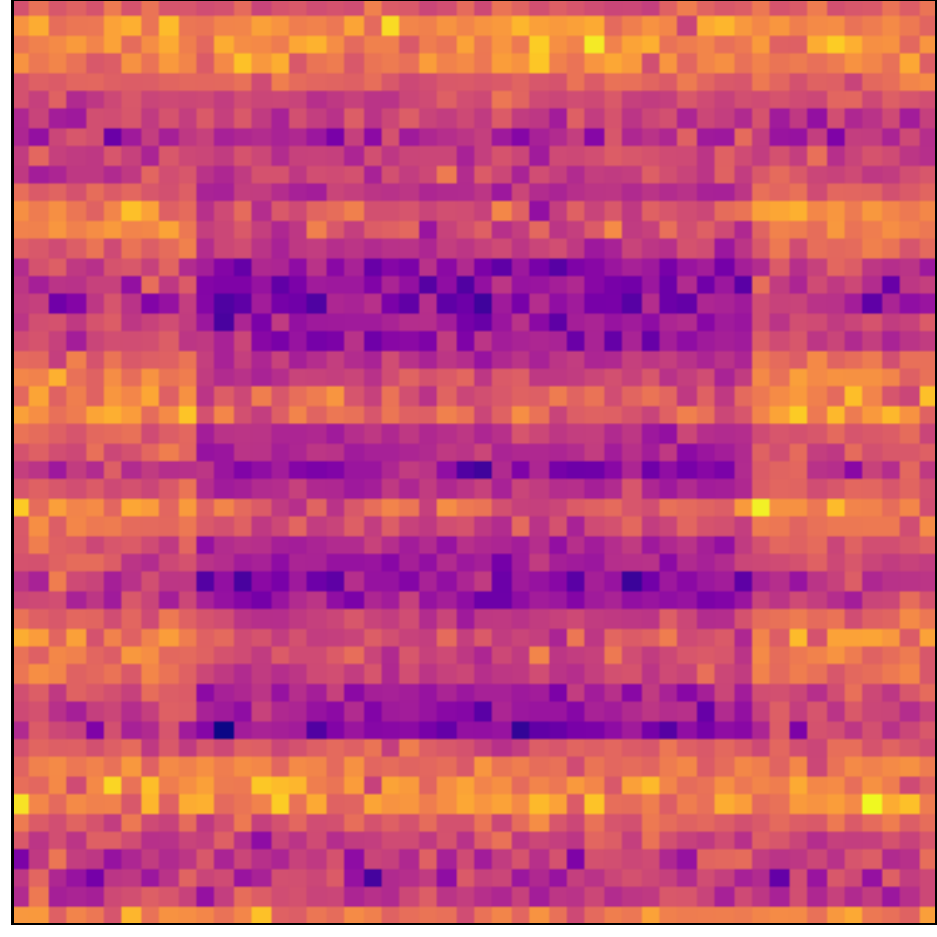}
    \elasticfigure{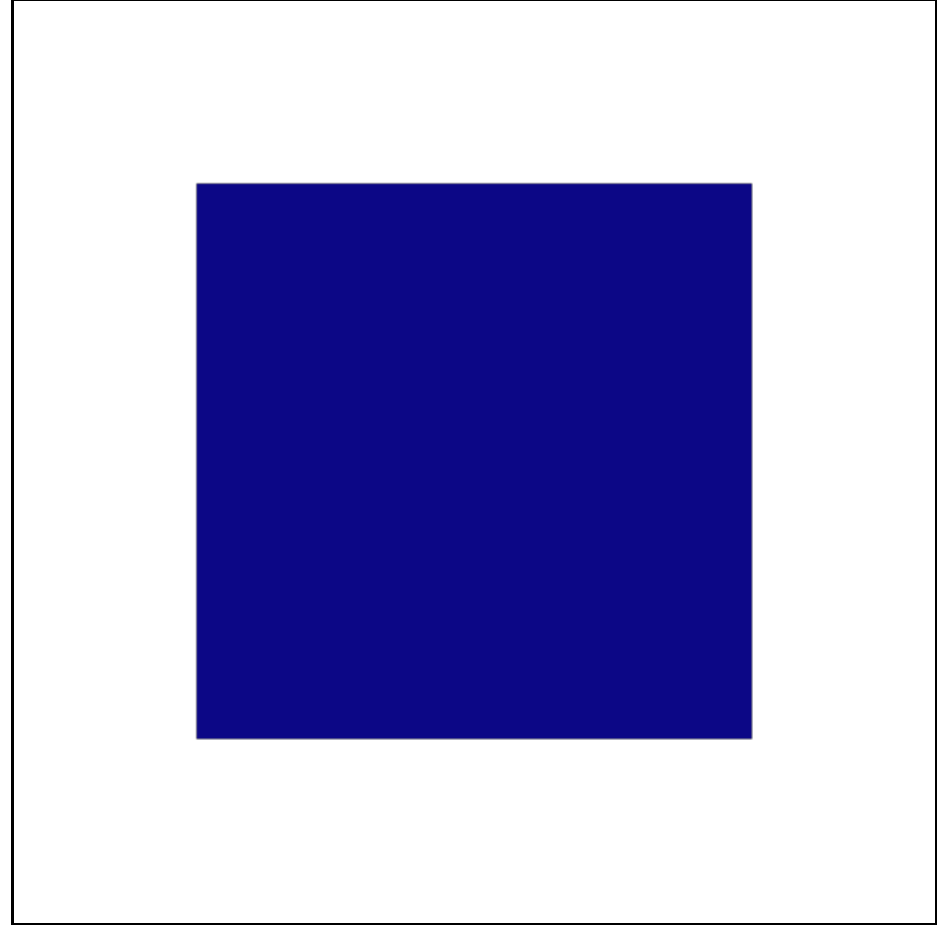}
    \elasticfigure{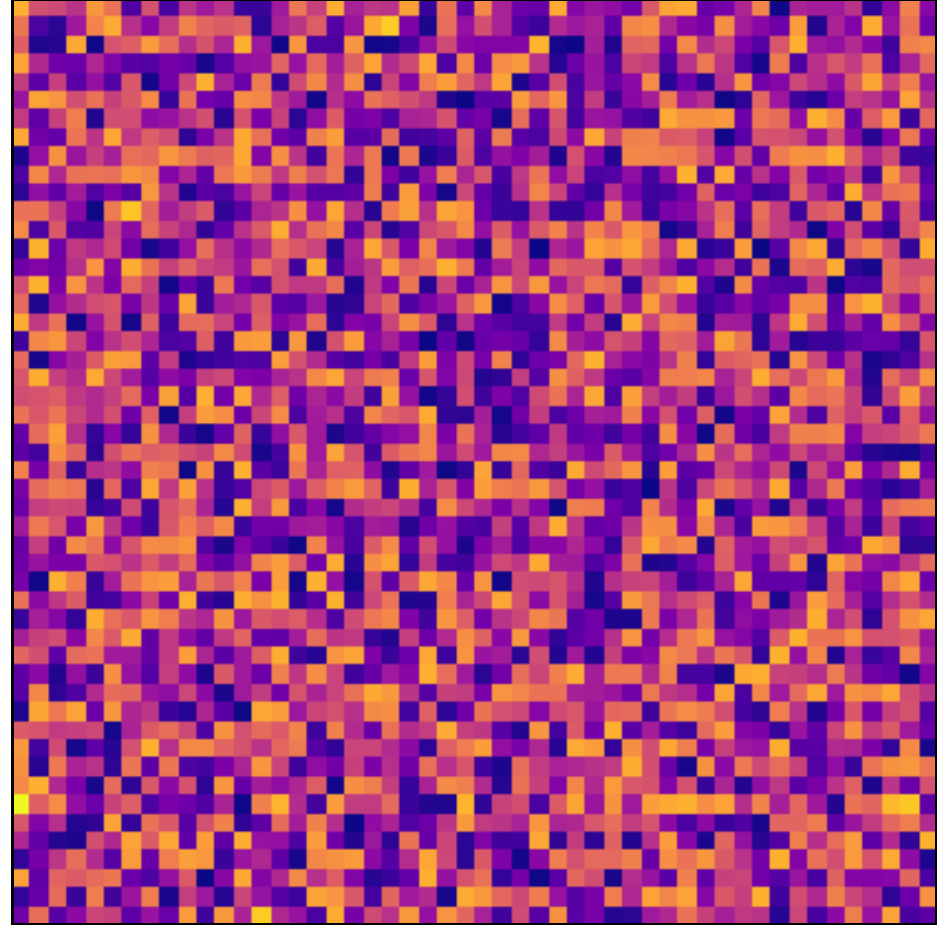}
    \elasticfigure{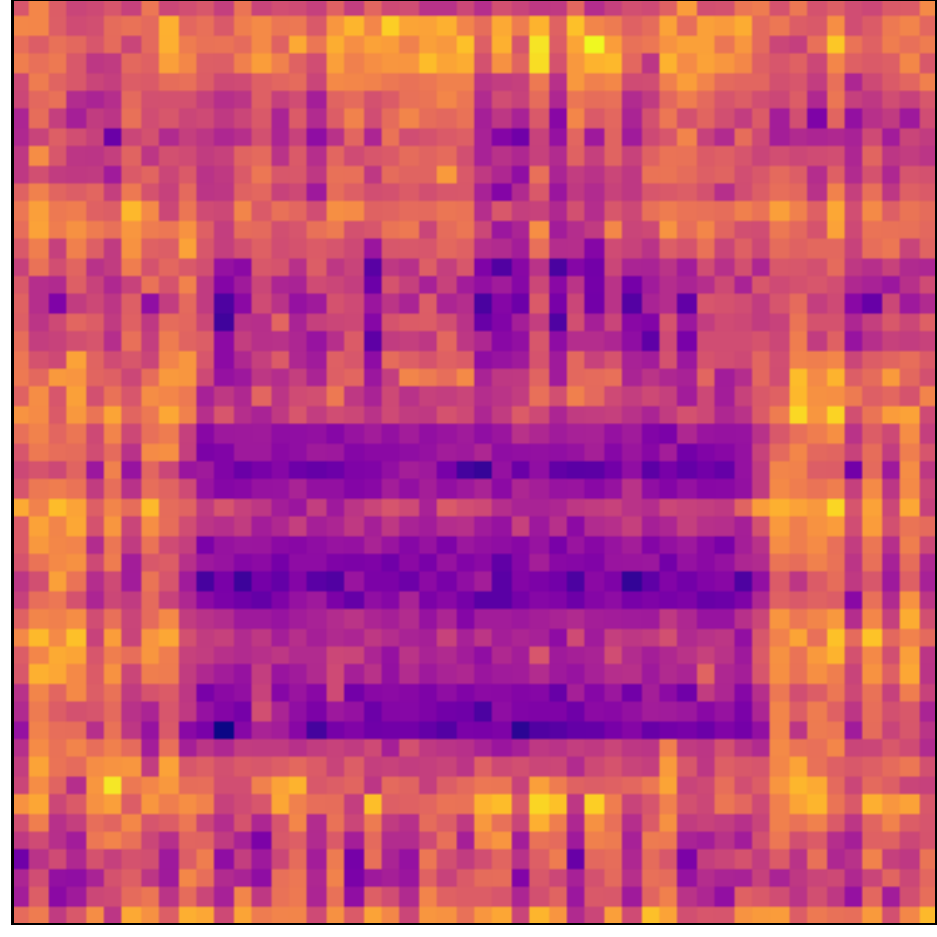}
    \elasticfigure{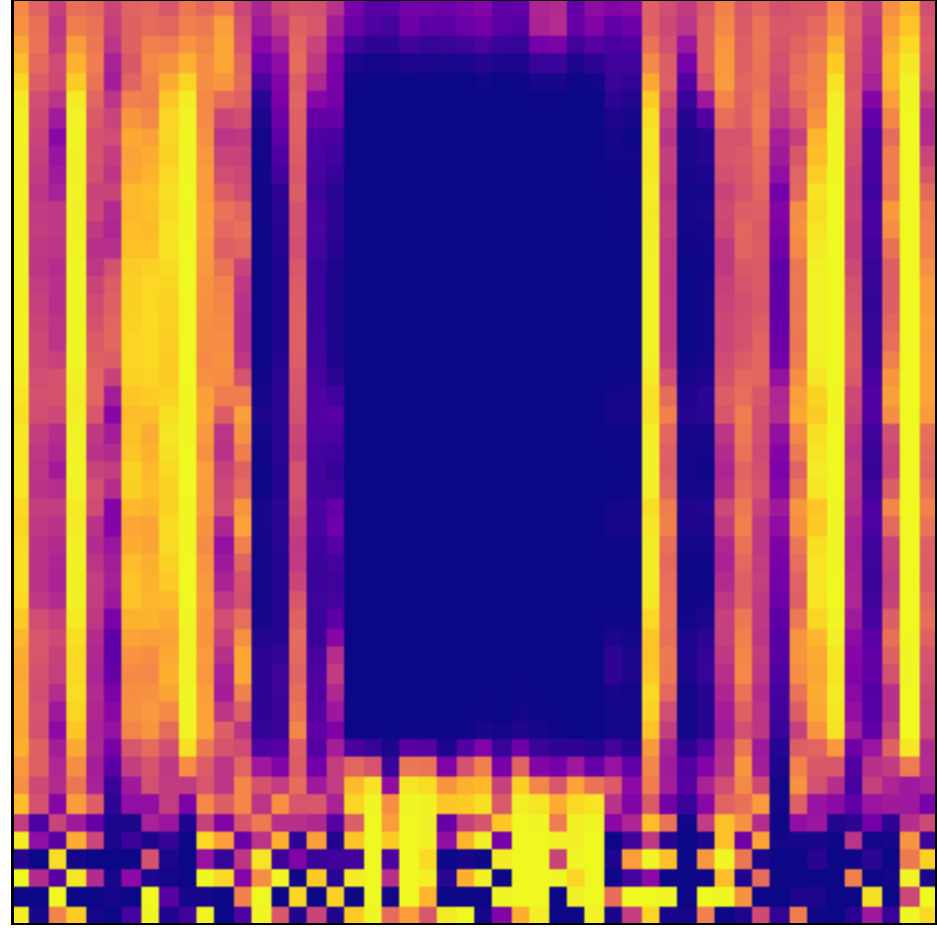}
    \elasticfigure{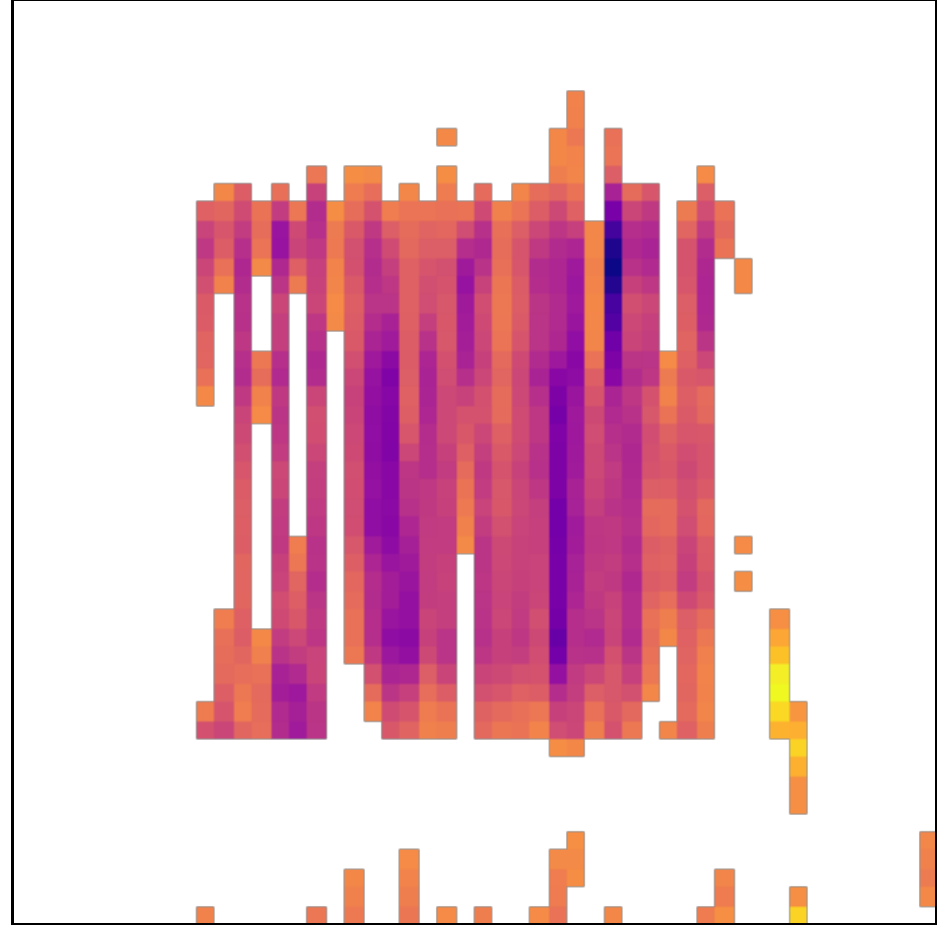}
\end{elasticrow}
\vskip\imagepaddingtiny
\begin{elasticrow}[\imagepaddingy]
    \elasticfigure{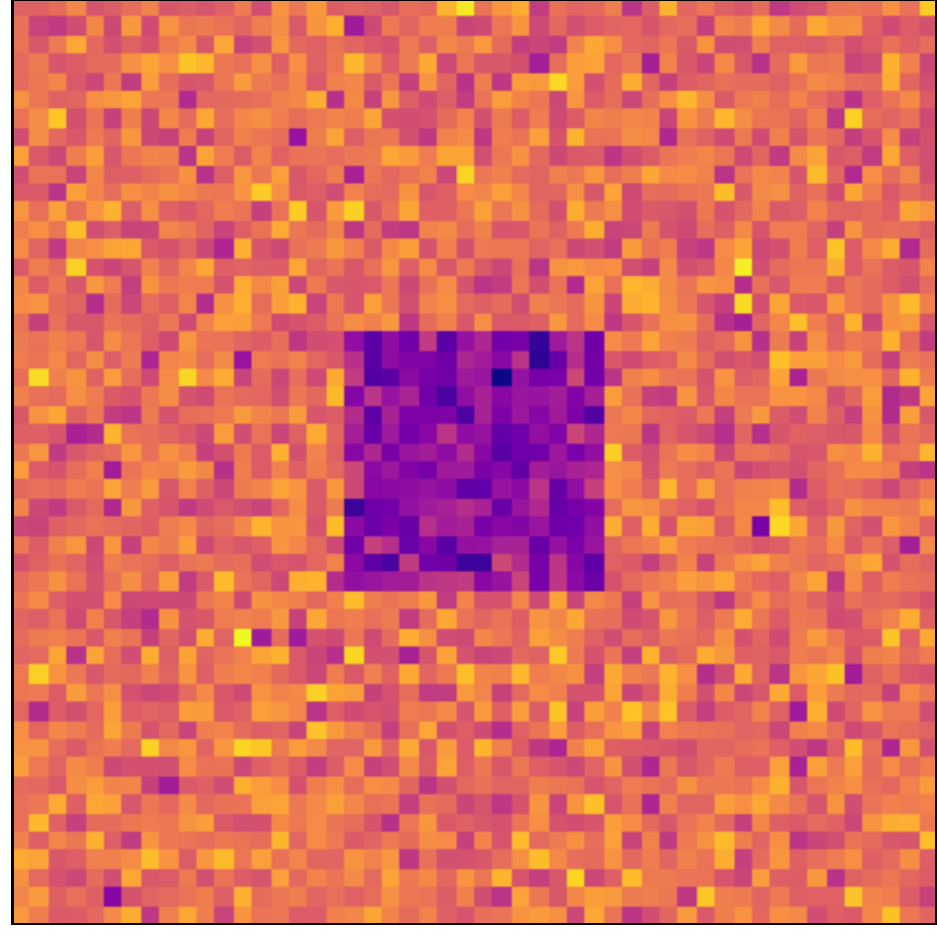}
    \elasticfigure{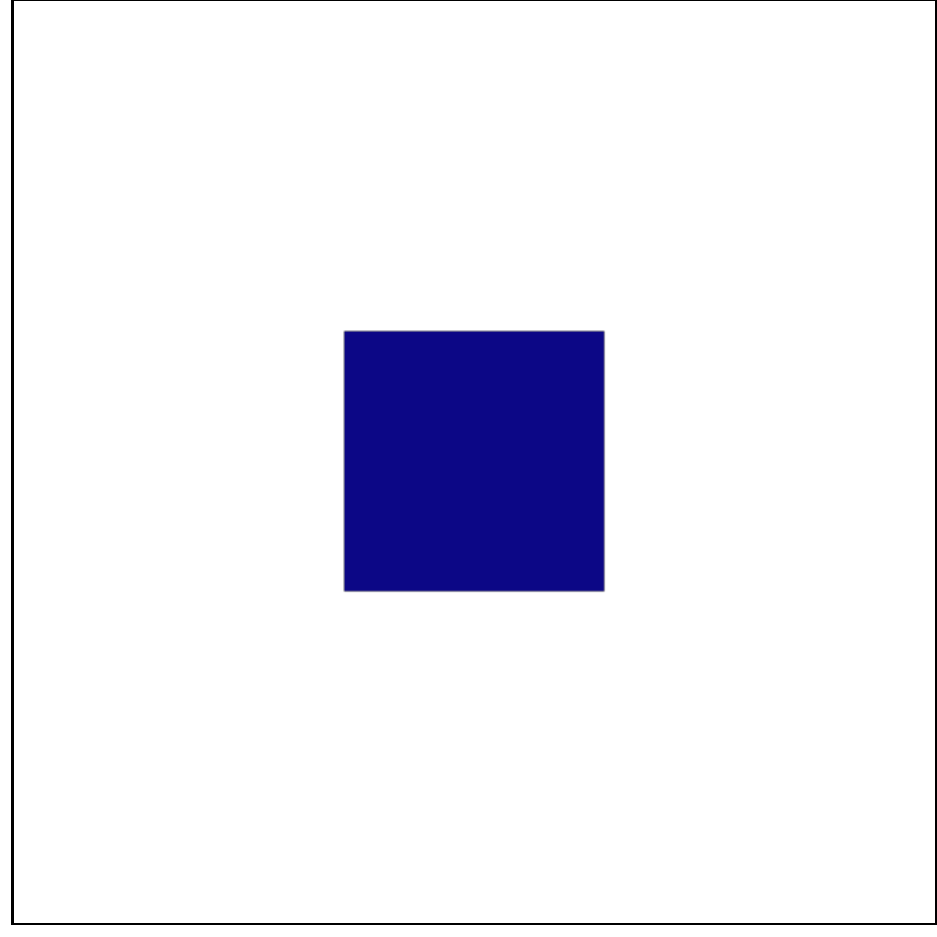}
    \elasticfigure{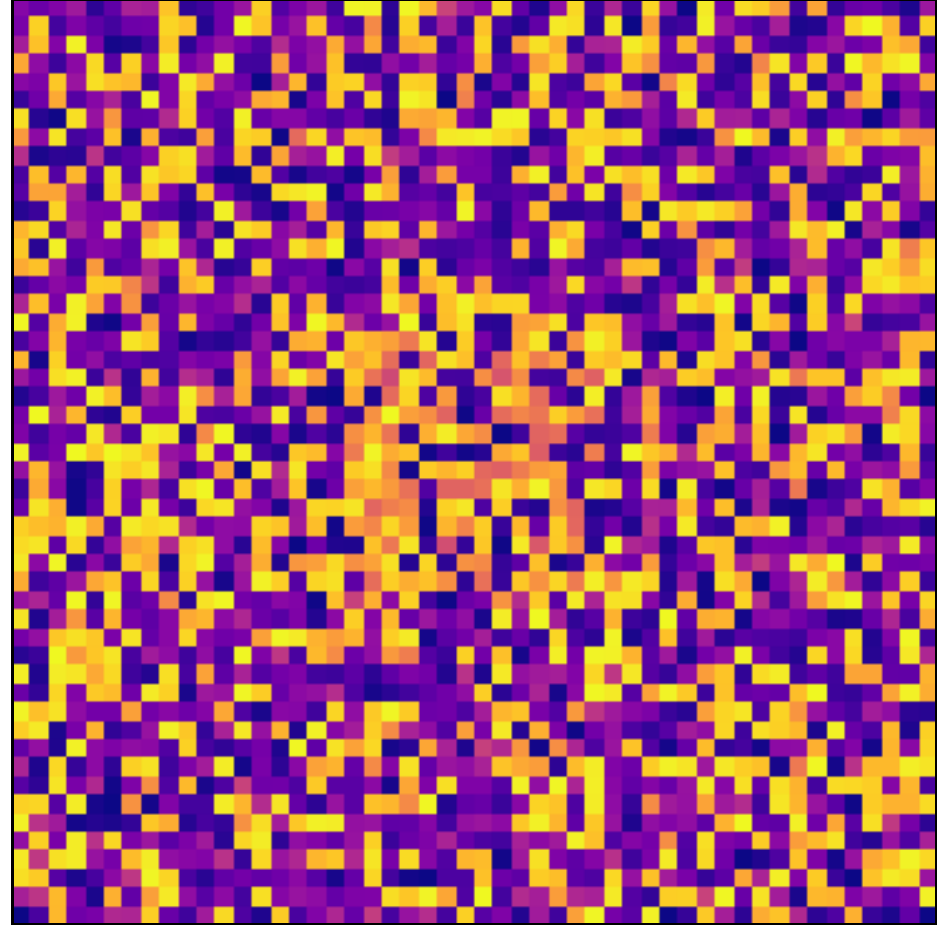}
    \elasticfigure{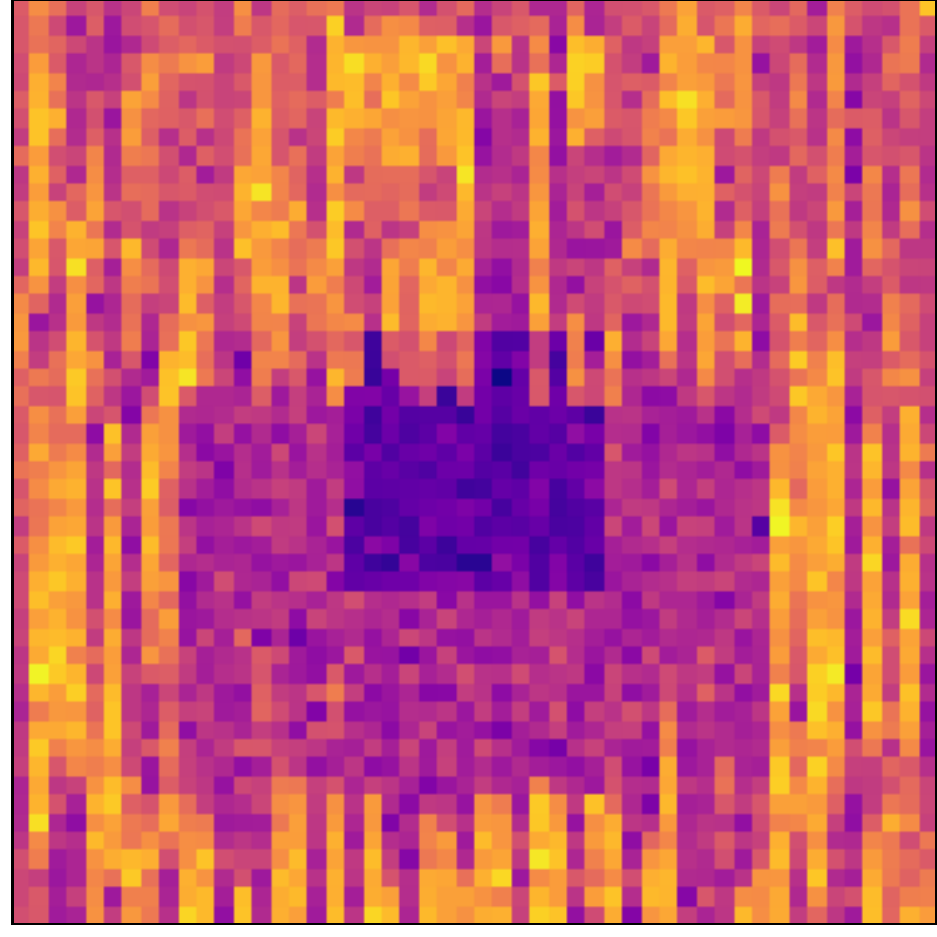}
    \elasticfigure{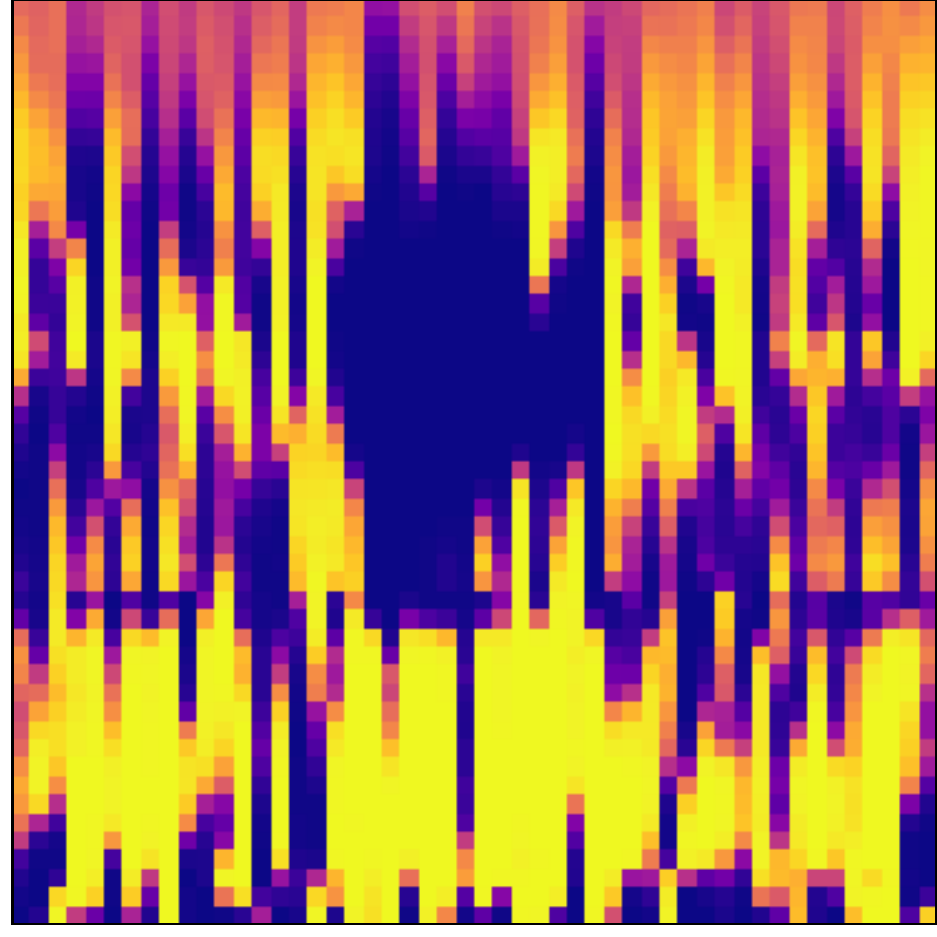}
    \elasticfigure{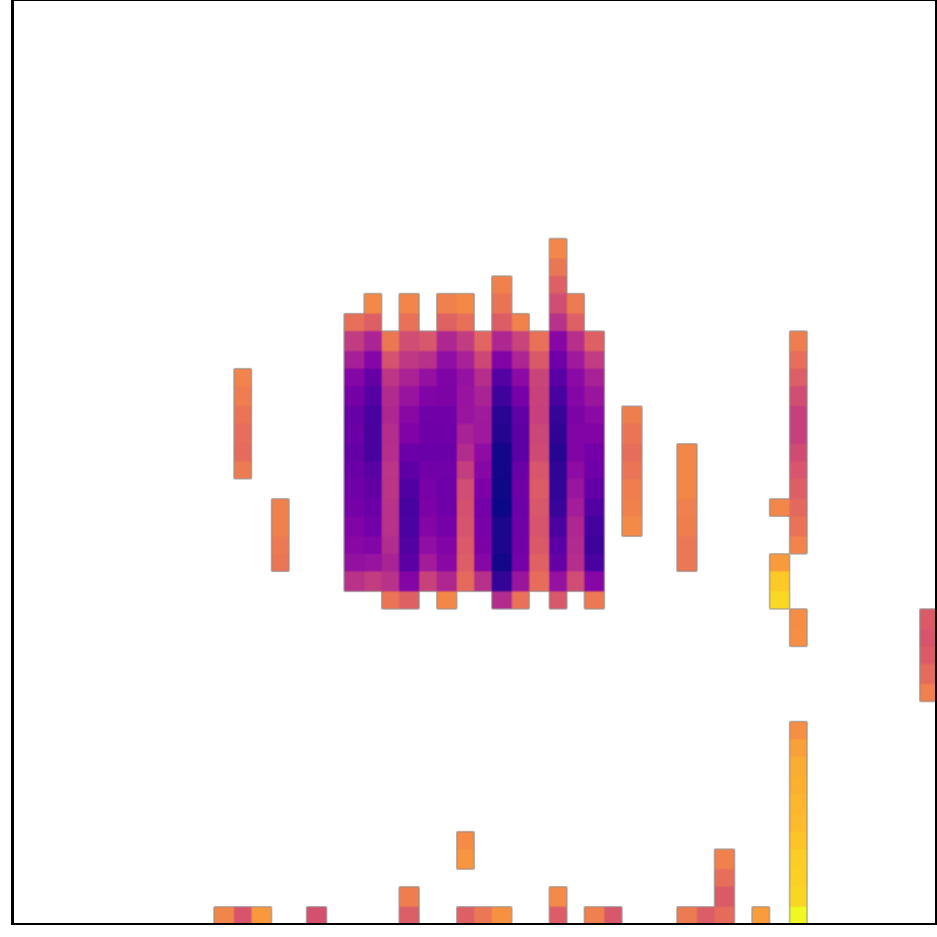}

    \end{elasticrow}
\vskip\imagepaddingtiny
\begin{elasticrow}[\imagepaddingy]
    \elasticfigure{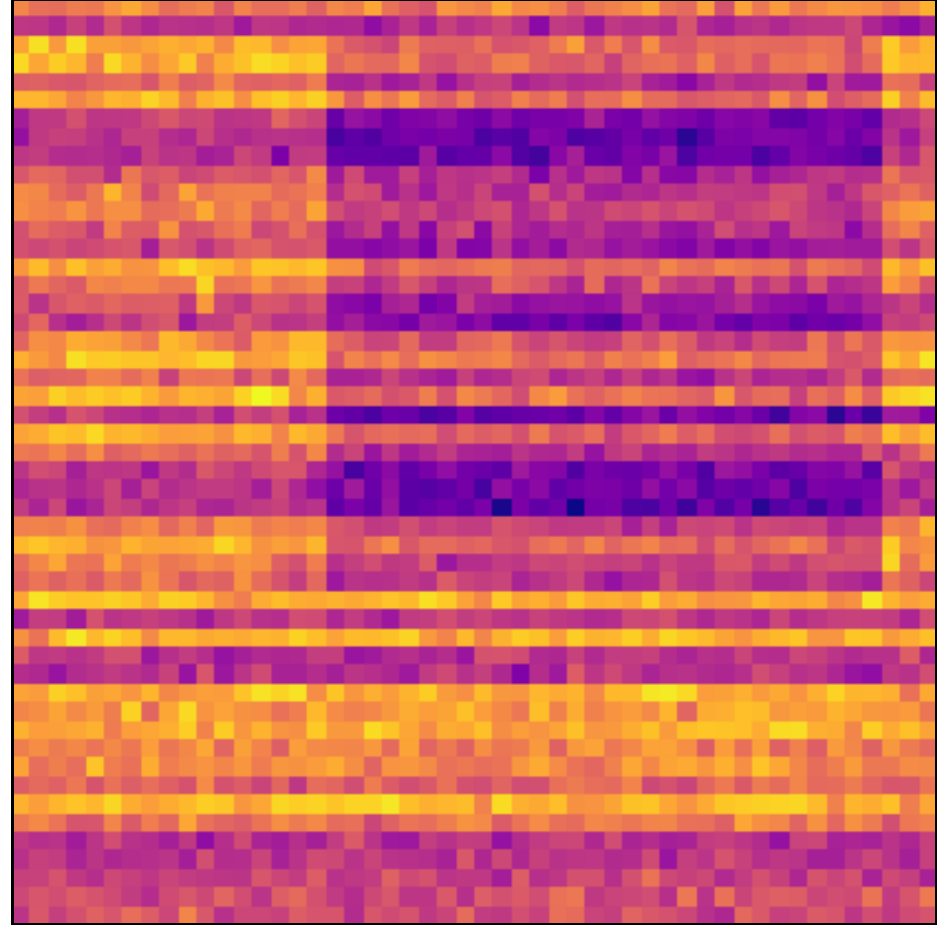}
    \elasticfigure{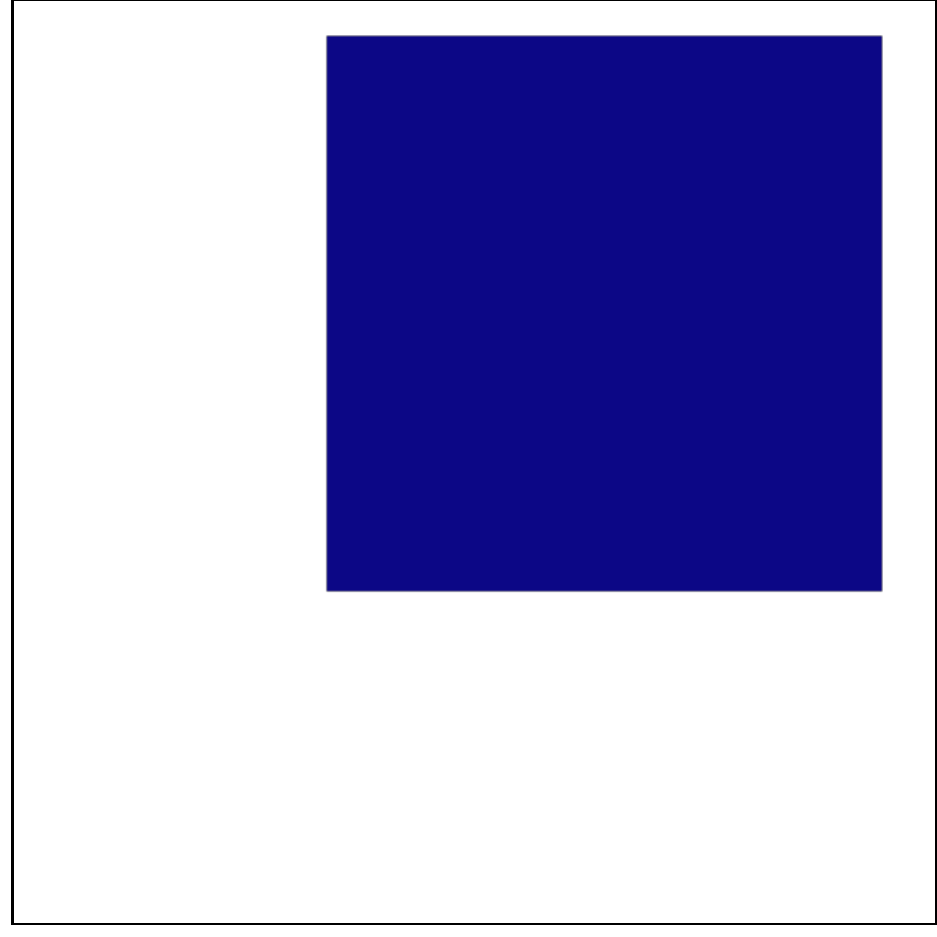}
    \elasticfigure{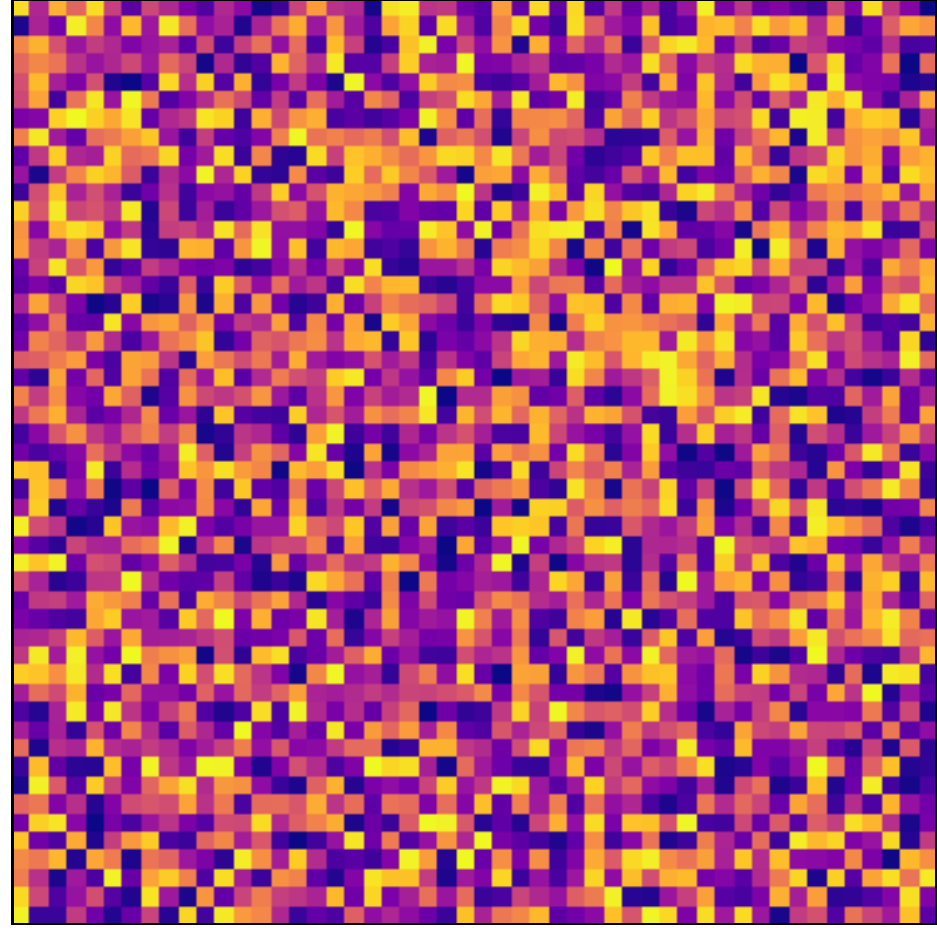}
    \elasticfigure{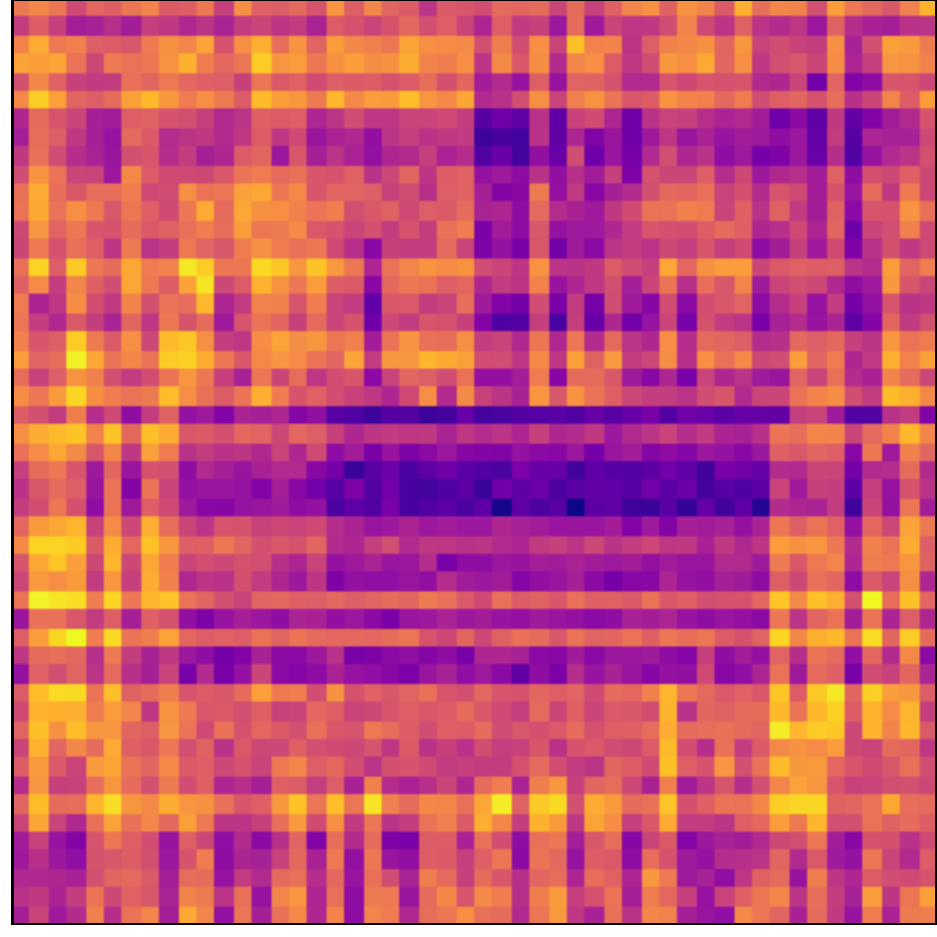}
    \elasticfigure{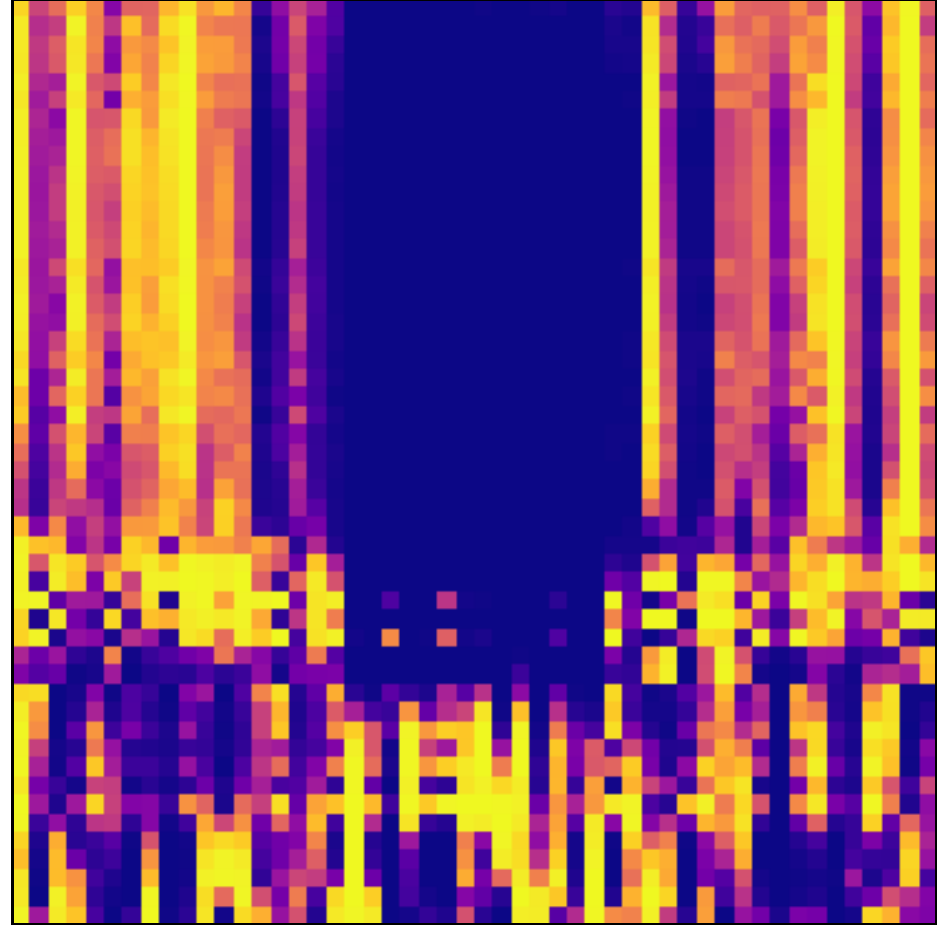}
    \elasticfigure{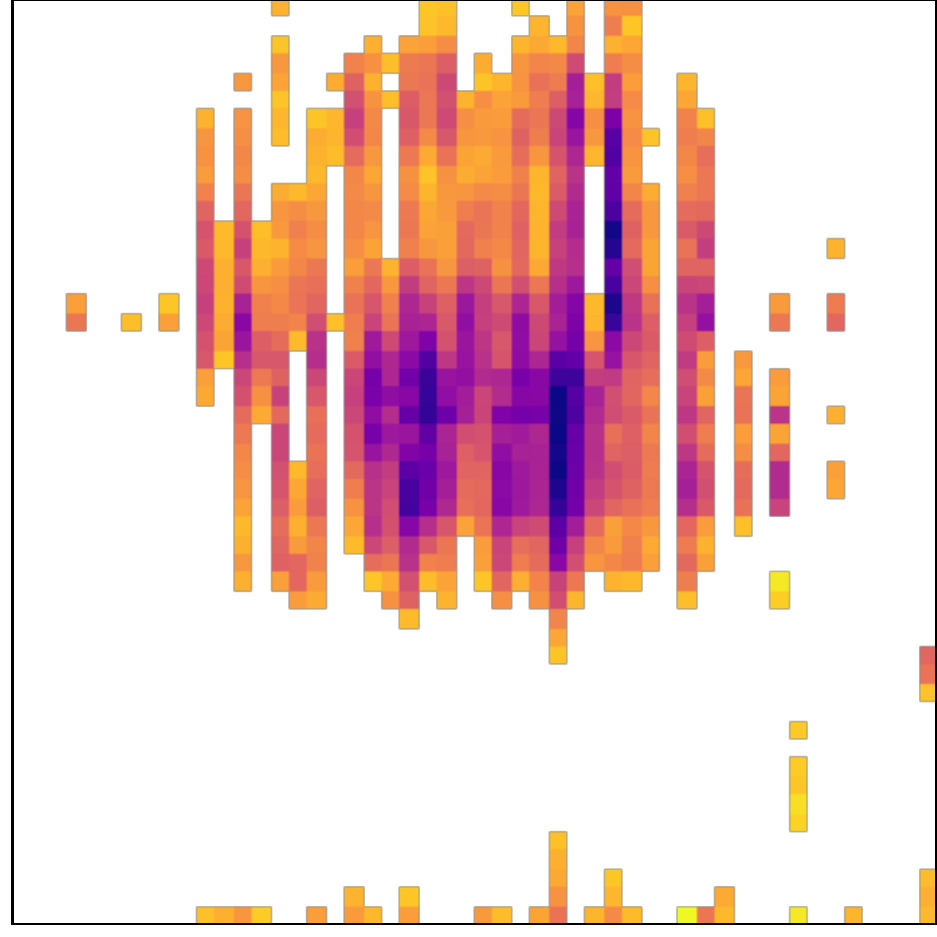}
\end{elasticrow}
\vskip\imagepaddingtiny
\begin{elasticrow}[\imagepaddingy]
    \elasticfigure{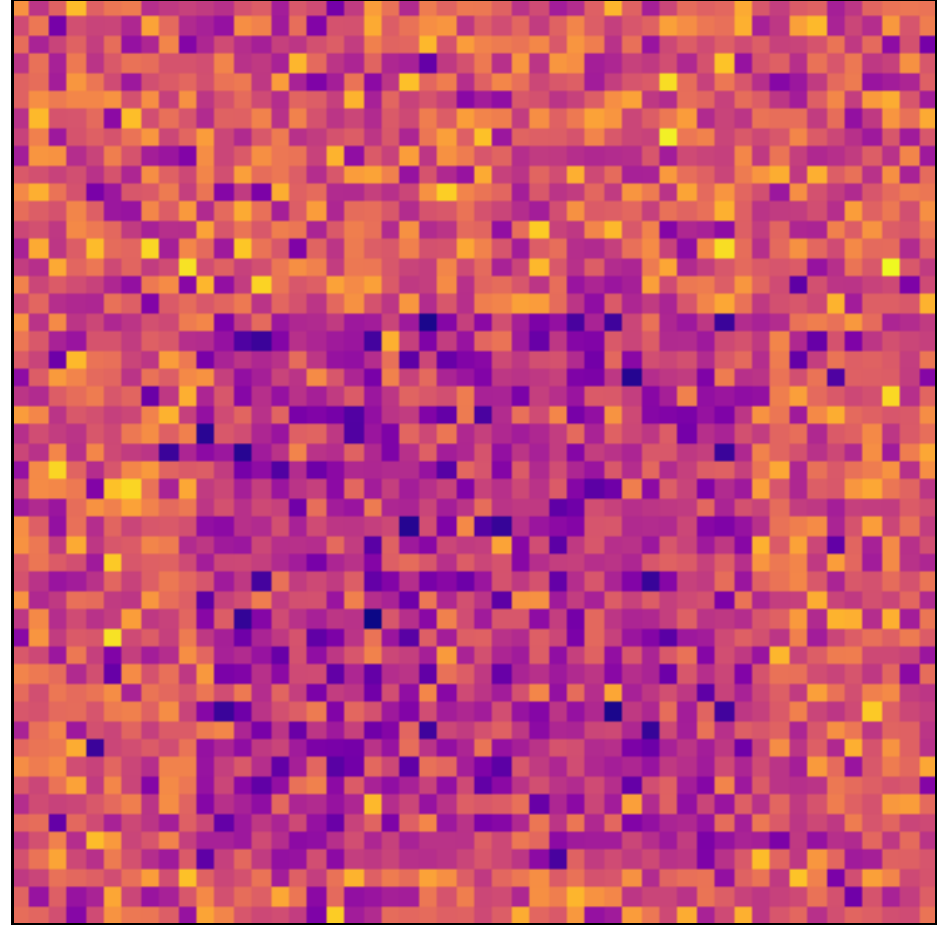}
    \elasticfigure{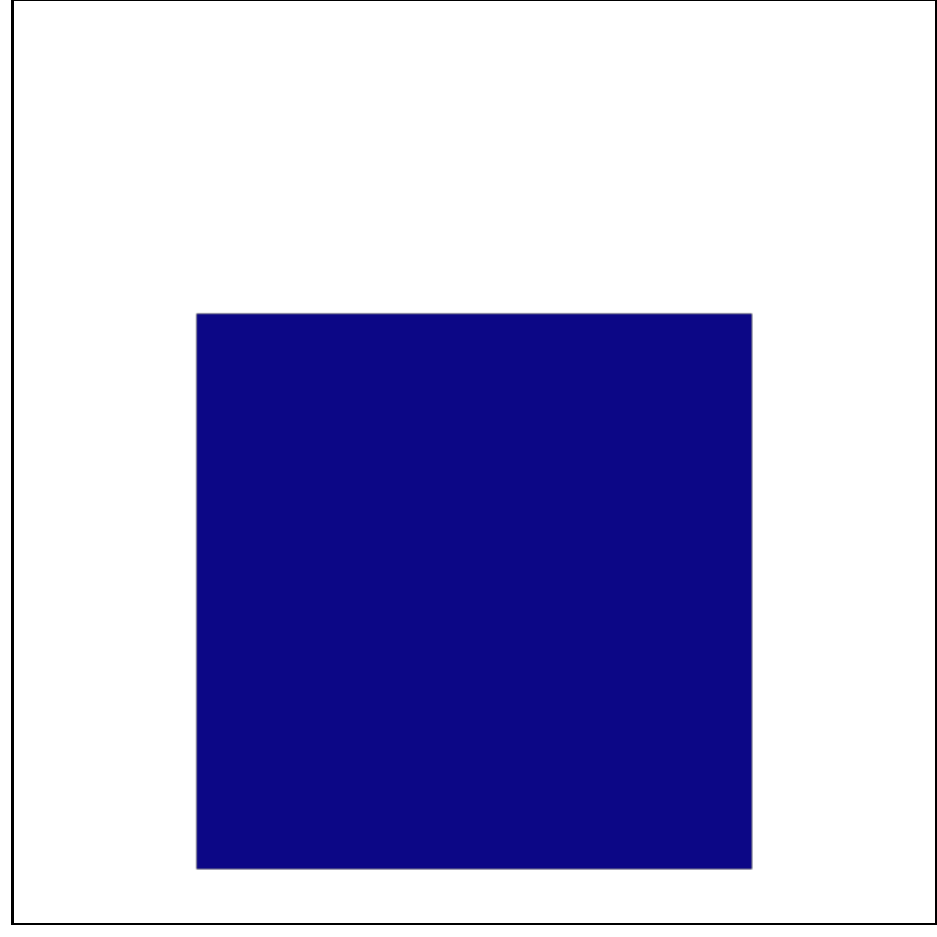}
    \elasticfigure{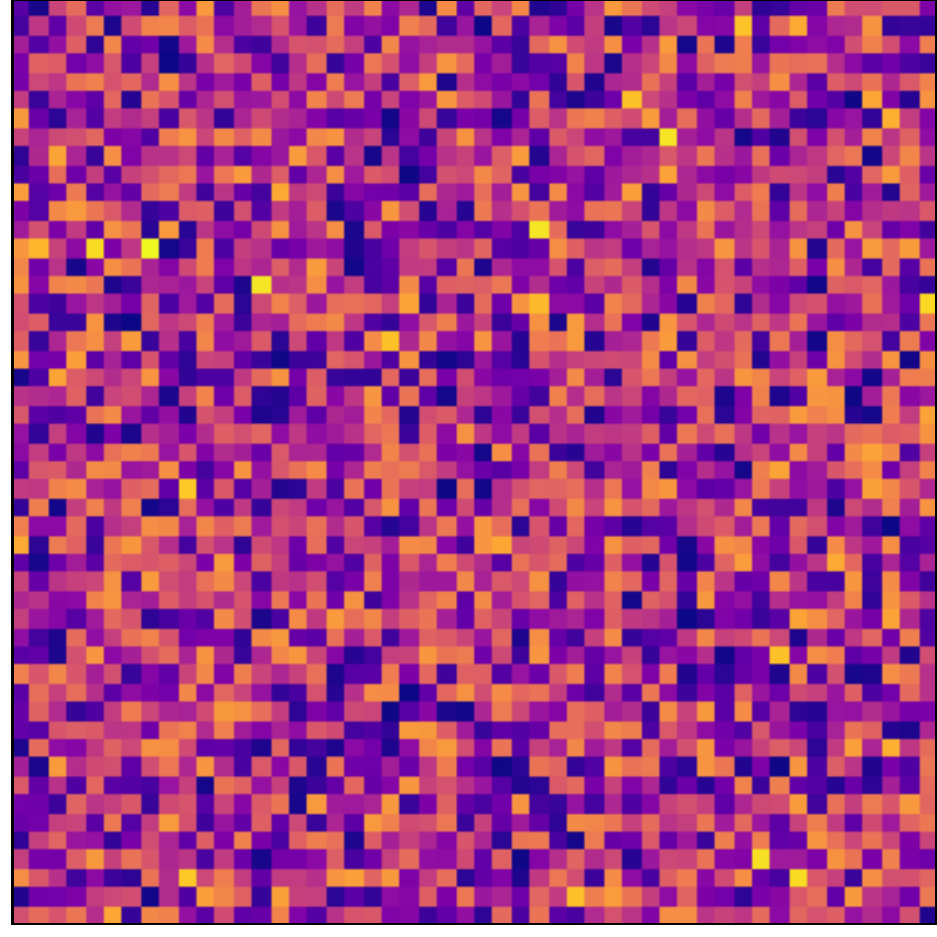}
    \elasticfigure{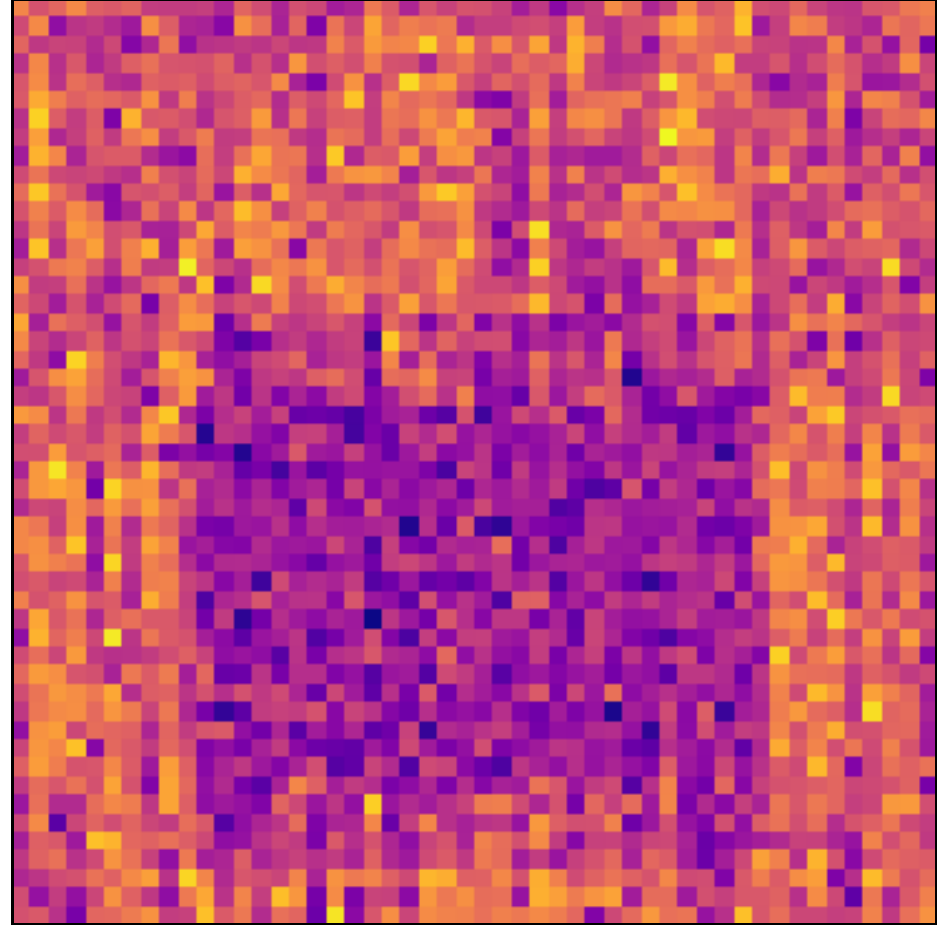}
    \elasticfigure{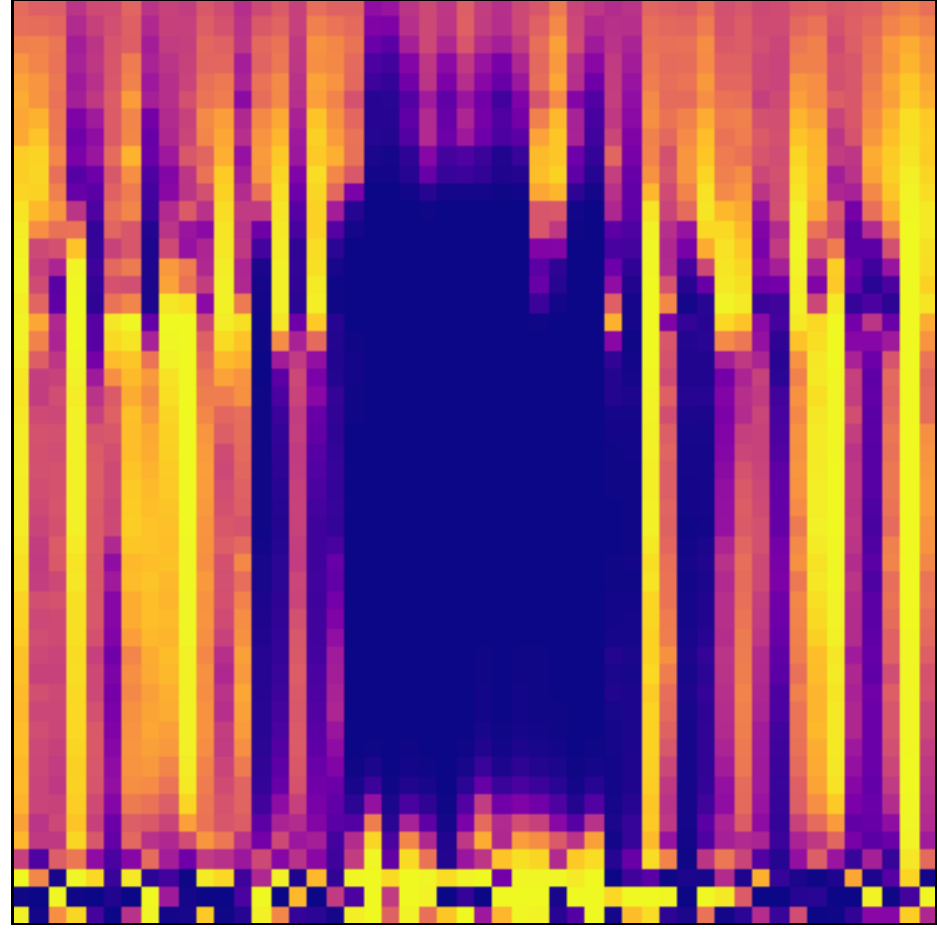}
    \elasticfigure{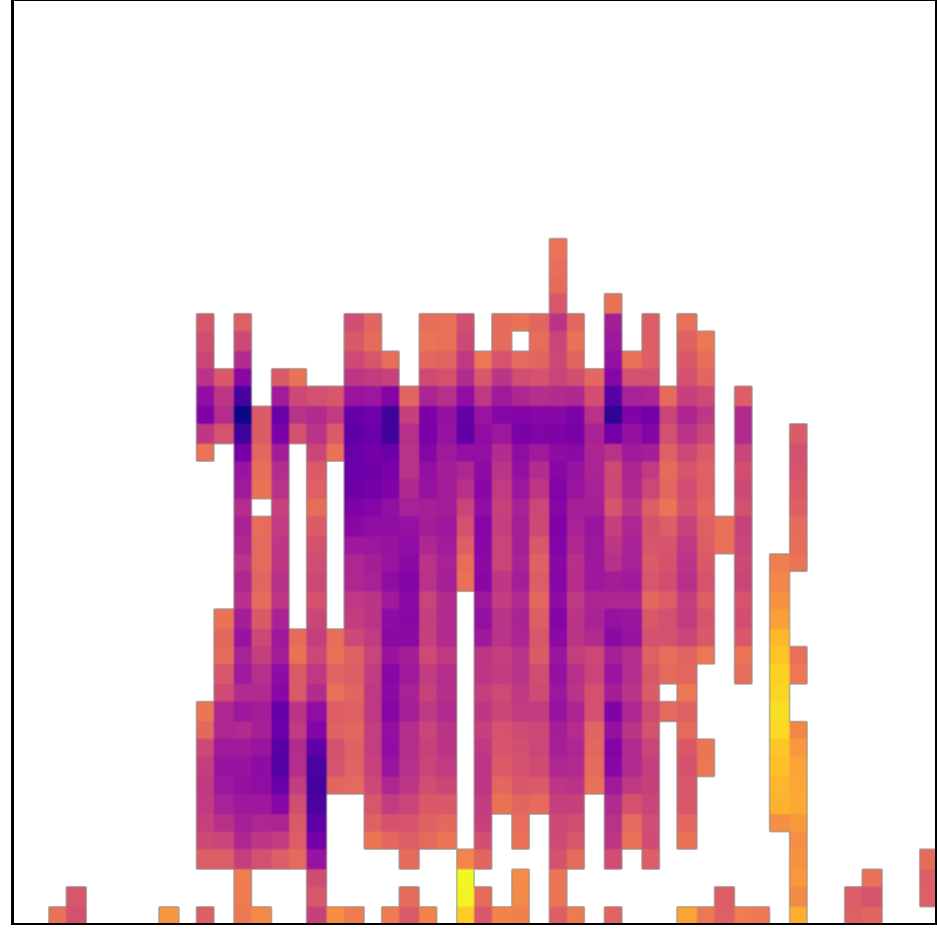}
\end{elasticrow}

\alignedlabel{0.13\linewidth}{Original}\hfill
\alignedlabel{0.13\linewidth}{Salient}\hfill
\alignedlabel{0.13\linewidth}{ICS}\hfill
\alignedlabel{0.13\linewidth}{GAN}\hfill
\alignedlabel{0.13\linewidth}{CounteRGAN}\hfill
\alignedlabel{0.13\linewidth}{SPARCE}\hfill
%    Original & Salient & ICS & GAN & CounteRGAN & \ours (Ours) & Ours $\lambda_{4,5} = 0$
    \caption{Visualization of predefined salient inputs in comparison to counterfactual modifications for the \textit{Moving Box} dataset.}
    \label{fig:salient2}
   
\end{figure}

Figure \ref{fig:salient2} visualizes heatmaps for four additional queries from the \textit{Moving Box} dataset in comparison to counterfactual modifications made by ICS, GAN, CounteRGAN and our approach (SPARCE). In this dataset, salient inputs are known in advance. A reasonable counterfactual explanation shows a high overlap of non-zero modifications and salient inputs (second column), as well as zero-modifications and non-salient inputs. In the figure, zero-residuals are plotted in white. The figure demonstrates that our approach generates sparse counterfactuals with a clear distinction between salient and non-salient time steps and features. The modifications made by our approach show a high overlap with the original saliency heatmap. Other approaches also partially recognize salient inputs, but modify salient as well as non-salient inputs. Although these approaches also successfully change the class label and remain similar to the query, they fail to create sparse counterfactuals. As pointed out in the main paper, sparsity is a key factor in ensuring the actionability of counterfactual explanations.

\subsubsection{Influence of Different Target Classes}

In contrast to other approaches, our classifier is not trained to distinguish samples belonging to the target class from other samples. Instead, it learns to discriminate all classes that are present in the original dataset. As a consequence, we can flexibly switch the target class for counterfactuals without retraining the classifier. Table \ref{tab:motionsense-target-class} denotes the performance of our approach on all metrics for the four action categories of \textit{MotionSense} (dws: walking downstairs, ups: walking upstairs, wlk: walking, jog: jogging). Our approach is able to generate counterfactuals with high informative value regardless of the selected target class. The results for \textit{Catching} and \textit{Moving Box} datasets are provided in Tables \ref{tab:catching-target-class}, \ref{tab:simulated-target-class}.

\begin{table}[h!]
  \caption{Influence of different target classes for \textit{MotionSense}}
  \label{tab:motionsense-target-class}
  \centering
  \footnotesize
  \begin{tabularx}{\linewidth}{XCCCC}
    \toprule
    %\cmidrule(r){1-5}
       $\targetclass$: & dws & ups & wlk & jog \\
    \midrule
    Precision  & 0.04 $\pm$ .04 & 0.04 $\pm$ .06 & 0.19 $\pm$ .08 &  0.04 $\pm$ .07 \\
    Similarity & 0.21 $\pm$ .10 & 0.22 $\pm$ .13 & 0.21 $\pm$ .09 & 0.24 $\pm$ .09  \\
    Sparsity  & 0.35 $\pm$ .15 & 0.22 $\pm$ .14 &  0.42 $\pm$ .17 &   0.45 $\pm$ .15  \\
    Smoothness & 0.02 $\pm$ .01 & 0.04 $\pm$ .03 & 0.03 $\pm$ .01 &  0.03 $\pm$ .01    \\
    \bottomrule
  \end{tabularx}
\end{table}

\begin{center}

\begin{table}[h!]
\begin{minipage}[c]{0.48\linewidth}
  \caption{Influence of different target classes for \textit{Catching}}
  \label{tab:catching-target-class}
  \centering
  \footnotesize  
  \begin{tabularx}{\textwidth}{lRRR}
    \toprule
    %\cmidrule(r){1-5}
       $\targetclass$: &  Ataxia & Autism & Control \\
    \midrule
    Precision  & 0.11 & 0.04  & 0.01           \\
    Similarity & 0.15 & 0.13 & 0.09     \\
    Sparsity  & 0.47 & 0.42 &  0.27       \\
    Smoothness & 0.01 & 0.01 & 0.01       \\
    \bottomrule
  \end{tabularx}
\end{minipage}
\hfill
\begin{minipage}[c]{0.48\linewidth}
  \caption{Influence of different target classes for \textit{Moving Box}}
  \label{tab:simulated-target-class}
  \centering
  \footnotesize  
  \begin{tabularx}{\textwidth}{lRR}
    \toprule
    %\cmidrule(r){1-5}
       $\targetclass$: & 0 & 1 \\
    \midrule
    Precision  & 0.00 & 0.00 \\
    Similarity & 0.52 & 0.40 \\
    Sparsity  & 0.33 & 0.30 \\
    Smoothness & 0.02 & 0.02 \\
    \bottomrule
  \end{tabularx}
\end{minipage}
\end{table}

\end{center}

\subsubsection{Influence of Regularization Terms}
In addition to the similarity constraint used in most related approaches, we apply sparsity and jerk regularization. To investigate the importance of these additional factors, we report results of experimental runs activating either only $\advloss$, $\clsloss$ and $\simloss$ ($\lambda_{1,2,3} = 1 $), or additionally $\sparseloss$ ($\lambda_{1,2,3,4} = 1 $), or additionally $\jerkloss$ ($\lambda_{1,2,3,5} = 1 $) or all regularization terms ($\lambda_{1,2,3,4,5} = 1 $). As shown in Table \ref{tab:catching-lambda-influence} the sparsest counterfactuals for the \textit{Catching} dataset are generated activating all five loss terms. However, the most significant difference in sparsity between other approaches and ours does not come from the $\lone$ sparsity regularization, but from the sparsity layer. The meaningfulness of jerk regularization can vary between datasets and can - as shown here - easily be switched off if high differences between features in subsequent time steps are expected or even desired. Results for the other datasets are reported in Tables \ref{tab:motionsense-lambda-influence}, \ref{tab:simulated-lambda-influence}.

\begin{table}[h!]
  \caption{Influence of regularization terms for \textit{Catching}}
  \label{tab:catching-lambda-influence}
  \centering
  \footnotesize
  \begin{tabularx}{\linewidth}{XCCCC}
    \toprule
    %\cmidrule(r){1-5}
         & $\lambda_{1,2,3} = 1 $  & $\lambda_{1,2,3,4} = 1 $ & $\lambda_{1,2,3,5} = 1 $ & $\lambda_{1,2,3,4,5} = 1 $ \\
    \midrule
    Precision  & \textbf{0.01} $\pm$ .01 & \textbf{0.01} $\pm$ .01 & 0.02 $\pm$ .03 & \textbf{0.01} $\pm$ .01           \\
    Similarity & \textbf{0.08} $\pm$ .02 & 0.14 $\pm$ .04 & 0.10 $\pm$ .03 & 0.09 $\pm$ .04       \\
    Sparsity  & 0.32 $\pm$ .06 & 0.33 $\pm$ .13 & 0.28 $\pm$ .14 & \textbf{0.27} $\pm$ .10      \\
    Smoothness & \textbf{0.01} $\pm$ .00 & 0.02 $\pm$ .01 & \textbf{0.01} $\pm$ .01 & \textbf{0.01} $\pm$ .01       \\
    \bottomrule
  \end{tabularx}
\end{table}

\begin{table}[h!]
  \caption{Influence of regularization terms for \textit{MotionSense}}
  \label{tab:motionsense-lambda-influence}
  \centering
  \footnotesize
  \begin{tabular}{lllll}
    \toprule
    %\cmidrule(r){1-5}
         & $\lambda_{1,2,3} = 1 $  & $\lambda_{1,2,3,4} = 1 $ & $\lambda_{1,2,3,5} = 1 $ & $\lambda_{1,2,3,4,5} = 1 $ \\
    \midrule
    Precision  & \textbf{0.01} $\pm$ .01 & 0.03 $\pm$ .04 & 0.03 $\pm$ .04 & 0.04 $\pm$ .06           \\
    Similarity & 0.25 $\pm$ .15 & 0.23 $\pm$ .11 & 0.29 $\pm$ .13 & \textbf{0.22} $\pm$ .13       \\
    Sparsity  & \textbf{0.22} $\pm$ .11 & \textbf{0.22} $\pm$ .08 & 0.32 $\pm$ .16 & \textbf{0.22} $\pm$ .14      \\
    Smoothness & 0.06 $\pm$ .04 & \textbf{0.03} $\pm$ .03 & \textbf{0.03} $\pm$ .03 & 0.04 $\pm$ .03      \\
    \bottomrule
  \end{tabular}
\end{table}

\begin{table}[h!]
  \caption{Influence of regularization terms for \textit{Moving Box} (only one repetition for columns 1-3)}
  \label{tab:simulated-lambda-influence}
  \centering
  \footnotesize
  \begin{tabular}{lllll}
    \toprule
    %\cmidrule(r){1-5}
         & $\lambda_{1,2,3} = 1 $  & $\lambda_{1,2,3,4} = 1 $ & $\lambda_{1,2,3,5} = 1 $ & $\lambda_{1,2,3,4,5} = 1 $ \\
    \midrule
    Precision  & \textbf{0.00} & \textbf{0.00} & \textbf{0.00} & \textbf{0.00} $\pm$ .00            \\
    Similarity & 0.39 & \textbf{0.38} & 0.42 & 0.40 $\pm$ .06       \\
    Sparsity  & \textbf{0.26} & 0.35 & 0.29 & 0.30 $\pm$ .05       \\
    Smoothness & \textbf{0.02} & \textbf{0.02} & \textbf{0.02} & \textbf{0.02} $\pm$ .00       \\
    \bottomrule
  \end{tabular}
\end{table}

\subsection{Dataset Statistics}
All datasets used in this work contain multivariate time series. Sequences are either truncated to equal lengths or are of equal length by design. Table \ref{tab:dataset-samples} denotes the number of samples per class, the total number of samples and the number of features and time steps for each dataset. Note that for \textit{MotionSense}, classes 2 and 3 (sitting and standing) are excluded in all experiments.

\begin{table}[h!]
  \caption{Samples per class for all datasets}
  \label{tab:dataset-samples}
  \centering
\footnotesize
    \begin{tabularx}{\linewidth}{lrrrRRRRRR}
    \toprule
    %\cmidrule(r){1-5}
%         & \# Samples & \# Time & \# Features & $\targetclass{=}0$ &
         & Samples & Time steps & Features & $\targetclass{=}0$ &          $\targetclass{=}1$ & $\targetclass{=}2$ & $\targetclass{=}3$ & $\targetclass{=}4$ & $\targetclass{=}5$ \\
    \midrule
    \textit{Catching} & 1\,975 & 60 & 20 & 522 & 789 & 664 & - & - & -    \\
    \textit{MotionSense} & 13\,950 & 100 & 12 & 1\,283 & 1\,317 & 3\,364 & 3\,042 & 1\,536 & 3\,408 \\
    \textit{Moving} Box & 17\,600 & 50 & 50 & 8\,806 & 8\,794 & - & - & - & -  \\
    \bottomrule
  \end{tabularx}
\end{table}

\subsection{Computing Times}
All experiments are conducted on a Dell Precision 7920 Tower computer with an Intel Xeon Silver processor running at 2.10 GHz, 64 GB RAM, Windows 10 and an NVIDIA Quadro RTX 5000 GPU. Computing times include one complete training run (for GAN-based models) and the generation of counterfactuals for the entire test set (Table \ref{tab:computing-times}).

 \begin{table}[h!]
  \caption{Computing times (in seconds) for all dataset}
  \label{tab:computing-times}
  \centering
  \footnotesize
  \begin{tabularx}{\linewidth}{lRRRR}
    \toprule
    %\cmidrule(r){1-5}
         & ICS  & GAN & CounteRGAN & Ours \\
    \midrule
    \textit{Catching} & 693 $\pm$ 34 & 432 $\pm$ 5 & 435 $\pm$ 7 & 398 $\pm$ 4       \\
    \textit{MotionSense} & 2\,678 $\pm$ 13 & 1\,483 $\pm$ 21 & 1\,660 $\pm$ 23 & 1\,669 $\pm$ 32       \\
    \textit{Moving Box} & 1\,755 $\pm$ 5 & 22\,789 $\pm$ 1\,071 & 22\,290 $\pm$ 1\,198 & 21\,138 $\pm$ 914       \\
    
    \bottomrule
  \end{tabularx}
 \end{table}

\end{document}